%% file: main.tex
\documentclass[twoside,11pt]{article}

\usepackage{blindtext}

\usepackage{jair2e}
\usepackage{graphicx}
\usepackage{booktabs}
\usepackage{tabularx}
\usepackage{multirow}
\usepackage{longtable}
\usepackage{array}
\usepackage{caption}
\usepackage{subcaption}
\usepackage{enumitem}
\usepackage{pifont}
\usepackage[dvipsnames]{xcolor}

\usepackage{lastpage}

\ShortHeadings{Giving AI Personalities Leads to More Human-Like Reasoning}{Nighojkar, Moydinboyev, Duong, Licato}

\begin{document}

\title{Giving AI Personalities Leads to More Human-Like Reasoning}

\author{\name Animesh Nighojkar \email anighojkar@usf.edu\\
    \name Bekhzodbek Moydinboyev \email bmoydinboyev@usf.edu\\
    \name My Duong \email myduong@usf.edu\\
    \name John Licato \email licato@usf.edu\\
    \addr Advancing Machine and Human Reasoning (AMHR) Lab\\
    Department of Computer Science and Engineering, University of South Florida\\
    Tampa, FL, USA
}



\maketitle

\begin{abstract}
In computational cognitive modeling, capturing the full spectrum of human judgment and decision-making processes, beyond just optimal behaviors, is a significant challenge. This study explores whether Large Language Models (LLMs) can emulate the breadth of human reasoning by predicting both intuitive, fast System 1 and deliberate, slow System 2 processes. Unlike traditional AI research focused on optimizing accuracy, this paper investigates the potential of AI to mimic diverse reasoning behaviors across a human population, addressing what we call the {\em full reasoning spectrum problem}. We designed reasoning tasks using a novel generalization of the Natural Language Inference (NLI) format to evaluate LLMs' ability to replicate human reasoning. The questions were crafted to elicit both System 1 and System 2 responses. Human responses were collected through crowd-sourcing and the entire distribution was analyzed and modeled, rather than just the majority of the answers. We used personality-based prompting inspired by the Big Five personality model to elicit AI responses reflecting specific personality traits, capturing the diversity of human reasoning, and exploring how personality traits influence LLM outputs. Combined with genetic algorithms to optimize the weighting of these prompts, this method was tested alongside traditional machine learning models. The results show that LLMs can mimic human response distributions, with open-source models like Llama and Mistral unexpectedly outperforming proprietary GPT models. Personality-based prompting, especially when optimized with genetic algorithms, significantly enhanced LLMs' ability to predict human response distributions, suggesting that capturing suboptimal, naturalistic reasoning may require modeling techniques incorporating diverse reasoning styles and psychological profiles. The study concludes that personality-based prompting combined with genetic algorithms is promising for enhancing AI's \textit{human-ness} in reasoning, proposing a new methodology for studying and applying human reasoning by acknowledging and leveraging the vast differences in individual reasoning styles at a granular level.
\end{abstract}


\input{latex/introduction}
\input{latex/background}
\input{latex/survey}
\input{latex/experiments}
\input{latex/conclusion}


\acks{This material is based on work supported by the National Science Foundation under Grant No. 2311286.}
\input{latex/appendix}

\clearpage
\vskip 0.2in
\bibliographystyle{apacite}
\bibliography{animesh,bib,john}

\end{document}

%% file: latex/introduction.tex
\section{Introduction}\label{sec:introduction}

    Capturing the nuanced, often imperfect, and highly diverse reasoning processes of humans presents significant challenges. Traditional uses of AI for this purpose have focused primarily on optimizing accuracy and efficiency, but this approach often neglects the complexity of human cognition, where decisions can be significantly influenced by intuition, emotion, and prior experiences. The central issue is that current AI models struggle to capture the full spectrum of human reasoning, particularly in problem spaces where decisions are not simply correct or incorrect but rather involve a plurality of diverse thought processes. Addressing this limitation is crucial for advancing AI systems that can genuinely understand and interact with humans on a deeper, more intuitive level, making it a critical area of research that extends beyond technical optimization.

    Understanding and predicting human reasoning is as fascinating now as it has been for thousands of years \citep{Wason1966-WASR, aristotle_organon_2013}. Dual Process Theory, one of the most influential theories in contemporary cognitive science \citep{stanovich_individual_2000, Kahneman2011}, distinguishes between two types of cognitive processes: System 1 and System 2.\footnote{Although in this paper we will use the more popular terms System 1 and System 2, it should be noted that they are somewhat misleading, in that they imply a coordinated system of processes working together. For this reason, some authors prefer the terminology ``Type 1'' and ``Type 2,'' e.g., \cite{Evans2018}.} System 1 operates automatically, rapidly, and effortlessly, guiding intuitive and habitual decisions without conscious control. In contrast, System 2 is slower, more deliberate, and responsible for managing mental activities that require focused attention, such as complex calculations and deliberate decision-making. These dual processes are fundamental to understanding how humans navigate decisions, both trivial and significant, and provide a framework for assessing whether AI can genuinely replicate human-like reasoning.


    Dual process theories have been given a second look in the age of the large language model (LLM), which have become a cornerstone in AI-driven reasoning due to their tremendous success on a variety of tasks. Some researchers have suggested---or even demonstrated through experimental paradigms---that LLMs with lower complexity or simpler prompts tend to engage in System 1 reasoning, while more complex architectures or prompts encourage System 2-like reasoning \citep{kojima_large_2022, hagendorff_human-like_2023, weston_system_2023, saha_system-1x_2024, yu_distilling_2024}. However, many such claims operate under the assumption that System 1 is inferior and should be avoided, focusing instead on making AI reasoning more like human System 2. The use of datasets such as the Cognitive Reflections Test (CRT) \citep{frederick_cognitive_2005}, designed to provoke System 1 into giving the wrong answer while System 2 provides the correct one, is a reflection of this assumption. This focus on accuracy neglects the reality that human reasoning is rarely so clear-cut.

    Given these complexities, how well can our best AI approaches capture human reasoning? AI reasoning encompasses a broad range of approaches that have been described at times as efficient \citep{maclure_ai_2021}, logical \citep{hagedorn_knowledge_2020}, mechanistic, or capable of making sense of complex scenarios \citep{zollman_analyzing_2023}. However, it has also been criticized for being biased and opaque \citep{oneil_weapons_2017}. Numerous studies have aimed to reduce bias in AI reasoning and promote fairness \citep{ma_fairness-guided_2023}, yet human reasoning is naturally biased and flawed \citep{gilovich_heuristics_2002, ariely_predictably_2010, stanovich_rationality_2018}. Eliminating these flaws can therefore make AI reasoning less human-like \citep{chemero_llms_2023, prescott_understanding_2023}. Furthermore, human reasoning processes are deeply influenced by context, individual differences, and prior knowledge, making it challenging to predict human decisions based purely on logical correctness \citep{nisbett_telling_1977, kahneman_choices_1984, stanovich_individual_2000}. And thus, artificial reasoners that prioritize normative correctness by reducing errors \citep{sun_survey_2024} may become less capable of modeling human-like reasoning.
    

    In this paper, we advocate for an approach to computational cognitive modeling that considers an approach successful only to the extent that it can predict the \textit{full distribution} of naturalistic behaviors in some task of interest. We will refer to this challenge as the \textit{full reasoning spectrum problem}. 
    Perhaps nowhere is this problem more apparent than in the blurred boundaries between System 1 and System 2. Faced with a reasoning problem, individuals may arrive at the same conclusion through different cognitive paths, or different individuals may use the same cognitive process to reach different conclusions \citep{de_neys_logic_2019, Nye2021}. \citet{hamade2024designing} argue that ``achieving superhuman performance is not sufficient; AI needs to account for suboptimal actions or idiosyncratic style from humans.'' Thus, it is important not only to capture whether human reasoners get the answer right or wrong, but \textit{what types} of wrong answers they guess, in what proportions those wrong answers are given in a sample of human reasoners, and how those proportions are affected by the environments in which the individuals perform their reasoning. The \textit{full reasoning spectrum problem} is the challenge of designing artificially intelligent reasoning systems that can predict all of these.

    \begin{figure}
        \centering
        \includegraphics[width=\textwidth]{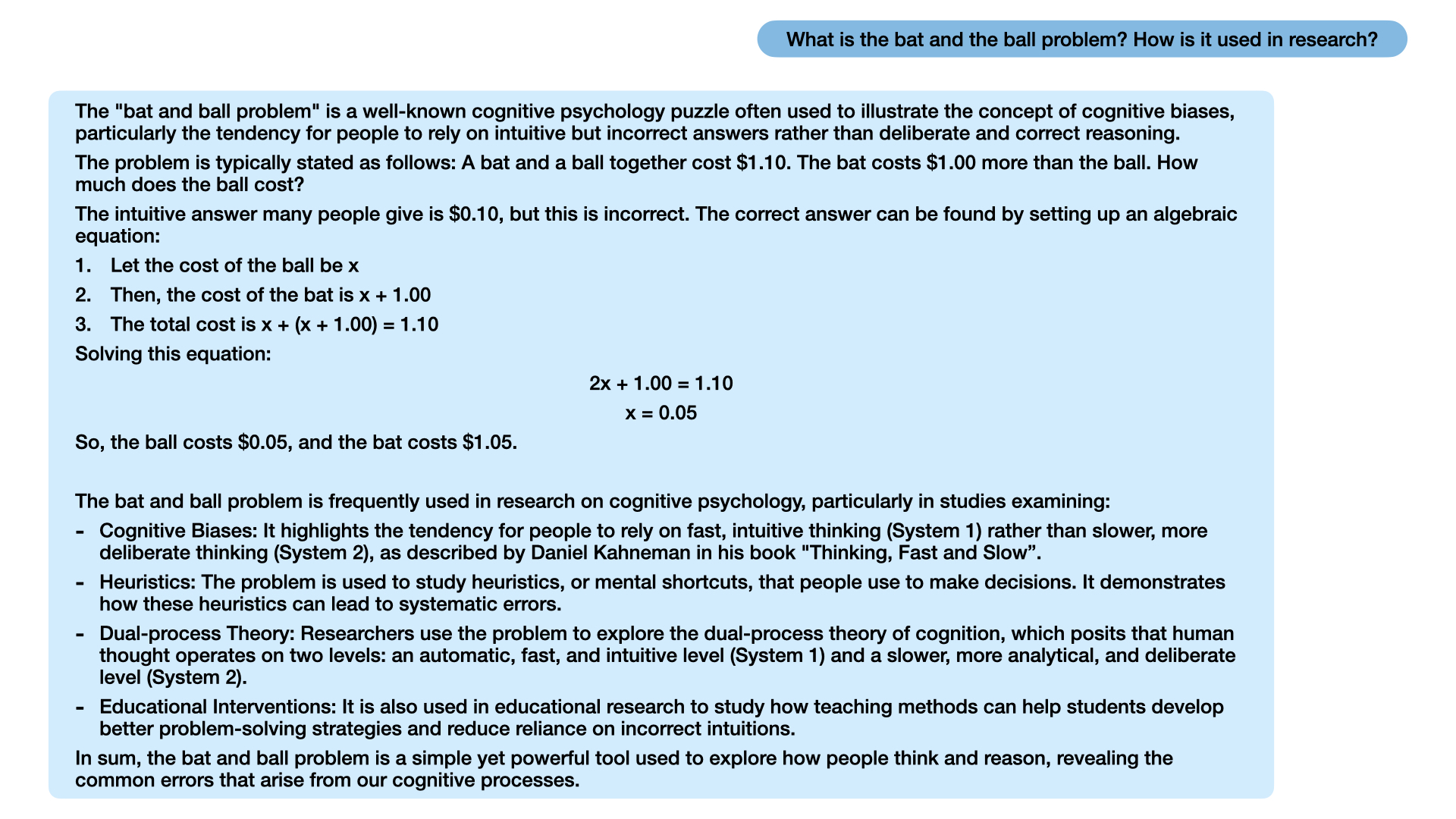}
        \caption{A question to and response by ChatGPT (GPT-4o)---without using browsing capabilities---shows that the LLM is familiar with this common question from cognitive science already, and thus questions of this type have questionable validity when used to assess its reasoning.}
        \label{fig:bat-ball}
    \end{figure}
    
    Moreover, familiarity with specific problems can significantly influence the cognitive process employed by both humans and LLMs. For humans, prior experience with a problem might lead them to rely on System 1 for quick, intuitive responses, while unfamiliarity might prompt more deliberate System 2 reasoning \citep{evans_dual-processing_2008, Kahneman2011, klein_sources_2017}. For LLMs, this raises concerns about the \textit{data contamination problem} \citep{sainz_nlp_2023, balloccu_leak_2024}, where LLMs might rely on memorized information from their training data rather than engaging in genuine reasoning. Given that many modern LLMs \citep{brown_language_2020, almazrouei_falcon_2023, jiang_mistral_2023, dubey_llama_2024} are trained on vast datasets collected from the internet \citep{penedo_refinedweb_2023, raffel_exploring_2023, together2023redpajama}, there is a significant risk that the questions used in studies were part of the LLMs' training data, further complicating efforts to assess true reasoning capabilities. To illustrate this issue, consider the study by \citet{hagendorff_human-like_2023}, which examined ChatGPT's reasoning abilities using just three problems from the CRT as templates to create 150 nearly identical problems, altering only the numbers and objects. A very simple question to ChatGPT (shown in Figure \ref{fig:bat-ball}) reveals that the LLM is familiar with not only the problem but also the conclusions researchers draw from human responses to it. This raises doubts about the validity of inferences made about LLMs' reasoning abilities. Employing LLMs to generate fresh, original questions that distinguish between System 1 (intuitive, fast) and System 2 (deliberate, slow) reasoning processes helps avoid this problem.

    \begin{figure}
        \centering
        \includegraphics[width=\textwidth]{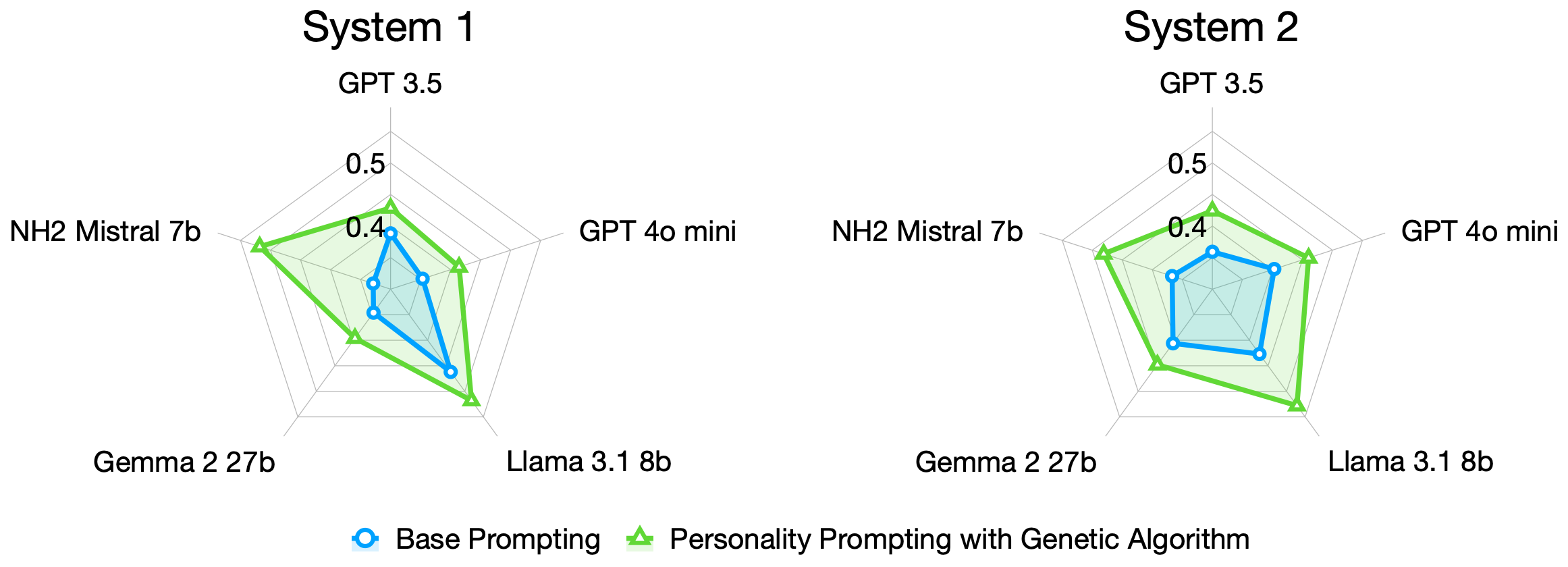}
        \caption{Similarity between human reasoning (System 1 and System 2) with LLMs' reasoning. Base prompting just prompts the LLMs for the answer and personality prompting prompts the LLM with different personality prompts and finds the best weight for each prompt using a genetic algorithm with K-fold cross validation. More details in Section \ref{sec:experiments}.}
        \label{fig:intro_results}
    \end{figure}

    We design questions in the Natural Language Inference (NLI) format (discussed in Section \ref{sec:nli}), incorporating diverse reasoning structures like syllogisms, fallacies, and belief biases \citep{evans_conflict_1983, sap_annotators_2022, yang_belief_2023}. Additionally, we modify the NLI format from three options to six, providing more granularity and nuance in human responses. Using the concept of personality prompting \citep{de_paoli_improved_2023, gu_effectiveness_2023, kamruzzaman_prompting_2024}, we aim to predict not just the majority response (or the gold label) but the \textit{entire distribution} of human responses to a given question, accounting for the likelihood that the humans participating in our study may possess different personality traits. These personality prompts are created from the traits in the Big Five Model \citep{roccas_big_2002}: openness, conscientiousness, extraversion, agreeableness, and neuroticism. We find that performance on this task improves when we allow the LLM to emulate different human personalities. Each prompt in this suite of personality prompts is assigned a weight that determines the frequency of that prompt's response in our final pool of responses. These weights (one for each prompt) are fine-tuned using a genetic algorithm \citep{mitchell_introduction_1996}, and the final distribution is compared with human responses in Figure \ref{fig:intro_results}. This entire process and its results are detailed in Section \ref{sec:experiments}.

    
    \paragraph{Contributions and Key Findings}
        \begin{itemize}
            \item We are the first to investigate whether AI (LLMs) can model System 1 and System 2 as separate, distinct reasoning processes, specifically in terms of their ability to capture the \textit{full distribution} of human responses.
            \item We introduce a six-way division of the natural language inference problem that allows for a more granular analysis of response distribution compared to existing research.
            \item We present a pioneering approach to the full reasoning spectrum problem, demonstrating the necessity of capturing the full range of correct and incorrect human responses to advance the fidelity of AI in cognitive modeling.
            \item Our study establishes a novel metric for assessing AI models' capability to encompass the entire spectrum of human reasoning, enhancing the evaluation of AI's mimicry of diverse cognitive processes from intuitive to analytical reasoning.
            \item Though Big Five personality traits have been used for personality detection with LLMs \citep{kazemeini_interpretable_2021, pellert_ai_2024, wen_affective-_2024}, we are the first to use the Big Five traits for personality prompting for NLI.
            \item We find that personality-based prompting significantly enhances LLMs' ability to mimic human response distributions compared to vanilla zero-shot prompting.
            \item The open-source Llama and Mistral models outperform closed-source GPT architectures, supporting the argument that current open-source LLMs are at least as effective as (and possibly better than) closed-source LLMs in capturing naturalistic human reasoning \citep{ahmed_studying_2024}.
        \end{itemize}

%% file: latex/background.tex
\section{Background}\label{sec:background}
    
    \subsection{System 1 vs System 2 Reasoning in LLMs}\label{sec:s1vs2}
        
        Dual Process Theory distinguishes between two types of cognitive processes: System 1, which is fast, intuitive, and automatic, and System 2, which is slower, more deliberate, and rational \citep{evans_dual-process_2013}. This framework is highly influential on current theories of human reasoning, judgment, and decision-making. For instance, classic problems like the Bat and Ball Problem \citep{frederick_cognitive_2005}, also shown in Figure \ref{fig:bat-ball}, demonstrate how System 1 can produce rapid but often flawed responses, while System 2 enables more accurate outcomes through reflective thinking \citep{evans_dual-processing_2008, evans_reflections_2019}.

        Recent advancements in AI have sought to replicate System 1 and System 2 processes using various prompting strategies in LLMs \citep{Nye2021, kojima_large_2022, hagendorff_human-like_2023, weston_system_2023, saha_system-1x_2024, yu_distilling_2024}, with mixed results \citep{vatsal_survey_2024}. Researchers have attempted to simulate System 1 and System 2 reasoning in LLMs by experimenting with common prompt templates, such as zero-shot and chain-of-thought prompts \citep{kojima_large_2022, wei_chain_2022}. Additionally, efforts to emulate System 2 thinking have led to the development of novel prompting techniques that expand on chain-of-thought, mimicking aspects of human problem-solving processes using search algorithms \citep{wang_plan-and-solve_2023, wang_self-consistency_2023, yao_tree_2023, zhang_cumulative_2024}. In contrast, modeling System 1-like processes in LLMs has involved heuristic-driven, rapid decision-making approaches \citep{hagendorff_human-like_2023}, which can generate responses without immediate reasoning steps \citep{yu_distilling_2024}. On the other hand, simulating System 2 requires more complex logical reasoning and problem-solving tasks that involve deliberate, step-by-step processing \citep{Nye2021}. Some researchers have also explored hybrid models that integrate both cognitive systems, aiming to harness their combined strengths for more effective decision-making in AI \citep{saha_system-1x_2024}.

        In their survey, \citet{vatsal_survey_2024} examine how prompt engineering can evoke distinct reasoning and responses from LLMs. They categorize prompt engineering techniques into zero-shot, one-shot, and few-shot prompts, each with specific benefits and limitations. Zero-shot prompting involves giving the model a task description and expecting it to perform the task based on its pre-existing knowledge, though this approach may be constrained by the model's interpretation of the task description. One-shot prompting offers a single example to provide a clear reference for the task, while few-shot prompting supplies multiple examples, offering richer context and enhancing task generalization, which has been shown to significantly improve performance. The survey explores how these methods influence LLMs' effectiveness in various NLP tasks, including question answering, text generation, and language inference.
        
        A prevalent assumption in the literature is that System 1 leads to the wrong answer. System 1 processes, often seen as intuitive but error-prone, are typically simulated in LLMs through straightforward, concise zero-shot prompts \citep{kojima_large_2022, hagendorff_human-like_2023, yu_distilling_2024}. In contrast, System 2 processes, associated with more accurate and thoughtful responses, are mimicked through chain-of-thought prompting, which enhances performance on tasks like the Cognitive Reflection Test (CRT) \citep{hagendorff_human-like_2023}. However, this assumption limits the datasets used in these studies to those with clear correct or incorrect answers, overlooking the nuance of real-world reasoning. It is important to note that System 1 reasoning is not simply erroneous reasoning, but rather it is a set of reasoning processes (akin to heuristics) that has distinctive properties, and thus it can and should be modeled separately from System 2 reasoning. Moreover, System 1 processes are an inseparable step of human reasoning \citep{gilovich_heuristics_2002} and cannot be ignored. System 1 reasoning enables us to perform many tasks effortlessly, such as walking, eating, or driving, where over-analyzing every detail would be impractical or impossible. System 1 is almost effortless for humans \citep{gladwell_blink_2007} but as we demonstrate in this paper, replicating System 1 behaviors in LLMs is more complex than simply applying zero-shot prompts, especially with datasets lacking normatively correct or incorrect responses.

     \subsection{The Natural Language Inference (NLI) Format}\label{sec:nli}
        In our study, we utilize an enhanced version of the Natural Language Inference (NLI) task, building upon prior foundational work \citep{bowman_large_2015}. NLI, a successor to Recognizing Textual Entailment (RTE) \citep{quinonero-candela_pascal_2006},\footnote{RTE focused only on predicting whether the relationship between two sentences was entailment, but NLI bifurcated the ``False'' label for more granularity} is centered around evaluating whether a hypothesis $h$ can logically follow from a given premise $p$. For illustration, consider the premise $p=$ ``A crowd gathers as three blue cars begin a race.'' Possible relationships between this premise and a hypothesis $h$ could be:
        \begin{enumerate}
            \item Entailment: $h=$ ``A race is taking place.'' must be true if $p$ is true.
            \item Contradiction: $h =$ ``There are no cars racing.'' cannot be true if $p$ is true.
            \item Neutral: $h =$ ``Three men are competing in a race.'' is neither necessarily true nor necessarily false given $p$.
        \end{enumerate}

        The most common goal of NLI datasets is to study how a set of human reasoners \textit{naturally} reasons about the inferential relationships in the NLI items. LLMs are traditionally trained and benchmarked on NLI datasets to enhance their naturalistic reasoning capabilities \citep{bowman_large_2015, williams_broad-coverage_2018, nie_adversarial_2020, williams-etal-2022-anlizing}. 
        
        While typically trifurcated into distinct categories, this specific categorization can suffer from problems of underspecification and a lack of granularity \citep{nighojkar_no_2023}. Addressing these complexities, researchers have proposed various modifications to the standard categorization, including: 
        
        \begin{itemize}
            \item Introducing an ``entailment strength'' parameter, reflecting either model confidence or perceived likelihood, though such measures have historically grappled with issues like ambiguous evaluation standards and inconsistent annotator judgments \citep{chen_uncertain_2020, meissner_embracing_2021}.
            \item Examining the variability in annotator perspectives, considering whether a `neutral' judgment indicates balanced reasons for both agreement and disagreement or a complete absence of decisive evidence \citep{pavlick_inherent_2019, zhang_capturing_2021, zhang_identifying_2021, zhou_distributed_2022}.
            \item Differentiating between various degrees of entailment, such as ``absolutely entails'' versus ``likely entails,'' which echoes ongoing research into probabilistic reasoning \citep{Kahneman2011}.
        \end{itemize}

        To refine these categories further, we propose a structured NLI framework with six detailed categories, broadening the inferential spectrum from absolute contradiction to definite entailment. These categories are: (A) \textit{Absolutely must be false}, (B) \textit{Is more likely to be false}, (C) \textit{Has strong reasons to be true and strong reasons to be false}, (D) \textit{Has no reasons to be either true or false}, (E) \textit{Is more likely to be true}, and (F) \textit{Absolutely must be true}.

        This broader categorization not only clarifies the guidance provided to annotators but also enriches the nuances in the data, aligning with earlier initiatives to expand NLI’s analytical depth \citep{pavlick_inherent_2019, zhang_ordinal_2017}. This framework allows for more detailed response options while still fitting into the three traditional NLI categories, making it more flexible. The use of NLI is widespread, affecting fields such as measuring psychological traits \citep{Laverghetta2021c, Laverghetta2022, Laverghetta2022a, Laverghetta2023b, Laverghetta2023c}, comparing text similarity \citep{nighojkar_mutual_2021}, and evaluating the quality of translations and paraphrases \citep{nighojkar_improving_2021}. This demonstrates its versatility and usefulness in different areas.

%% file: latex/survey.tex
\section{Data Collection} \label{sec:data_collection}
    \subsection{Survey Design}
        
\begin{figure}[t]
            \centering
            \begin{minipage}{0.49\textwidth}
                \centering
                \includegraphics[width=\textwidth]{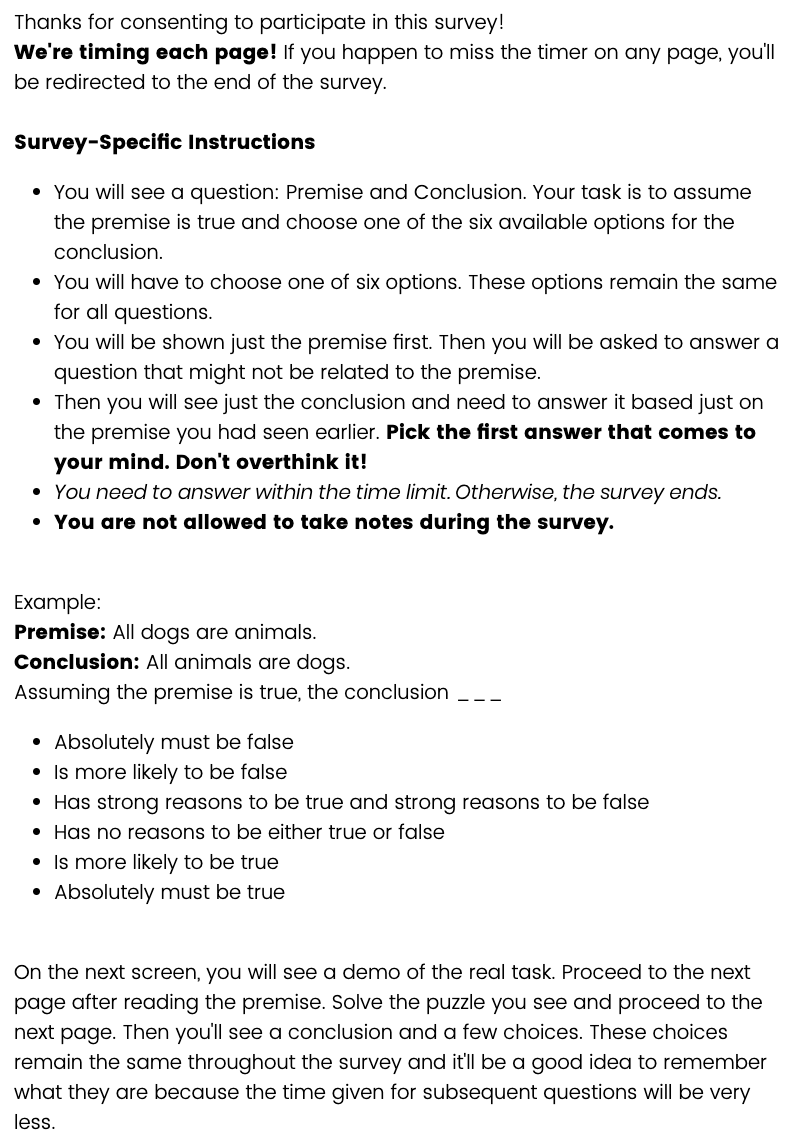}
                \caption{Survey instructions}
                \label{fig:instructions}
            \end{minipage}
            \hfill
            \begin{minipage}{0.49\textwidth}
                \centering
                \includegraphics[width=\textwidth]{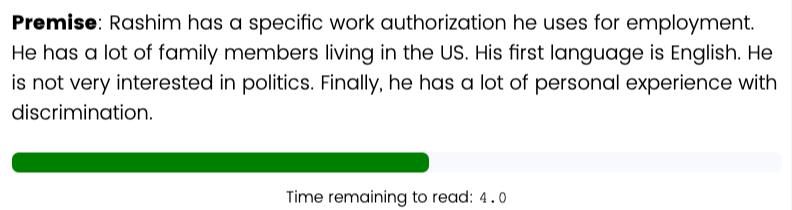}
                \caption{Survey screen 1: Premise}
                \label{fig:premise}
                \includegraphics[width=\textwidth]{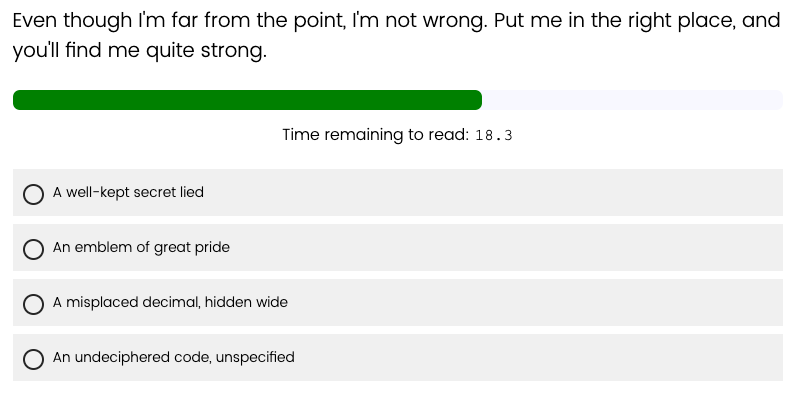}
                \caption{Survey screen 2: Distractor}
                \label{fig:distractor}
                \includegraphics[width=\textwidth]{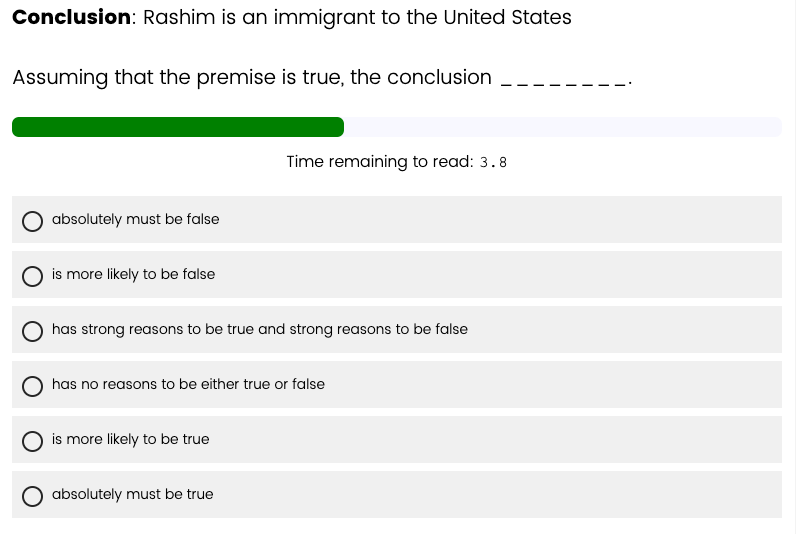}
                \caption{Survey screen 3: Conclusion}
                \label{fig:conclusion}
            \end{minipage}
           \label{fig:puzzle-distraction}
        \end{figure}

        Our study utilizes a dual-phase survey design, hosted on Qualtrics, with 60 participants recruited via Prolific. The survey consists of 27 questions in the first phase, which are repeated in the second phase, following the design of the two-response paradigm \citep{bago_fast_2017}. In this paradigm, participants are asked to provide two separate responses to the same question or task. The first response is typically their initial, instinctive answer, while the second response is given after some deliberation or additional information is provided. This method allows us to analyze the differences between intuitive and reflective thinking, or System 1 and System 2, shedding light on cognitive processes like reasoning, decision-making, and how people change their minds when given more time or information. In the first phase, aimed at eliciting System 1 reasoning, participants are presented with NLI questions while simultaneously solving a puzzle as a distraction. Each question is divided into two parts: the premise and the conclusion, with six response options (A-F) available, as detailed in Section \ref{sec:nli}. The exact instructions provided to the participants are in Figure \ref{fig:instructions}. Participants read the premise (Figure \ref{fig:premise}), solve the puzzle (Figure \ref{fig:distractor}), then read the conclusion before selecting one of the six options (Figure \ref{fig:conclusion}). Note that participants never have access to the premise and the conclusion together in this phase and need to rely on whatever they retain from the premise when answering the question about the conclusion on Screen 3. The survey begins with an example question to familiarize participants with the task. To reduce fatigue, one-minute breaks are incorporated after each question (a set of all three screens).

        In the first phase, a time limit is imposed to encourage rapid reasoning. The initial time allocated per question is determined using the following formula:
        \[ \textit{base time} = 0.0787 \times \textit{sentence length} + 0.0016 \times \textit{perplexity} + 6.3276 \]
        These coefficients are derived from ordinary least squares regression \citep{michalos_ordinary_2014}, based on data collected in-house. We recorded the time each participant (the co-authors and other volunteering students) took to read a question and trained the regression model to predict the 90th percentile reading time based on the sentence length and its perplexity \citep{jelinek_perplexitymeasure_1977}. Additionally, the reading time is adaptively adjusted based on participant pacing, either increasing or decreasing the time allotted depending on their speed, with a minimum engagement time set at 10 seconds. If a participant moves on from Screen 1 or answers the question on Screen 3 within 70\% of the base time for that screen, we reduce their time for all the following questions by 10\%. This reduction is compounded, so if the participant takes less than 70\% of the reduced time, we reduce the time for the subsequent questions by 19\% (10\% compounded twice). We never reduce the time below 10 seconds. A similar calculation works for slow participants. If they take over 90\% of the base time, we increase the time for the subsequent questions by 10\%. Failure to comply with the time limits results in warnings and may lead to survey termination after repeated infractions.
        
        The puzzles, essential for the distraction component, are generated using OpenAI's \texttt{GPT-4-0314} model. A structured query produces rhyming riddles with four potential answers, designed to engage participants critically without overwhelming them. Figures \ref{fig:premise}-\ref{fig:conclusion} illustrate the three components of a question in the first phase. The second phase omits the puzzles and combines the premise and conclusion into a single screen per question, allowing participants two minutes to carefully consider each NLI item and record their reasoning in a text box, explicitly targeting System 2 reasoning.
        
        To detect bad faith participation, embedded attention checks resembling standard questions are distributed randomly through phase 1 (questions in phase 2 follow the same order as phase 1), drawn from ChaosNLI \citep{nie_what_2020} with at least 90\% inter-annotator agreement. ChaosNLI is an NLI dataset created with 100 annotators for each question. This lets us set thresholds on inter-annotator agreement (like 90\%) that would not be possible with other NLI datasets like SNLI \citep{bowman_large_2015} or MultiNLI \citep{williams_broad-coverage_2018} that only have 5 annotations per question. If a participant's responses fail to match more than 5 out of the 8 attention check questions across both phases, their data is excluded, and a new participant is recruited to ensure 30 valid, good-faith responses per question. Additionally, responses from participants who do not complete the entire survey are disregarded.
        
    \subsection{Item Types}
    
        In Section \ref{sec:introduction}, we introduced the full reasoning spectrum problem (where existing datasets and methods fail to capture the whole spectrum of human reasoning patterns) and the data contamination problem (the possibility that publicly available datasets may be included in the training data of large language models, or LLMs). In Section \ref{sec:nli}, we also explained our decision to use NLI as our preferred format. Commonly available NLI datasets do not attempt to capture annotations that distinguish between System 1 and System 2 reasoning in humans. Additionally, it is highly likely that existing public NLI datasets have already been used to train LLMs. Consequently, these datasets suffer from both the full reasoning spectrum problem and the data contamination problem. To address these issues, we created our own dataset of NLI questions using OpenAI's GPT-4 model (specifically \texttt{GPT-4-0314}) to generate all the items. We provided the model with multiple prompts and used different OpenAI API calls to generate the premises and conclusions separately, ensuring that the model never produced both simultaneously. Finally, we made two additional API calls to rephrase the premises and conclusions individually. All our prompts are in Appendix \ref{secA:dataset_creation} along with details about our data generation pipeline. We generate different types of items to prevent monotonicity in the dataset and to prevent any pattern recognition for the humans doing our survey and the models we use to predict the human responses. These item types are explained in more detail below.
        
        \paragraph{StereoNLI}
        StereoNLI connects NLI with stereotypes, building on previous research showing that human System 1 reasoning is often biased and influenced by stereotypes \citep{Kahneman2011, geeraert_when_2013}. We utilized the StereoSet dataset \citep{nadeem_stereoset_2021} to select seed words for generating StereoNLI questions. StereoSet comprises 17,000 sentences that assess stereotype bias in language models concerning gender, race, religion, and profession. Each sentence is labeled by multiple human annotators as either `anti-stereotype', `stereotype', `unrelated', or `related'. From StereoSet, we randomly chose nouns associated with gender, race, religion, and profession, then prompted GPT-4-0314 to generate a name for a hypothetical person called `X'. We asked GPT-4 to write a paragraph with three sentences reflecting common assumptions about `X' based on the given traits. This paragraph became our premise. Excluding gender due to its limited variety in the dataset, we randomly selected one of the three traits and prompted GPT-4 to generate an assumption about `X' with a truth value. For each scenario, we created three instances---contradiction, neutral, and entailment---over three separate conversations with the GPT-4 chatbot via the OpenAI API. Finally, the premises and all three conclusions were rephrased. Table \ref{tab:item_types} presents three examples of StereoNLI items that share the same premise.

        \paragraph{Fallacy}
        Fallacies of argumentation are arguments whose logical structures that have very little to no deductive validity, despite being commonly used in informal reasoning.\footnote{For more nuanced discussions of what constitutes a fallacy, see \cite{Fearnside1959,Walton1990,Walton1985,Walton2008}.} Previous cognitive psychology research \citep{boissin_debiasing_2022} has examined how System 1 can sometimes engage in fallacious reasoning. In our dataset, we focus on three specific types of fallacies. The first, \textit{post hoc ergo propter hoc}, occurs when someone mistakenly believes that because one event follows another, the first event caused the second. The second, \textit{slippery slope}, posits that an action will set off a chain of events leading to an undesirable outcome without establishing or quantifying the relevant contingencies. This fallacy, also known as ``the domino effect,'' often implies a long series of intermediate events connecting a seemingly harmless start to an undesirable end, assuming uncertain or unlikely consequences. The third fallacy, \textit{straw person}, involves misrepresenting an opponent’s argument to make it easier to refute. We created templates to generate premise-conclusion pairs for each type of fallacy. For example, in the case of \textit{post hoc ergo propter hoc}, the premise follows the structure ``X happened right before Y,'' with the conclusion stating ``X caused Y.'' GPT-4 fills in the details for X and Y, and we then rephrase the premises and conclusions to produce the final items. Table \ref{tab:item_types} displays examples of Fallacy items, one for each fallacy.

        \paragraph{Syllogism and Stereo Syllogism}
        Syllogisms, a core component of traditional logic used in philosophical reasoning, consist of a major premise (a general statement), a minor premise (a specific statement), and a conclusion. The structure of a syllogism, or its figure, depends on the positioning of the middle term (M), subject (S), and predicate (P). The complex structure of syllogisms has led many researchers to examine its effects on eliciting System 1 and System 2 reasoning \citep{evans_conflict_1983, khemlani_theories_2012, da_silva_system_2023}. For our dataset, we selected singular nouns as seed words to guide the GPT-4 model in generating the major and minor premises while preserving the order of M, P, and S. The four primary syllogistic figures are: (1) \textit{Premise: M is P. S is M. Conclusion: S is P.} (2) \textit{Premise: P is M. S is M. Conclusion: S is P.} (3) \textit{Premise: M is P. M is S. Conclusion: S is P.} (4) \textit{Premise: P is M. M is S. Conclusion: S is P.} Finally, we rephrased the premises and conclusions to produce the final items. We also created a variant of syllogism questions using seed words from StereoSet, referred to as Stereo Syllogism. Table \ref{tab:item_types} presents four examples of Stereo Syllogism items, one for each figure.

        \paragraph{Guilt}
        Drawing on studies that demonstrate dual-process effects when participants are asked to assess the guilt of a suspect in a hypothetical scenario \citep{Peer2013, Kassin2013, Rachlinski2015, Wistrich2015, Bergius2020}, we designed questions using the following template: First, we prompted GPT to generate a few sentences describing the details of a crime for which the perpetrator has not been identified. We then asked GPT-4 to list several features likely to be true of the culprit. This enabled us to create two forms of questions: the entailment form (\textit{E Guilt}), which included the crime description, a suspect characterized by features that made them appear likely to be guilty, and a conclusion asking whether the suspect is guilty; and the contradiction form (\textit{C Guilt}), where the suspect is described as not possessing the features, or as having opposite features. Table \ref{tab:item_types} shows examples where the E Guilt premise describes a suspect who initially seems likely to have committed the crime, while the C Guilt premise describes a suspect who does not seem to fit the crime's circumstances.

        \paragraph{Primacy and Recency} 
        The first and last pieces of information presented can disproportionately influence the reader's perception due to what is known as the \textit{serial position effect} \citep{murdock_serial_1962}. To incorporate questions leveraging this effect, we began by manually writing several sentences that simply stated an individual's name and a characteristic (e.g., ``Simon is a professor''). We then asked GPT-4 to generate five sentences about that individual that would likely be true if the characteristic were accurate (\textit{likely-true}). We selected two sentences that best aligned with our archetypal conceptions of the characteristic. Next, we asked GPT-4 to generate five sentences about that individual that were likely \textit{not} true (\textit{likely-false}) and selected three of these. We then created two forms of questions: \textit{P Primacy/Recency} questions had premises starting and ending with sentences from the \textit{likely-true} category, with three \textit{likely-false} sentences in the middle, and the original sentence describing the individual and characteristic as the conclusion. \textit{N Primacy/Recency} questions had the same premise, but the conclusion was the original sentence in a negated form (e.g., ``Simon is not a professor'' instead of ``Simon is a professor''). Examples of these questions are provided in Table \ref{tab:item_types}.

    \subsection{Dataset Characteristics} \label{subsec:dataset_characteristics}
        
        \begin{figure}
            \centering
            \includegraphics[width=\textwidth]{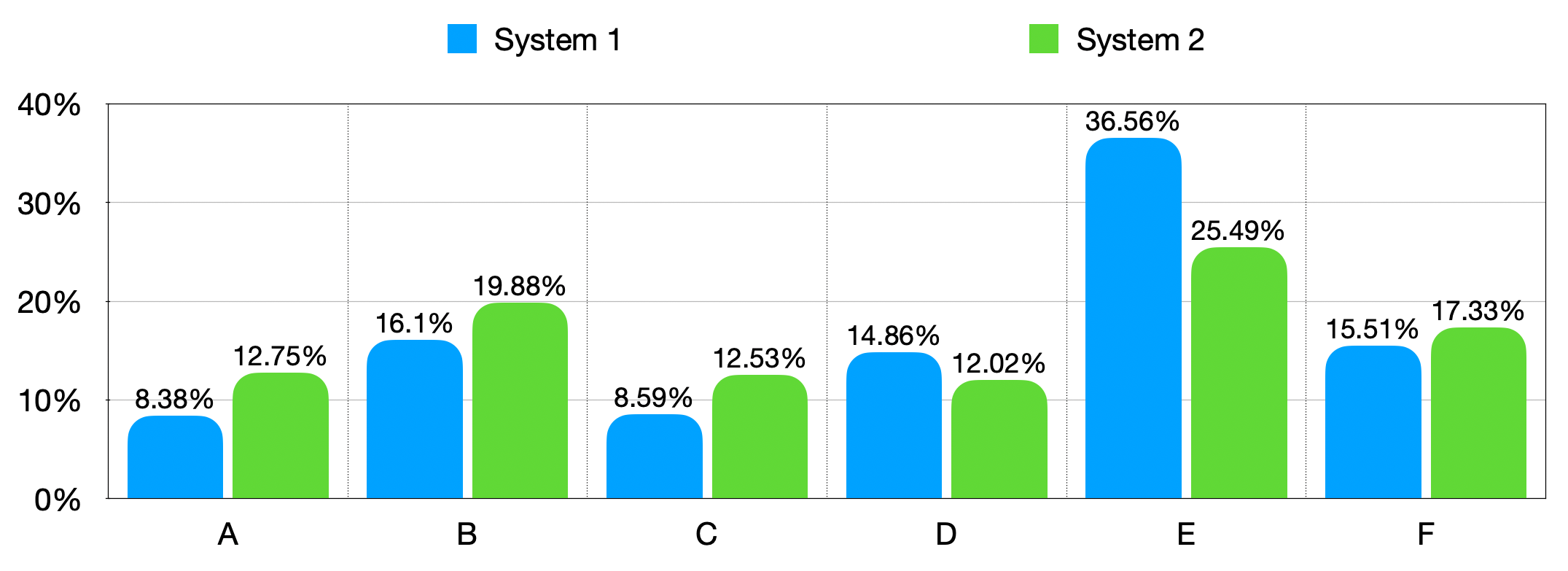}
            \caption{Label percentage distribution histogram}
            \label{fig:label_percentage_distribution}
        \end{figure}
    
        We gathered 45 items across all item types (10 StereoNLI, 7 Fallacy, 8 Syllogism/Stereo-Syllogism, 10 Guilt, and 10 Primacy/Recency), each annotated 30 times by humans during the first phase (System 1) and an additional 30 times by the same individuals in the second phase (System 2). Figure \ref{fig:label_percentage_distribution} provides a quantitative comparison of voting preferences between System 1 and System 2, highlighting how the distribution of votes across six options changed from System 1 to System 2. Notably, option E exhibited the most significant shift, decreasing by $11.07$ percentage points.

        \begin{figure}
            \centering
            \includegraphics[width=\textwidth]{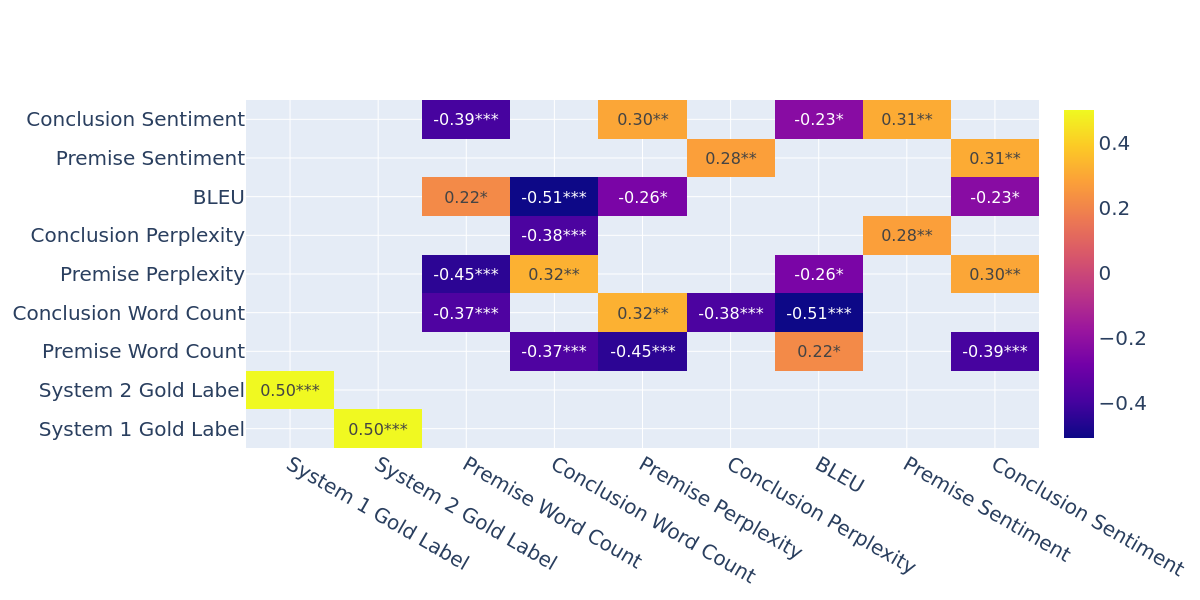}
            \caption{Kendall's $\tau$ between various features of the items in our dataset. Only statistically significant (p-value $<0.05$) $\tau$ values are shown. \textit{***} indicates a p-value $<0.001$, \textit{**} indicates a p-value $<0.01$, and \textit{*} indicates a p-value $<0.05$.}
            \label{fig:tau}
        \end{figure}
        
        To ensure the reliability of our results in subsequent experiments, we examined potential spurious correlations within our dataset, focusing on four primary factors: the length of the premise and conclusion, the perplexity of the premise and conclusion, BLEU scores \citep{papineni_bleu_2001} between the premise and conclusion, and VADER sentiment scores \citep{hutto_vader_2014} for both the premise and conclusion. Figure \ref{fig:tau} displays the Kendall's $\tau$ \citep{kendall_new_1938} between these factors and the mean value of participant responses. Ideally, no feature should correlate with the gold labels from either System 1 or System 2, as such a correlation could be considered spurious. The highest $\tau$ value is observed between System 1 and System 2 gold labels, indicating that a participant’s responses in System 1 may be predictive of their responses in System 2. However, in this paper, we treat System 1 and System 2 independently and do not attempt to predict System 2 responses based on System 1, leaving this for potential future work.
    
        \begin{figure}[t]
            \centering
            \begin{subfigure}{0.2\textwidth}
                \centering
                \includegraphics[width=\textwidth]{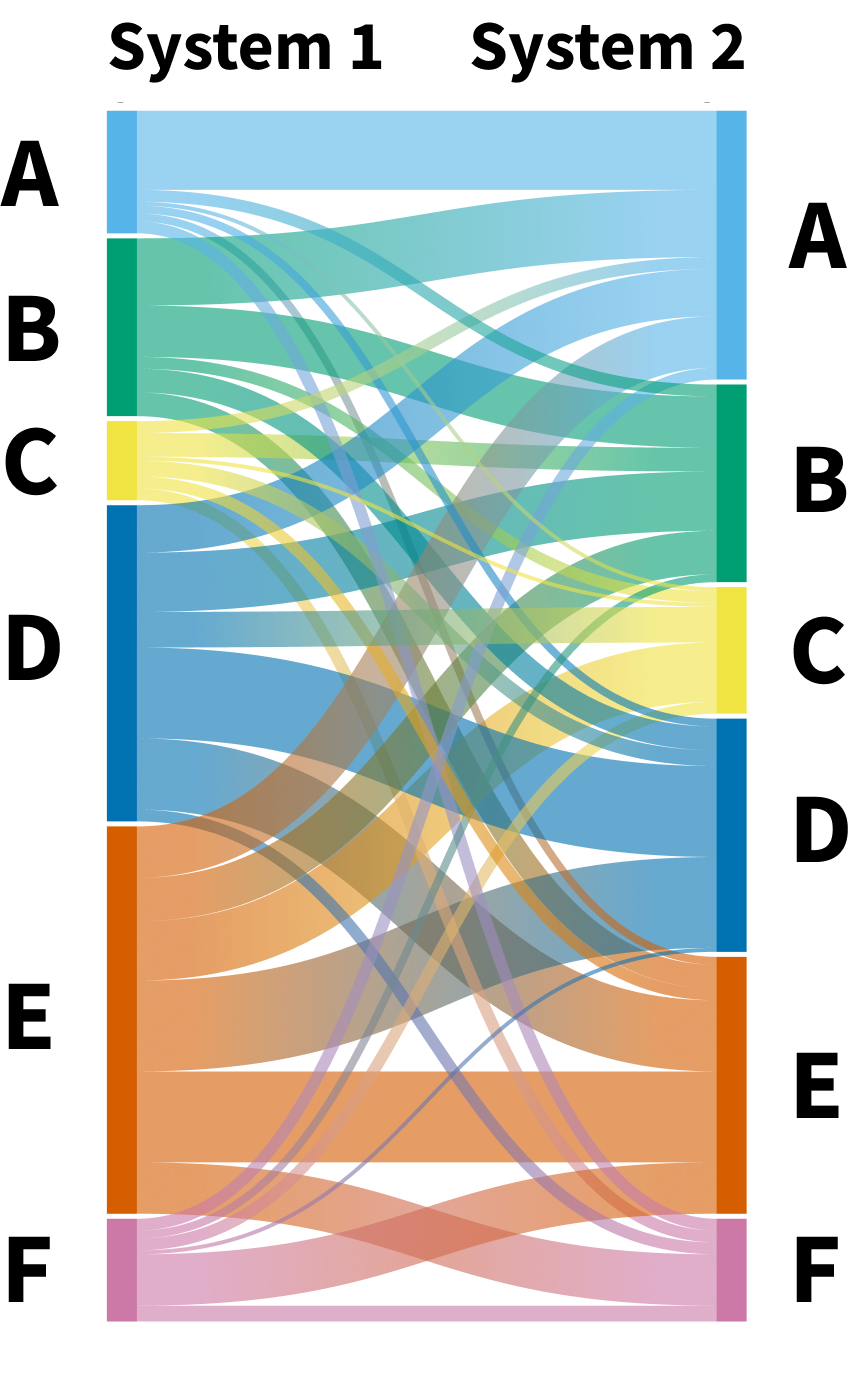}
                \caption{StereoNLI}
                \label{fig:sankey-stereonli}
            \end{subfigure}
            \hfill
            \begin{subfigure}{0.2\textwidth}
                \centering
                \includegraphics[width=\textwidth]{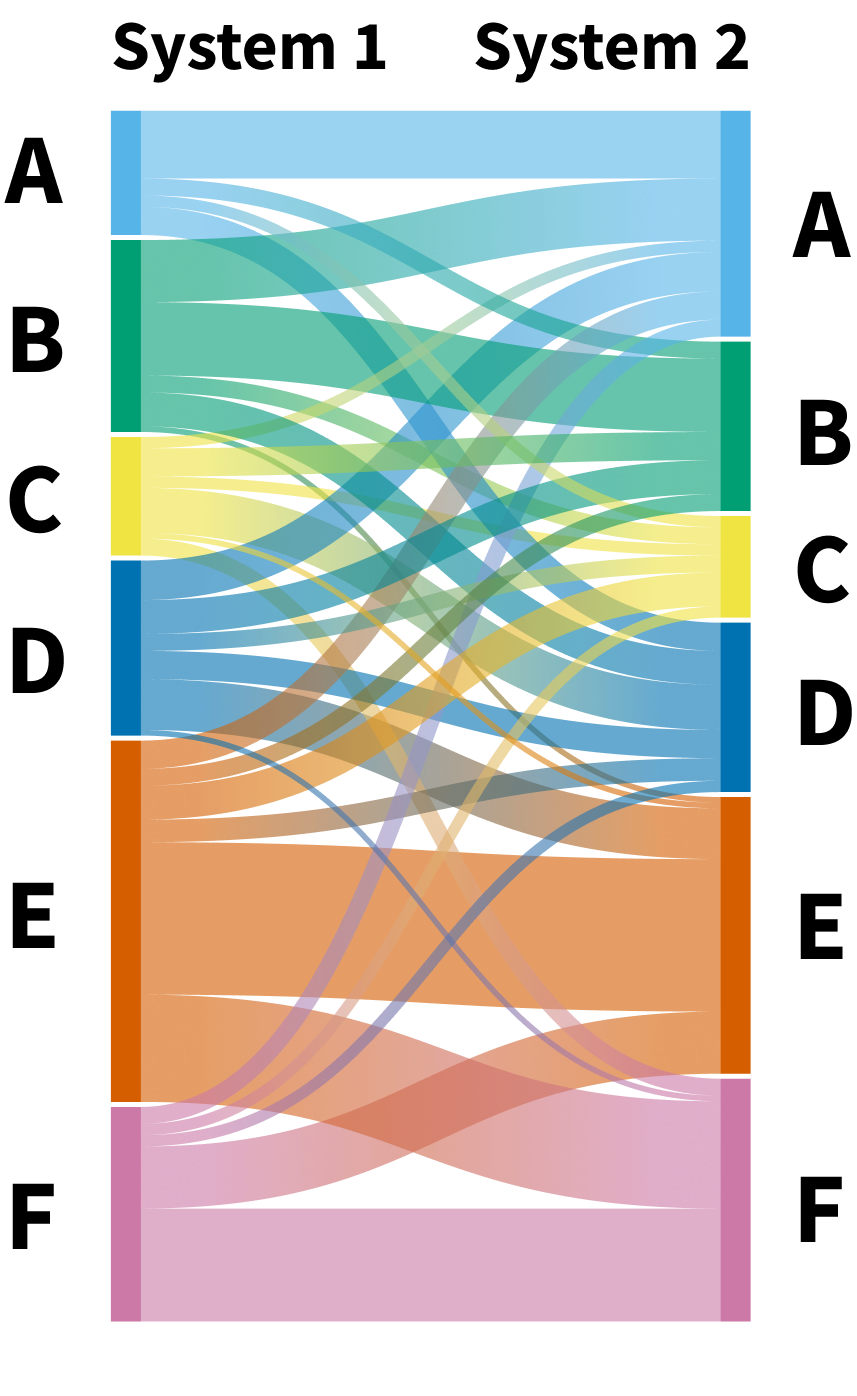}
                \caption{Fallacy}
                \label{fig:sankey-fallacy}
            \end{subfigure}
            \hfill
            \begin{subfigure}{0.2\textwidth}
                \centering
                \includegraphics[width=\textwidth]{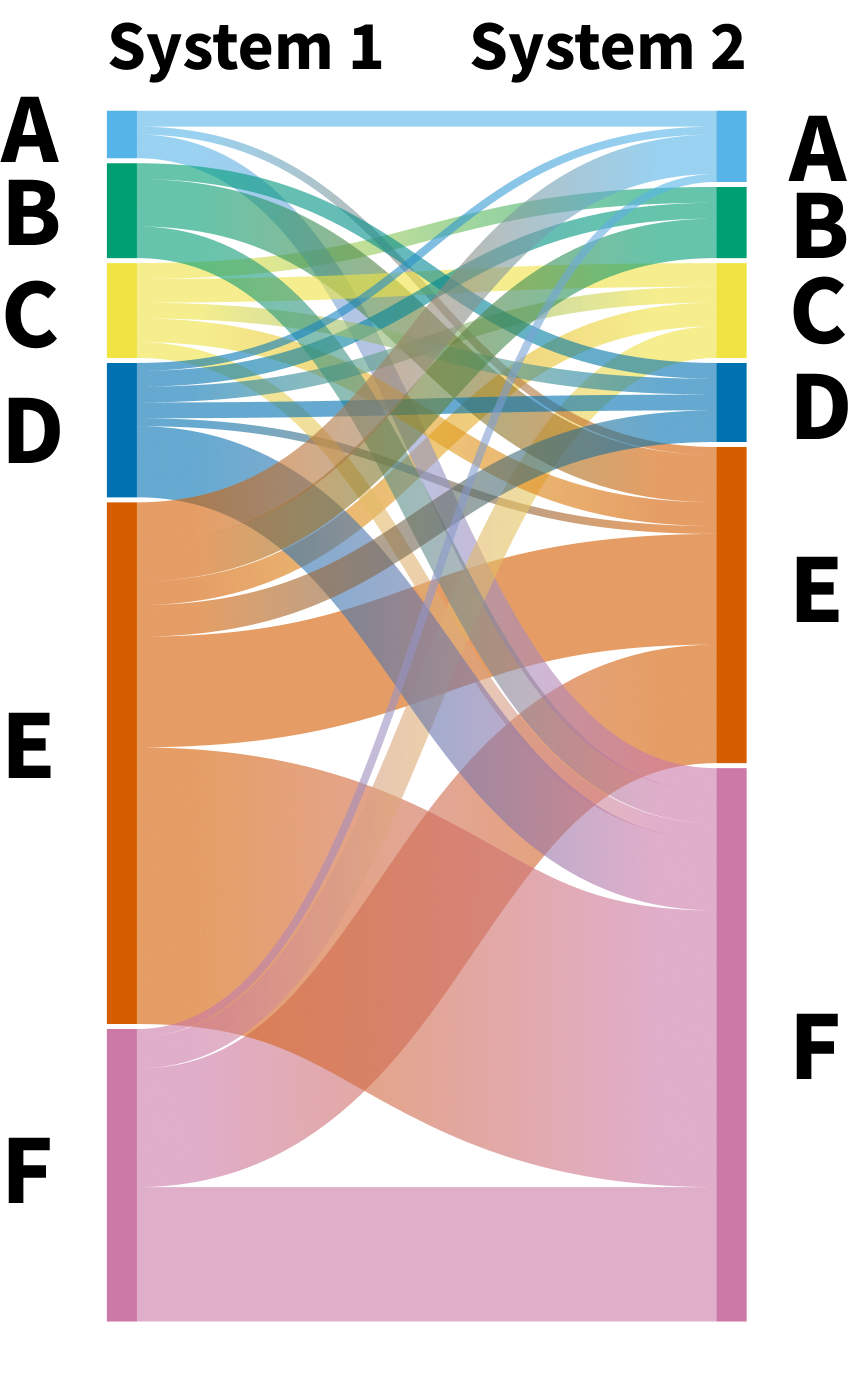}
                \caption{Syllogism}
                \label{fig:sankey-syllogism}
            \end{subfigure}
            \hfill
            \begin{subfigure}{0.2\textwidth}
                \centering
                \includegraphics[width=\textwidth]{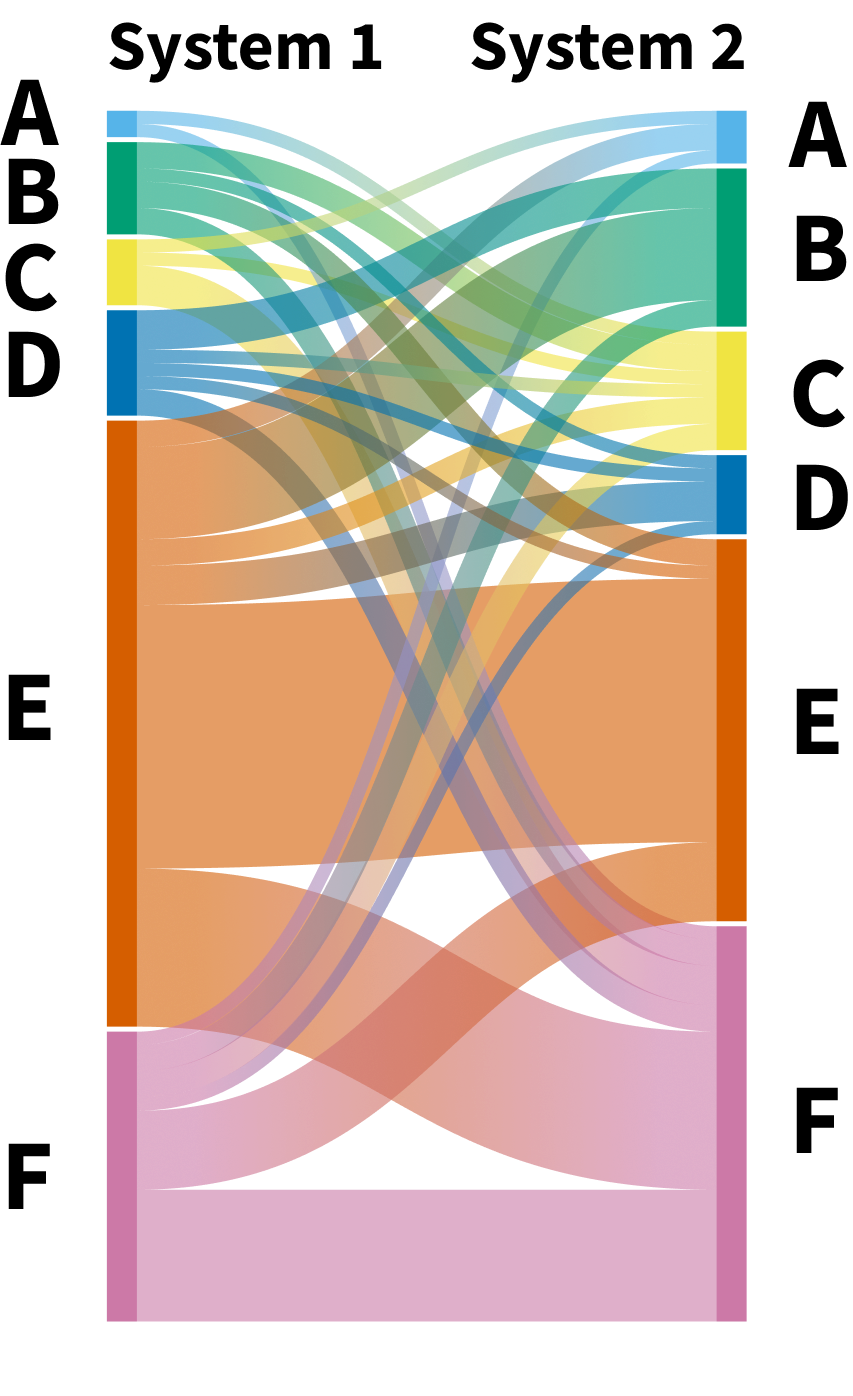}
                \caption{S. Syllogism}
                \label{fig:sankey-stereo-syllogism}
            \end{subfigure}
            \hfill
            \begin{subfigure}{0.2\textwidth}
                \centering
                \includegraphics[width=\textwidth]{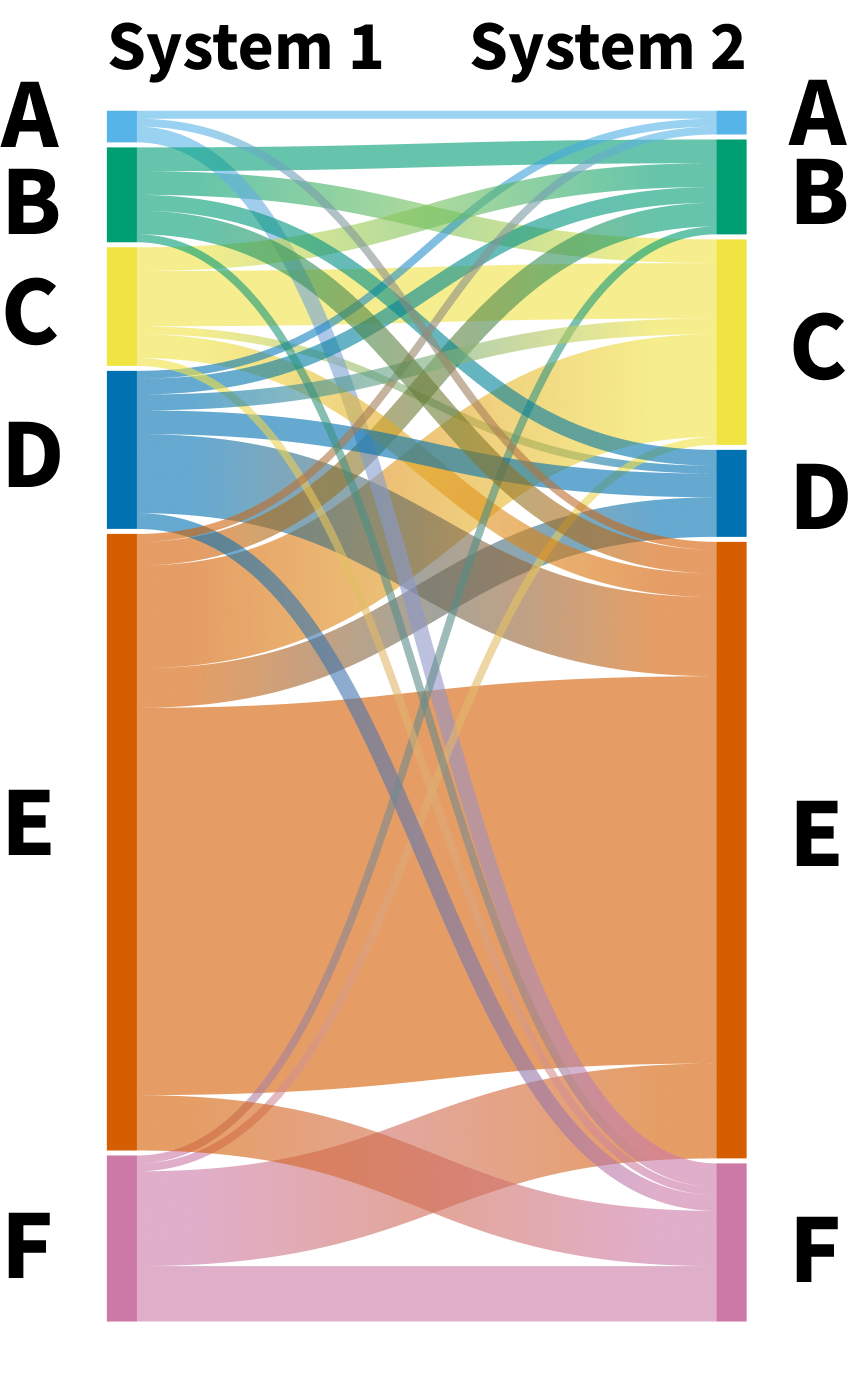}
                \caption{E Guilt}
                \label{fig:sankey-e-guilt}
            \end{subfigure}
            \hfill
            \begin{subfigure}{0.2\textwidth}
                \centering
                \includegraphics[width=\textwidth]{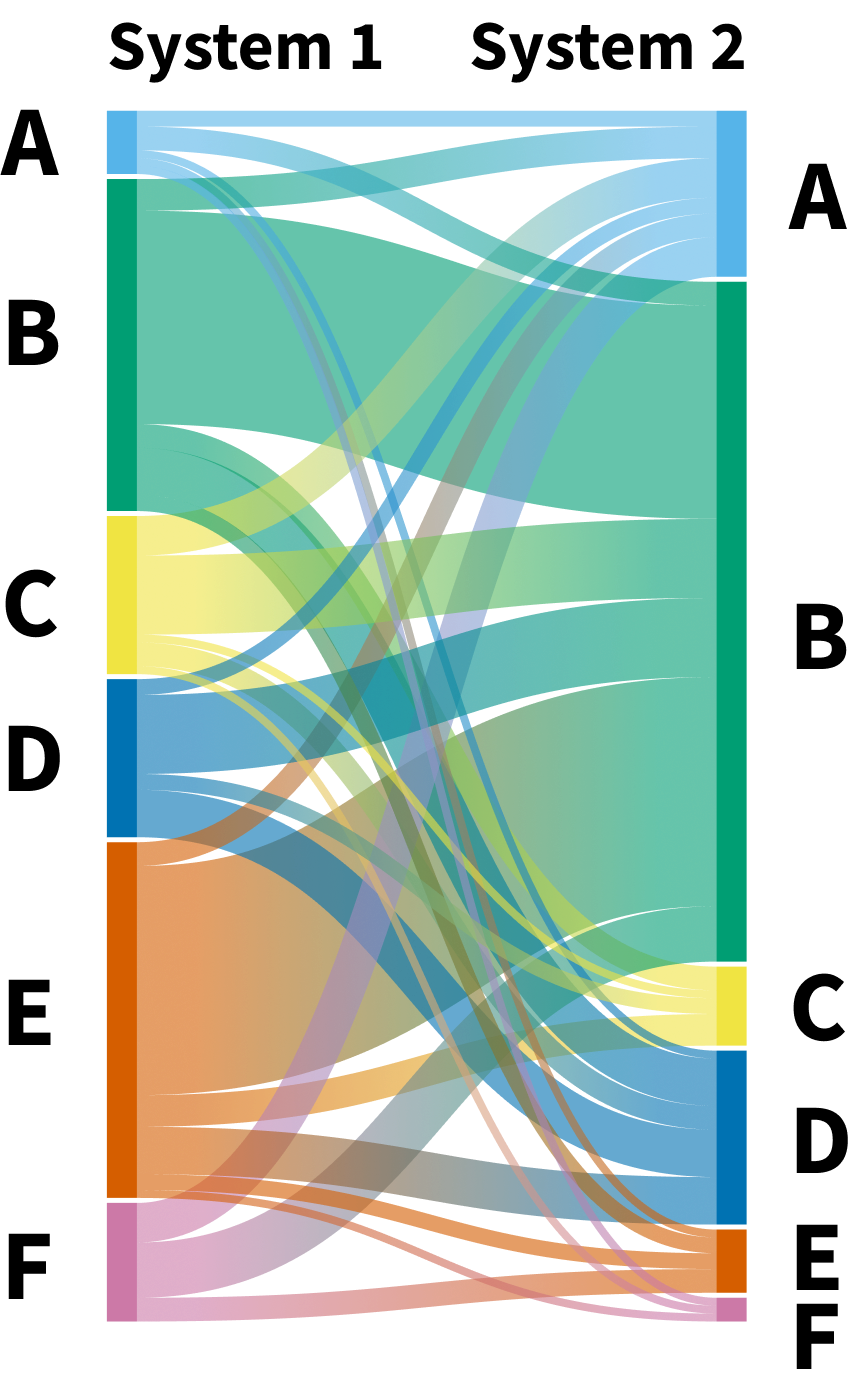}
                \caption{C Guilt}
                \label{fig:sankey-c-guilt}
            \end{subfigure}
            \hfill
            \begin{subfigure}{0.2\textwidth}
                \centering
                \includegraphics[width=\textwidth]{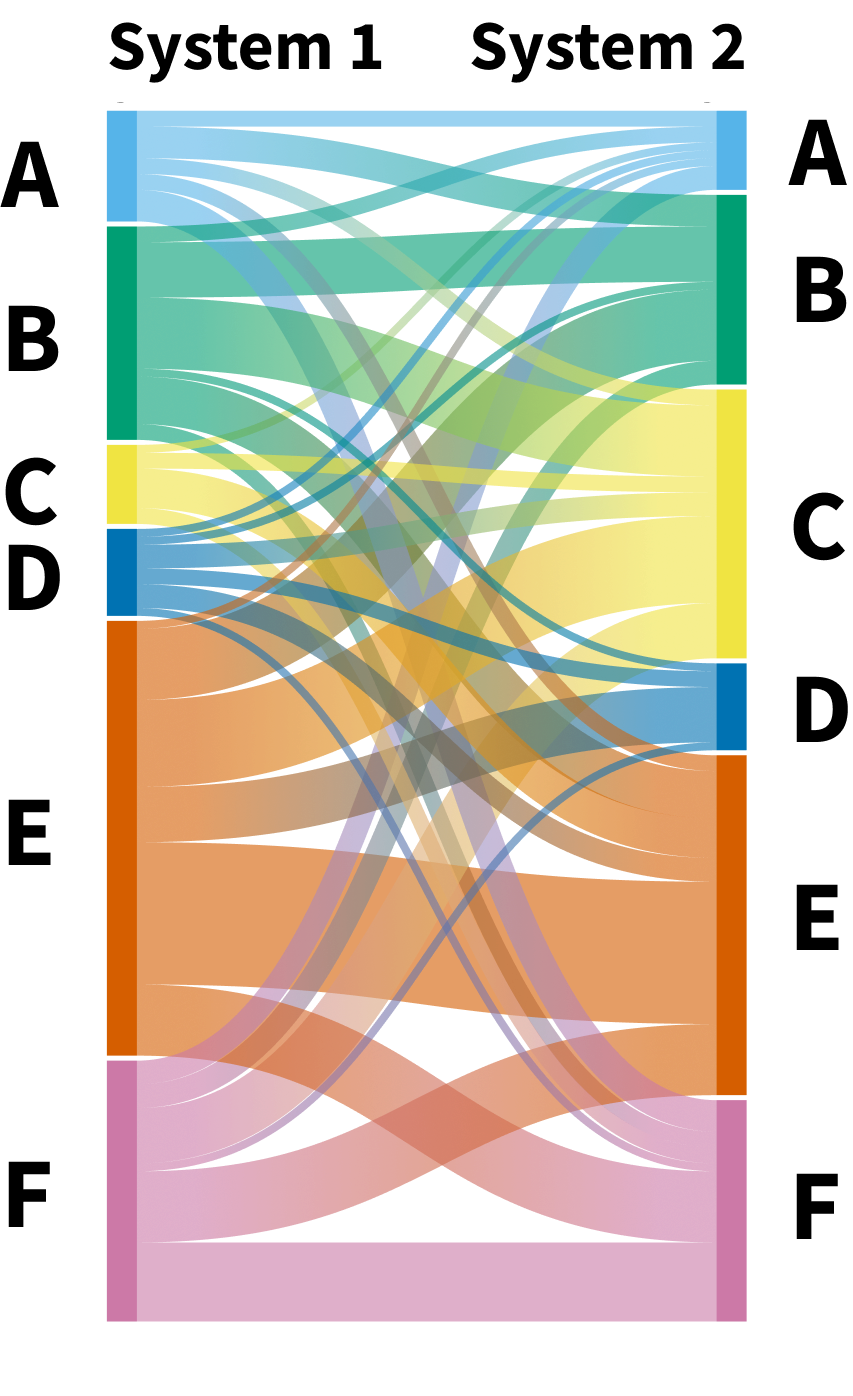}
                \caption{P Primacy}
                \label{fig:sankey-p-primacy}
            \end{subfigure}
            \hfill
            \begin{subfigure}{0.2\textwidth}
                \centering
                \includegraphics[width=\textwidth]{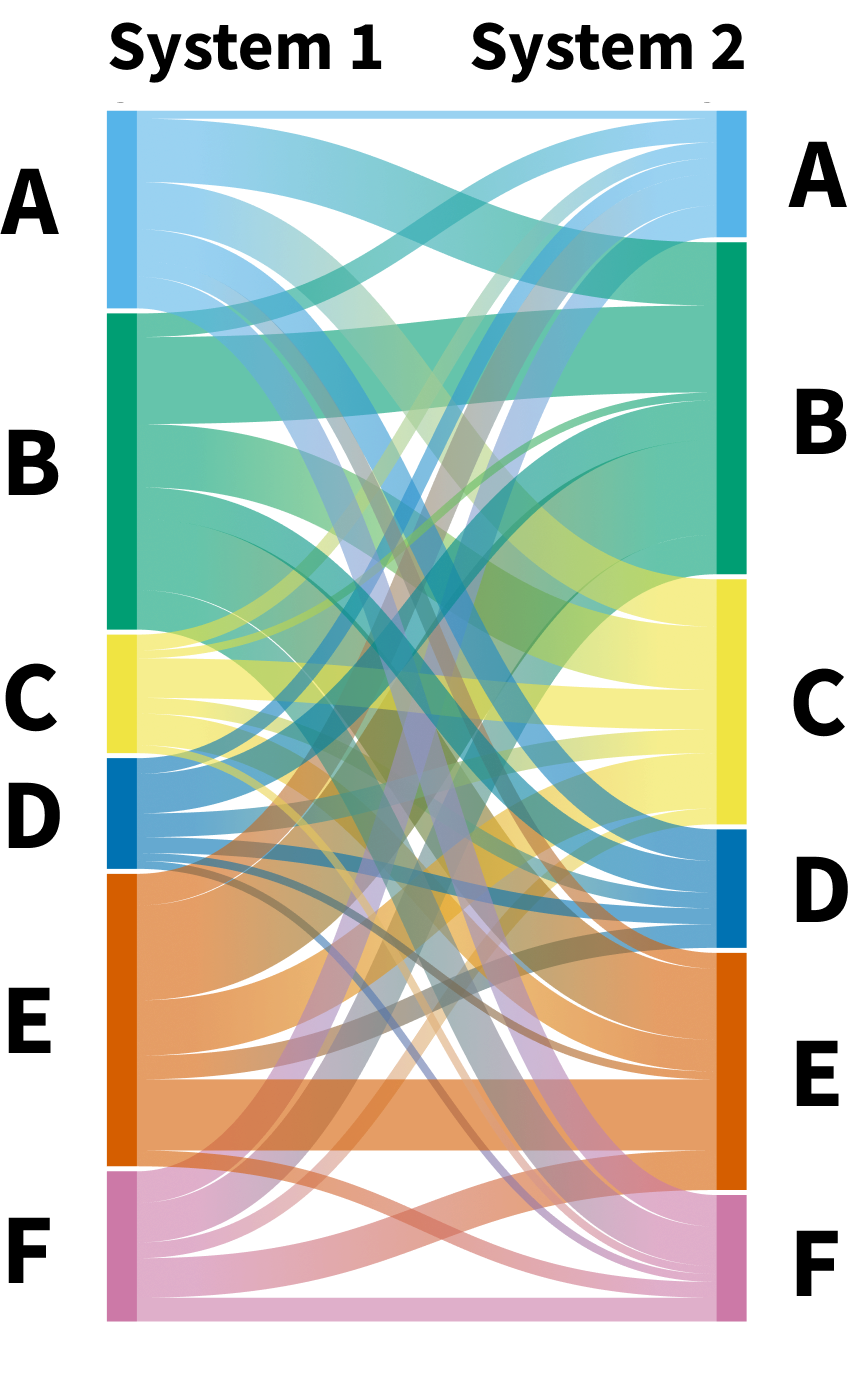}
                \caption{N Primacy}
                \label{fig:sankey-n-primacy}
            \end{subfigure}
           \caption{Sankey diagram of changes in human responses from System 1 (phase 1 of the survey) to System 2 (phase 2 of the survey) for all item types in our dataset.}
           \label{fig:sankey}
        \end{figure}

        Figure \ref{fig:sankey} presents Sankey diagrams to visually represent the transition of human responses from System 1 to System 2, illustrating the distribution and shifts across our six labels by varying the width of the links. This visualization allows us to easily see how successful each question type is at highlighting the differences between System 1 and System 2 reasoning are, as well as how the distribution of human responses differs with each question type. We observe that most participant responses fluctuated between E and D for StereoNLI, and between E and F for fallacy, syllogisms, and stereo syllogisms. Conversely, E guilt questions exhibited minimal fluctuation, with most participants maintaining their E response from System 1 to System 2. This consistency is expected, as is the trend of shifting from E to B for C Guilt, since these questions are designed to mislead participants in System 1. This misleading effect is also observed, though to a lesser extent, in P Primacy and N Primacy. Another noteworthy trend is that participants generally avoided selecting the neutral options C and D for any question type, suggesting an underlying bias. Further research may shed light on the reasons for this behavior.

        \paragraph{Inclusion and Ethics Statement} Our study was designed and conducted with careful consideration for ethical research practices and inclusivity. We recruited 60 participants via Prolific, aiming for a diverse sample. All participants provided informed consent and were clearly instructed about the study's nature. Data privacy and protection measures were implemented to ensure participant anonymity and confidentiality. We incorporated breaks to reduce fatigue and set time limits to manage cognitive load, considering participant well-being. The study design included various item types to prevent monotonicity and reduce potential biases. We used AI models (GPT-4) to generate survey items, addressing potential data contamination issues and creating a novel dataset. We have been transparent about our methodology, including the use of AI-generated content and both proprietary and open-source language models in our experiments. By adhering to these ethical principles and inclusive practices, we aim to contribute to the field of AI and cognitive science research in a responsible and equitable manner.

%% file: latex/experiments.tex
\section{Modeling Experiments}\label{sec:experiments}

    The overarching aim of our modeling experiments is to predict how human votes are distributed across each item. We begin by predicting the mode of the distribution (majority vote, as discussed in Section \ref{sec:accuracy}), then proceed to estimate the variance among the votes (Section \ref{sec:var}), and ultimately aim to predict the entire distribution of human votes (Section \ref{sec:mimic}). None of these tasks can be considered inherently easier than the others due to the significant variability in human reasoning, as highlighted in Section \ref{subsec:dataset_characteristics}. Given our development of the 6-way scheme, which includes six labels as detailed in Section \ref{sec:data_collection}, we also compare the performance of LLMs on the more conventional NLI scheme: the 3-way classification (contradiction, neutral, and entailment). It is important to note that we collect human responses solely for the 6-way scheme and subsequently map these responses to the 3-way scheme by categorizing A and B as \textit{contradiction}, C and D as \textit{neutral}, and E and F as \textit{entailment}. All LLMs are tested separately on both the 3-way and 6-way schemes. We utilize both proprietary LLMs (GPT-3.5,\footnote{\url{https://platform.openai.com/docs/models/gpt-3-5-turbo}} GPT-4 \citep{openai_gpt-4_2023}, GPT-4o-mini\footnote{\url{https://openai.com/index/gpt-4o-mini-advancing-cost-efficient-intelligence/}}) and open-source LLMs (Gemma 2 27b \citep{gemma_2024}, Llama 3.1 8b, Llama 3.1 70b, Llama 3.1 405b,\footnote{\url{https://github.com/meta-llama/llama-models/blob/main/models/llama3_1/MODEL_CARD.md}} Mistral 7b with direct performance optimization \citep{rafailov_direct_2023, Nous-Hermes-2-Mistral-7B-DPO}) to compare their performance against traditional (non-transformer-based) machine learning algorithms.

    \begin{table}
        \centering
        \begin{tabular}{ll}
            \toprule
            Trait & Prompt\\
            \midrule
             O+ & You're open to new experiences, creative, inventive, curious, and imaginative.\\
             O-- & You prefer routine and familiarity, consistent, conventional, and cautious.\\
             C+ & You're organized, efficient, reliable, and responsible.\\
             C-- & You're flexible, spontaneous, extravagant, and careless.\\
             E+ & You're friendly, outgoing, sociable, and energetic.\\
             E-- & You're reserved, quiet, introverted, and solitary.\\
             A+ & You're cooperative, warm, friendly, and compassionate.\\
             A-- & You're competitive, detached, critical, and judgemental.\\
             N+ & You're anxious, stressed, nervous, and emotionally sensitive.\\
             N-- & You're calm, stable, confident, and emotionally resilient.\\
             \bottomrule
        \end{tabular}
        \caption{Personality Prompts}
        \label{tab:prompts}
    \end{table}
    
    To predict the entire vote distribution using LLMs, we must prompt them in a way that allows the recreation of a vote distribution rather than producing a single vote per item. We do not fine-tune the LLMs because we do not have enough data to tune millions of parameters that we would need to tune even with techniques such as parameter-efficient fine-tuning \citep{ding_parameter-efficient_2023}. All the LLMs we use are chatbot models, which include a system prompt that influences the LLM's responses and a user prompt (both described below) that contains the specific query. We maintain consistency in the user prompt by limiting it to just the premise and hypothesis (we call these \textit{s1} and \textit{s2} in our prompts, not to be confused with System 1 and System 2, which are never shortened in this paper). We develop two distinct task definitions, one for each labeling scheme, and employ two different prompting styles. The task definitions are as follows:
    \begin{itemize}
        \item \textbf{6-way:}\\
            Assuming s1 is true, choose the statement that seems most accurate for s2:
            \begin{enumerate}[label=\Alph*.]
                \item Absolutely must be false
                \item Is more likely to be false
                \item Has strong reasons to be true and strong reasons to be false
                \item Has no reasons to be either true or false
                \item Is more likely to be true
                \item Absolutely must be true
            \end{enumerate}
        \item \textbf{3-way:}\\
            Choose one option about the inferential relationship between s1 and s2:\\
            Entailment: s2 entails s1\\
            Contradiction: s2 contradicts s1\\
            Neutral: Cannot pick either of the above or both are likely
    \end{itemize}
    The first prompting style, called \textit{base prompting}, simply explains the task and asks the LLM to provide its prediction. The exact base prompt is: \texttt{\{task definition\}} Pick exactly one option and write it on the first line. Do not write anything else.'' The second prompting style, \textit{personality prompting}, adds a brief description of the desired personality for the LLM to adopt, in addition to the base prompt. The personality prompt reads: ``Here’s your personality: \texttt{{personality}}. Focus on this personality and respond just like a person who has this personality. \texttt{\{task definition\}} Pick the first answer that you think of based on your personality and nothing else. Pick exactly one option and write it on the first line. Do not write anything else.'' The different personalities are derived from the Big Five personality traits (OCEAN model) \citep{roccas_big_2002}, with each trait simplified to a high (+) or low (--) value to create ten different personality prompts shown in Table \ref{tab:prompts}.

    For our experiments, we use the same set of responses generated by all LLMs. For base prompting, we obtain this set by prompting the LLM ten times with a temperature setting of 1 (to maximize response entropy) to generate a distribution with ten votes. For personality prompting, we use each of the ten personality prompts to generate ten votes, then select the majority vote from each prompt's votes to compile a set of ten votes. This approach yields ten votes each for the base and personality prompting styles. It is important to note that the LLMs are not fine-tuned for this task, in contrast to the classical machine learning algorithms, which are fine-tuned. This intentional disparity aims to test the LLMs' innate ability to replicate human responses.
    
    \subsection{Are human responses predictable?}\label{sec:accuracy}
        
        \begin{figure}[t]
             \centering
             \begin{subfigure}{0.49\textwidth}
                 \centering
                 \includegraphics[width=\textwidth]{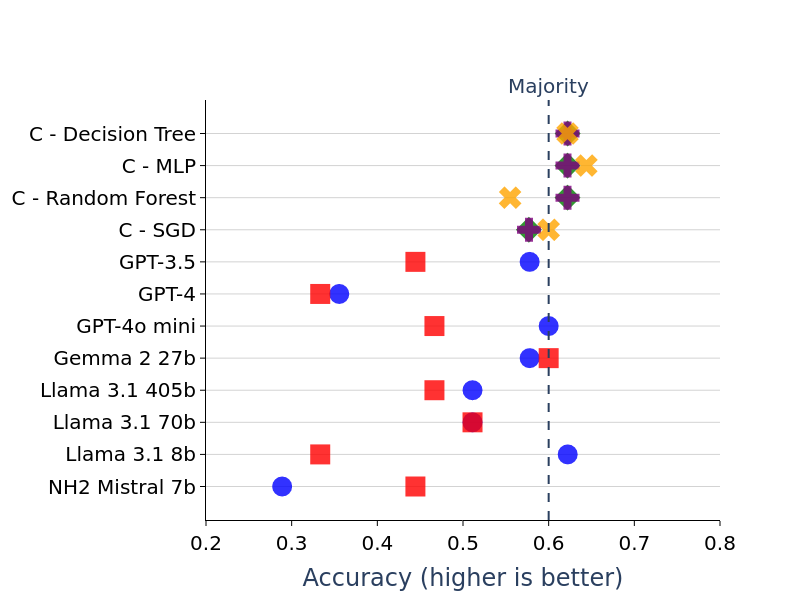}
                 \caption{6-way System 1}
                 \label{fig:6way_S1_accuracy}
             \end{subfigure}
             \begin{subfigure}{0.49\textwidth}
                 \centering
                 \includegraphics[width=\textwidth]{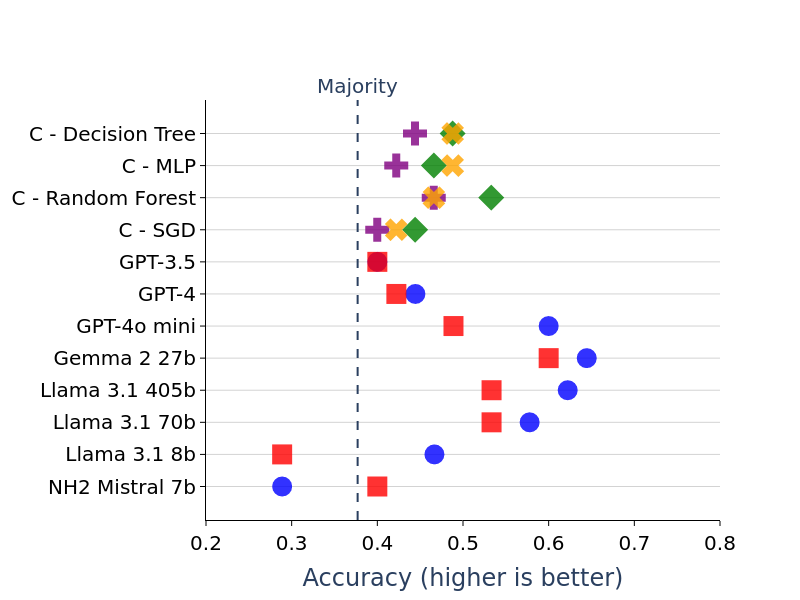}
                 \caption{6-way System 2}
                 \label{fig:6way_S2_accuracy}
             \end{subfigure}
             \begin{subfigure}{0.49\textwidth}
                 \centering
                 \includegraphics[width=\textwidth]{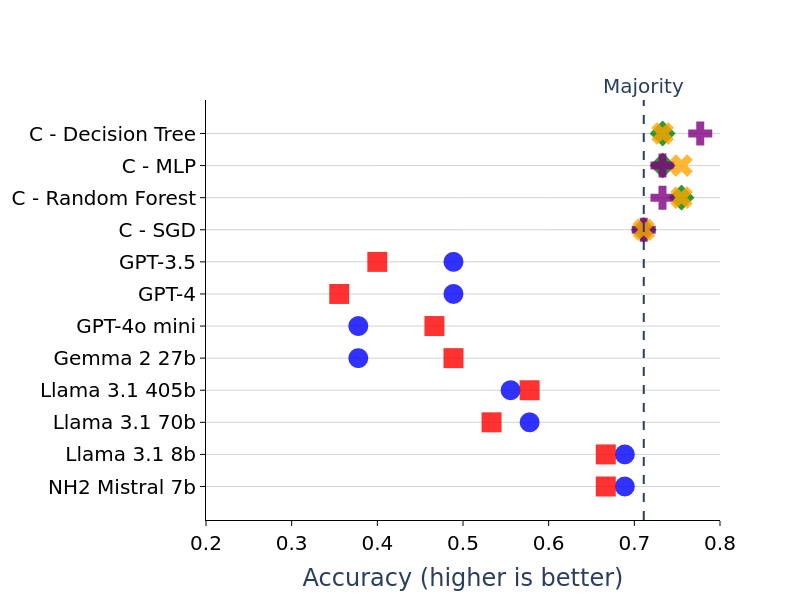}
                 \caption{3-way System 1}
                 \label{fig:enc_S1_accuracy}
             \end{subfigure}
             \begin{subfigure}{0.49\textwidth}
                 \centering
                 \includegraphics[width=\textwidth]{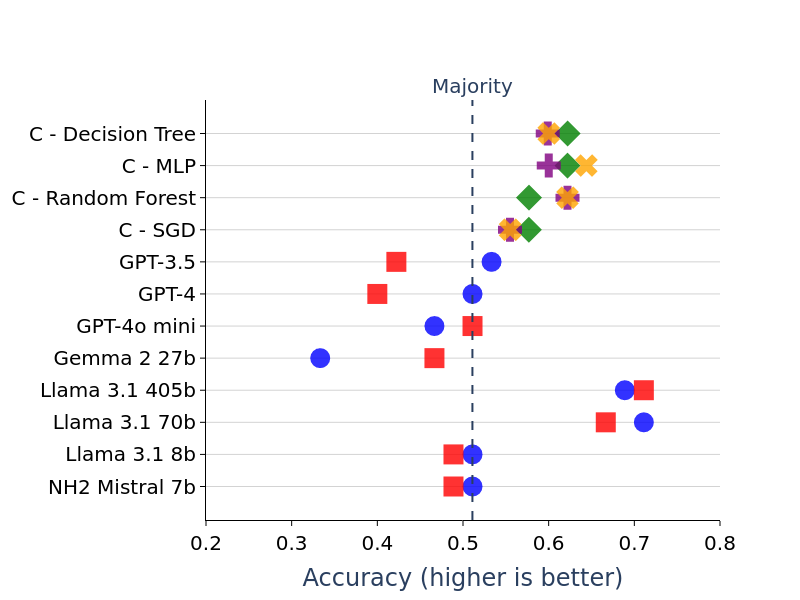}
                 \caption{3-way System 2}
                 \label{fig:enc_S2_accuracy}
             \end{subfigure}
                \caption{Human gold label prediction accuracy of classical machine learning models using various input schemes: base features only (green diamond \textcolor{Green}{\ding{117}}), base features with count vectorization of premise and conclusion (purple cross \textcolor{Plum}{\ding{58}}), and base features with TFIDF encoding of premise and conclusion (orange X \textcolor{YellowOrange}{\ding{54}}). LLMs are also included with base prompting (blue circle \textcolor{blue}{\ding{108}}) and personality prompting with equal weights (red square \textcolor{Red}{\ding{110}}). Dashed line represents the majority baseline for the classification task.}
                \label{fig:accuracy}
        \end{figure}
        
        The task of predicting the mode (also referred to as the gold label) of human responses can be approached as a classification problem, since all categories are different from one another. While it can also be framed as a regression problem, this approach presents several challenges. First, NLI has classically been modeled as a classification task, and efforts to convert it to a regression problem \citep{chen_uncertain_2020} have faced significant difficulties \citep{nighojkar_no_2023}. Second, although our 6-way scale resembles a Likert scale, neither our 3-way nor our 6-way scale can be directly interpreted as a regression task because there is no intermediate value between 1 and 2 (or A and B) on our 6-way scale. The debate on whether Likert-like scales should be treated as regression is discussed in \citet{sullivan_analyzing_2013}.

        To process human responses for the 6-way classification scheme, we convert labels ranging from A to F into a categorical format corresponding to numbers 1 through 6. Similarly, for the 3-way scheme, the labels contradiction, neutral, and entailment are mapped to 1, 2, and 3, respectively. We use four classical machine learning algorithms: Stochastic Gradient Descent (SGD) \citep{JMLR:v15:gupta14a}, Decision Tree \citep{quinlan_induction_1986}, Random Forest \citep{tin_kam_ho_random_1998}, and a Feed-forward Neural Network \citep{schmidhuber_deep_2015}. Various statistics and features that might influence the predictability of responses are added to the dataset, including (1) the word count of premises and conclusions, (2) the perplexity of premises and conclusions, (3) the BLEU score between premise and conclusion, and (4) sentiment analysis of premises and conclusions using the VADER tool. Collectively, these are referred to as ``base input features.''
        
        Given that these algorithms cannot natively process raw text data, we tokenize the text to remove stopwords and punctuation, then encode the premises and conclusions using Bag-of-Words and TF-IDF \citep{robertson_understanding_2004}, converting the text into numerical representations. The choice of encoding is crucial, as it impacts the model's ability to capture linguistic nuances. Due to the limited data available, which restricts the size of the test set, we employ grid search to optimize these algorithms. The models are trained on a simple classification task, where they predict the gold label for human responses based on the input features, separately for both System 1 and System 2 responses. The performance of the best hyperparameter combinations, averaged across 5-fold cross-validation, is presented in Figure \ref{fig:accuracy}. For LLMs, we rely on the majority vote from ten base prompts and the majority vote from ten personality prompts for each item.
        
        This task primarily aims to predict the population-level gold label, and while it may oversimplify individual reasoning differences, addressing whether human responses are predictable is a critical step before conducting more sophisticated analyses. As anticipated, the accuracy for the 3-way scheme (Figures \ref{fig:enc_S1_accuracy} and \ref{fig:enc_S2_accuracy}) surpasses that of the 6-way scheme (Figures \ref{fig:6way_S1_accuracy} and \ref{fig:6way_S2_accuracy}). The majority vote baseline, which simply predicts the most common gold label in the dataset, achieves higher accuracy on System 1 for both schemes; however, classical ML algorithms only slightly outperform this baseline. The baseline accuracy significantly declines for System 2 in both the 6-way and 3-way schemes, suggesting greater diversity in human responses for System 2 compared to System 1. Despite this decline, the classical ML algorithms maintain performance levels close to those observed in System 1, indicating a consistent degree of predictability in both System 1 and System 2 responses, with predictability being higher in System 1. Notably, Figure \ref{fig:accuracy} reveals that LLMs perform nearly equally well in predicting gold labels for System 1 and System 2 responses, and they significantly surpass the baseline in System 2, emerging as the most effective at predicting the label most agreed upon by humans using System 2 reasoning. The fact that LLMs perform similarly on the 6-way scheme as they do on the 3-way scheme is also intriguing. Despite fewer choices in the 3-way scheme, which theoretically should make it easier, the LLMs' performance suggests that the increased granularity of the 6-way scheme may render human responses more predictable for LLMs than initially expected.
    
    \subsection{Is the variance in human responses predictable?}\label{sec:var}
    
        \begin{figure}[t]
             \centering
             \begin{subfigure}{0.49\textwidth}
                 \centering
                 \includegraphics[width=\textwidth]{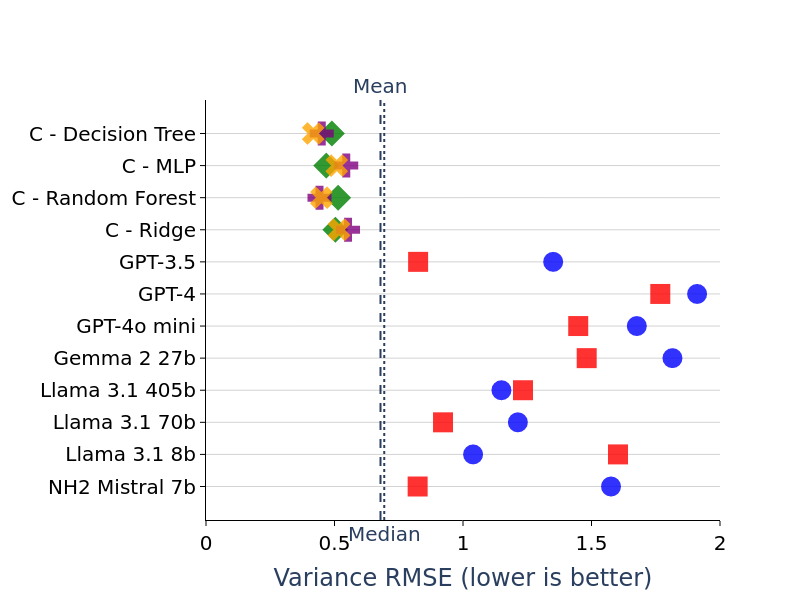}
                 \caption{6-way System 1}
                 \label{fig:6way_S1_rmse}
             \end{subfigure}
             \begin{subfigure}{0.49\textwidth}
                 \centering
                 \includegraphics[width=\textwidth]{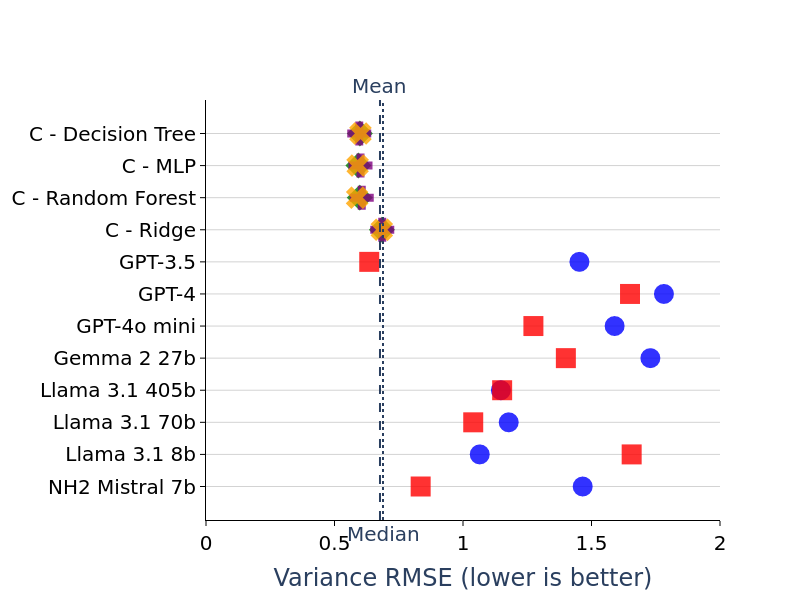}
                 \caption{6-way System 2}
                 \label{fig:6way_S2_rmse}
             \end{subfigure}
             \begin{subfigure}{0.49\textwidth}
                 \centering
                 \includegraphics[width=\textwidth]{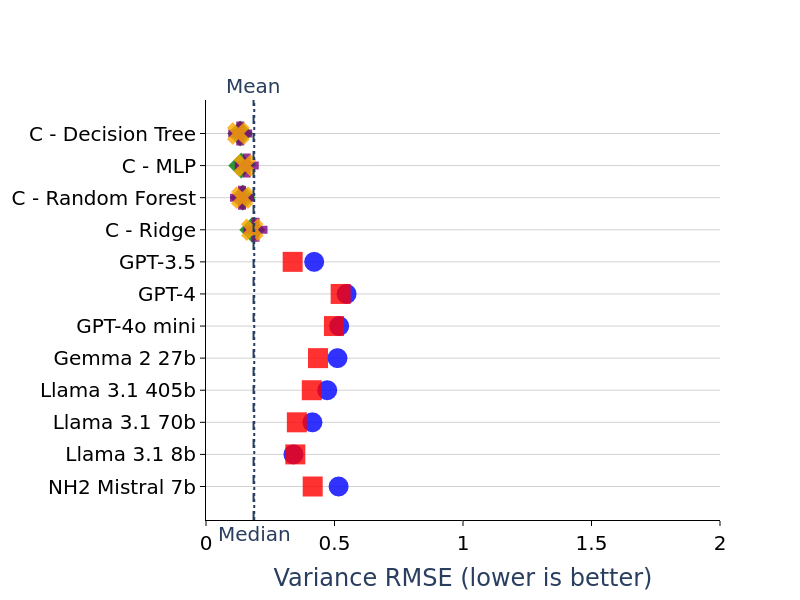}
                 \caption{3-way System 1}
                 \label{fig:enc_S1_rmse}
             \end{subfigure}
             \begin{subfigure}{0.49\textwidth}
                 \centering
                 \includegraphics[width=\textwidth]{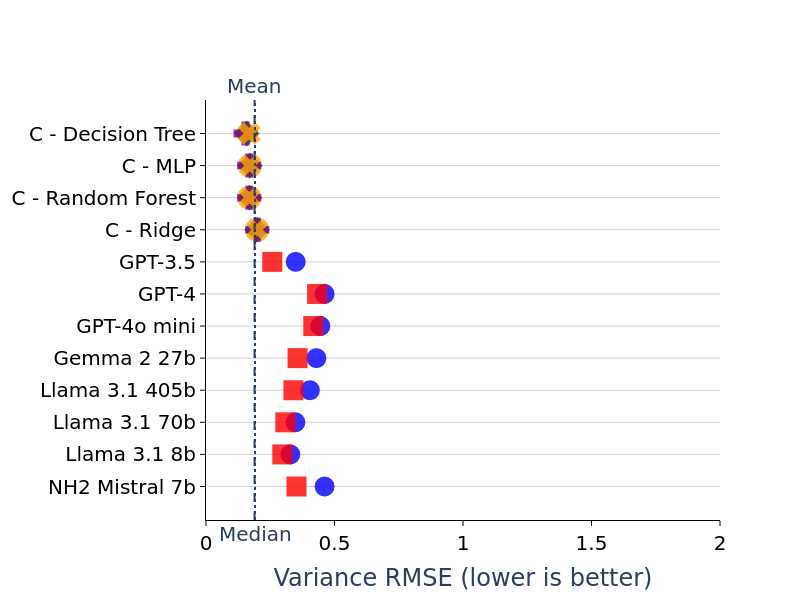}
                 \caption{3-way System 2}
                 \label{fig:enc_S2_rmse}
             \end{subfigure}
                \caption{RMSE for predicting variance between human responses using classical machine learning models with various input schemes: base features only (green diamond \textcolor{Green}{\ding{117}}), base features with count vectorization of premise and conclusion (purple cross \textcolor{Plum}{\ding{58}}), and base features with TFIDF encoding of premise and conclusion (orange X \textcolor{YellowOrange}{\ding{54}}). LLMs are also included with base prompting (blue circle \textcolor{blue}{\ding{108}}) and personality prompting with equal weights (red square \textcolor{Red}{\ding{110}}). Dashed line represents the mean baseline and dotted line represents the median baseline for the regression task.}
                \label{fig:var}
        \end{figure}
        
        After determining that we can predict the human gold label on both System 1 and System 2 with accuracy exceeding the majority baseline, we proceed to predict the variance among all human responses for a given item. This variance is calculated by mapping the human responses onto a scale from 1 to 6, as described in Section \ref{sec:accuracy}. It is important to note that, for this task, we treat the responses as ordinal rather than nominal, since the objective is to predict the variance between the human responses. Given that we are treating this as a regression task, we use root mean squared error (RMSE) as the evaluation metric, where a lower value indicates better performance \citep{armstrong_error_1992}. As in Section \ref{sec:accuracy}, we employ four different machine learning algorithms: ridge regression \citep{hoerl_ridge_2000}, decision tree, random forest, and a feed-forward neural network. We conduct a grid search to optimize hyperparameters and report the average results of a 5-fold cross-validation, as shown in Figure \ref{fig:var}. For LLMs, we calculate the variance from ten votes using base prompting and ten votes using personality prompting to compare with the human variance for each item. The baselines in this analysis predict either the mean or the median variance across all items.
        
        Since the 3-way scheme offers fewer options, the variance values fall within a narrower range, leading to lower prediction errors compared to the 6-way scheme, which is consistent with expectations. In both schemes, traditional ML algorithms outperform the baselines, although the margin of improvement is minimal. For both schemes, the variance between LLM votes significantly differs from the variance between human responses, with this disparity being much larger than what was observed in the accuracy of predicting the gold label (Figure \ref{fig:accuracy}).This indicates that predicting variance is a highly challenging task for LLMs, especially without any fine-tuning.
    
    \subsection{Can AI mimic the entire human response distribution?}\label{sec:mimic}
        
        Given the contrasting results obtained from our initial two experiments, further investigation is required to evaluate the ability of LLMs to emulate human reasoning. Additionally, while maxima and variance provide insight into a distribution, they represent only two of its characteristics. To more comprehensively evaluate the similarity between the response distributions of LLMs and humans, we now focus on comparing the entire response distribution. The Wasserstein Distance \citep{dobrushin_prescribing_1970}, also known as Earth Mover's Distance (EMD), serves as a metric for quantifying the difference between probability distributions across a specified metric space --- in this instance, the set of labels in a 3-way or 6-way classification scheme. Conceptually, if each distribution is visualized as a unit mass of earth, the EMD reflects the minimal \textit{cost} required to transform one distribution into the other, considering both the amount of earth that needs to be moved and the mean distance it must be moved. This makes EMD particularly suitable for our case, as it is sensitive to the ordinal nature of the metric.
        
        Since EMD requires a probability distribution, we transform each set of human responses into a vector of size $k$ (three or six for the 3-way or 6-way scheme), where each entry represents the frequency of the corresponding label. We then normalize this vector to create a probability distribution. Notably, EMD has an unbounded range, so we convert it into a similarity measure ranging from 0 to 1, which we term Earth Mover's Similarity (EMS). Our similarity function, defined as: \[\textit{EMS}(D_1, D_2) = 100^{\textit{EMD}(D_1, D_2)}\] takes the two normalized probability distributions $D_1$ and $D_2$ as inputs. While $e$ is often used as the base for exponents, we chose a base of $100$ in this study to better distinguish variations, as using $e$ would compress the values of interest into a narrow range.
        
        In our comparison, we previously evaluated LLMs against traditional ML algorithms trained on our dataset. However, these algorithms do not generate a distribution of votes. In this experiment, we assign weights to each of the ten personality prompts and use a genetic algorithm to identify the optimal set of weights for each of the five folds (as in Experiments \ref{sec:accuracy} and \ref{sec:var}). For each fold, we again conduct a grid search to find the best hyperparameters (parent selection type and crossover type) for training the genetic algorithm, while fixing the number of generations at 8, the population per generation at 256, and the number of mating parents at 128. These parameters were chosen to balance computational costs with performance, though we did not do a rigorous comparison of all possible values due to computational cost limitations. We refer to this setup as personality prompting with a genetic algorithm, or \texttt{P-GA}. Additionally, we test a setup using personality prompting with equal weights, termed \texttt{P-EQ}.
        
        \begin{figure}
            \centering
            \begin{subfigure}[b]{0.49\textwidth}
                \centering
                \includegraphics[width=\textwidth]{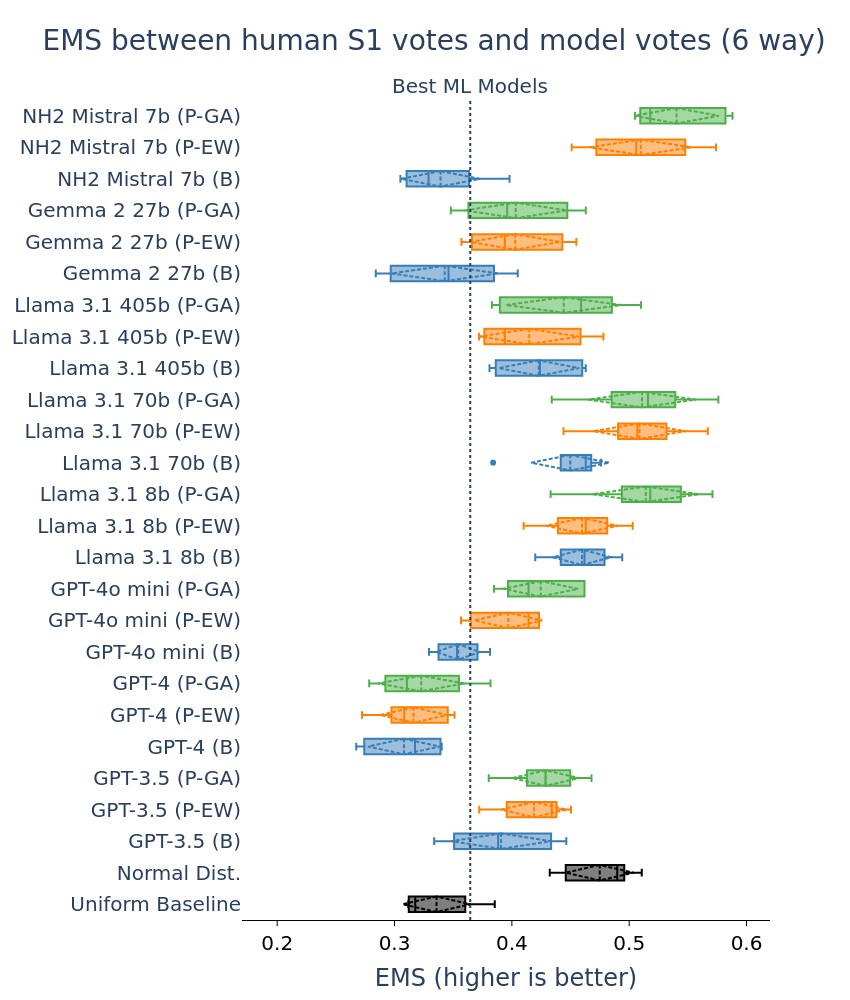}
                \caption{System 1 (6-way)}
                \label{fig:ems-s1-6}
            \end{subfigure}
            \begin{subfigure}[b]{0.49\textwidth}
                \centering
                \includegraphics[width=\textwidth]{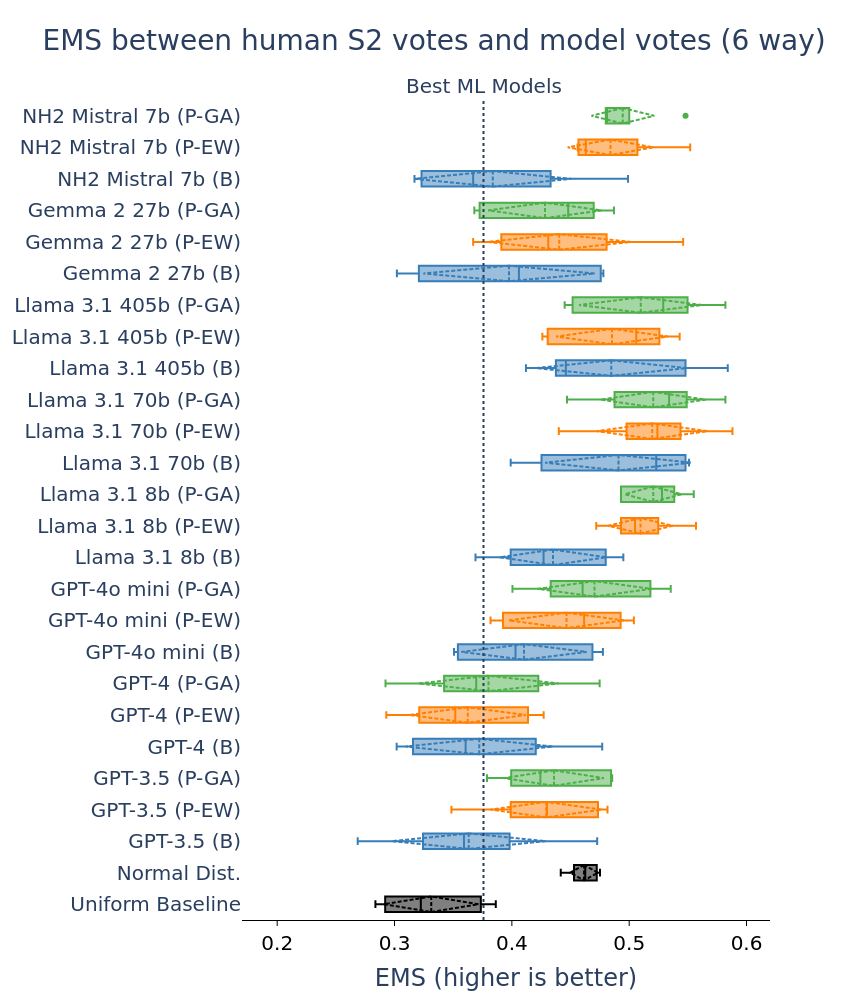}
                \caption{System 2 (6-way)}
                \label{fig:ems-s2-6}
            \end{subfigure}
            \begin{subfigure}[b]{0.49\textwidth}
                \centering
                \includegraphics[width=\textwidth]{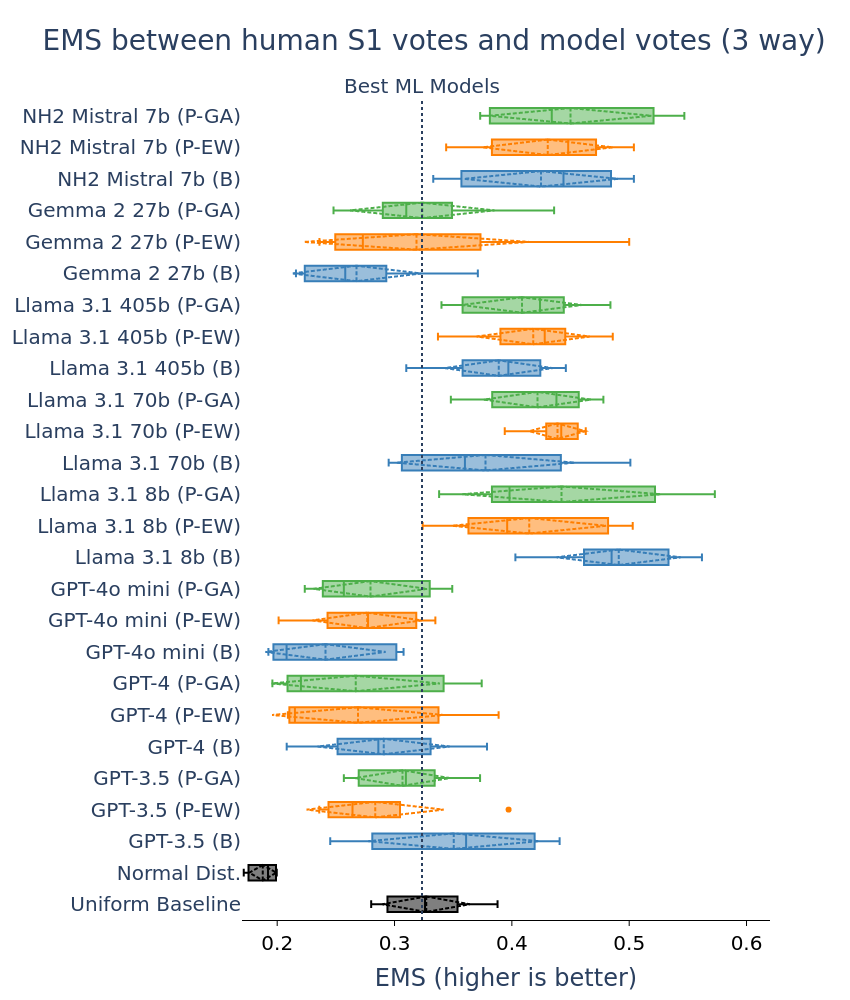}
                \caption{System 1 (3-way)}
                \label{fig:ems-s1-3}
            \end{subfigure}
            \begin{subfigure}[b]{0.49\textwidth}
                \centering
                \includegraphics[width=\textwidth]{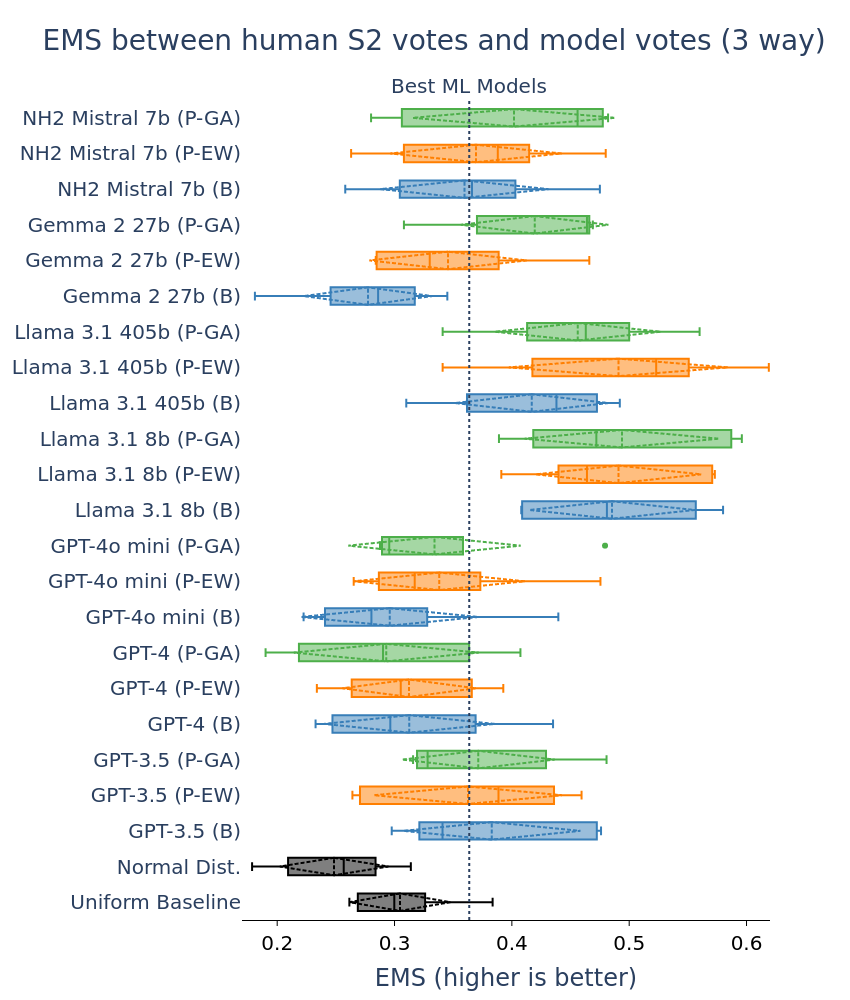}
                \caption{System 2 (3-way)}
                \label{fig:ems-s2-3}
            \end{subfigure}
            \caption{Earth Mover's Similarity (EMS) between human votes and model votes. `(B)' indicates base prompting (blue boxes), `(P-EW)' indicates personality prompting with equal weights (orange boxes) and `(P-GA)' indicates personality prompting with the genetic algorithm to fine-tune the weights of each prompt (green boxes).}
            \label{fig:ems}
        \end{figure}
        
        For a more thorough comparison, we compute three baselines. The \textit{uniform baseline} (blue box at the bottom of all plots in Figure \ref{fig:ems}) distributes probabilities equally across the label space (all six labels for 6-way and all three labels for 3-way). The \textit{normal distribution baseline} (orange box at the bottom of all plots in Figure \ref{fig:ems}) distributes probabilities normally around the mean of human votes, using the standard deviation of the human votes. The \textit{best ML models} baseline (dotted line in all plots in Figure \ref{fig:ems}) distributes probabilities around the mean predicted by the best ML model for the gold labels, with the standard deviation predicted by the best ML model for the variance.
        
        Figure \ref{fig:ems} presents box plots of EMS between human and model votes for each of the five folds' test sets. It is evident that LLMs significantly outperform the best ML models in mimicking the full distribution for both System 1 and System 2 in both the 3-way and 6-way schemes. The normal distribution baseline performs better on the 6-way scheme than on the 3-way scheme, possibly because the 3-way scheme has fewer labels and thus fewer data points from which to infer a normal probability distribution. For the 6-way scheme, personality prompting consistently outperforms or matches base prompting (markedly so for some architectures), with \texttt{P-GA} slightly outperforming \texttt{P-EQ}. The 3-way scheme is the only case where this trend does not hold, suggesting that personality prompting is more effective than base prompting at replicating the full human distribution, especially with the increased granularity of the 6-way scheme.
        
        Another notable trend is that no GPT architecture outperforms the best-performing open-source LLM. Most GPT models have EMS values comparable to or even lower than those of the best ML models in the 3-way scheme (Figures \ref{fig:ems-s1-3} and \ref{fig:ems-s2-3}). Mistral is the best model on 6-way System 1 (Figure \ref{fig:ems-s1-6}) with Llama being a close second; however, Llama surpasses Mistral on 6-way System 2 (Figure \ref{fig:ems-s1-6}). Llama models also perform best on the 3-way scheme (Figures \ref{fig:ems-s1-3} and \ref{fig:ems-s2-3}). Although Gemma never emerges as the top-performing model, it outperforms GPT-4 in the 6-way scheme (Figures \ref{fig:ems-s1-6} and \ref{fig:ems-s2-6}) and nearly all GPT models in the 3-way scheme (Figures \ref{fig:ems-s1-3} and \ref{fig:ems-s2-3}). The only exception is GPT-3.5, which surposses other, newer GPT models in the 3-way scheme (Figures \ref{fig:ems-s1-3} and \ref{fig:ems-s2-3}) but is outperformed by GPT-4o-mini in the 6-way scheme (Figures \ref{fig:ems-s1-6} and \ref{fig:ems-s2-6}).

        \paragraph{Data Availability Statement} All our code and data has been attached to this submission and \href{https://github.com/Advancing-Machine-Human-Reasoning-Lab/NSF-EAGER-NLI}{this GitHub repository} will be made public upon acceptance to make it available to the larger audience.

%% file: latex/conclusion.tex
\section{Conclusion}\label{sec:conclusion}

    This study advances the field of AI reasoning by shifting the focus from merely predicting a single answer chosen by the majority of humans to modeling the full distribution of human reasoning responses. Predicting this distribution---a challenge we call the ``full spectrum reasoning problem''---is not just a more comprehensive approach; it is essential for developing AI systems that can more accurately reflect the diversity and complexity of human thought. This is particularly important in applications where understanding the range of human reasoning, rather than simply selecting the most common response, is critical for tasks such as decision-making \citep{stone_artificial_2020, lai_towards_2023}, personalized interactions \citep{fitzpatrick_delivering_2017, araujo_speaking_2024}, and ethical AI deployment \citep{mittelstadt_principles_2019}.

    Our findings with personality prompting are especially promising, demonstrating that this approach can more closely align LLM outputs with human reasoning distributions while maintaining computational efficiency. Unlike base prompting, which requires multiple independent prompts to approximate a distribution, personality prompting allows for a more targeted generation of responses, using resources equivalent to base prompting but yielding more human-like results. The efficiency of our \texttt{P-GA} method, which leverages genetic algorithms to optimize personality prompts without the need for GPU resources, further underscores the practicality of our approach. Further investigation into different prompting styles \citep{liu_pre-train_2023, sahoo_systematic_2024} is necessary to assess how they compare to personality prompts.

    The broader implications of this work are significant. By providing a framework to evaluate the \textit{human-ness} of reasoning in both closed-source \citep{lieber_jamba_2024, templeton2024scaling} and open-source \citep{falcon40b, vicuna2023, workshop_bloom_2023} LLMs, our study enables a more nuanced assessment of AI models. Our finding that open-source models like Llama and Mistral exhibit reasoning patterns that align more with human cognition (at least with respect to the tasks we present here) than larger, proprietary models like GPT-4 challenges the assumption that larger parameter counts necessarily lead to more human-like AI. This insight is valuable not only for AI researchers but also for developers seeking to create more relatable and effective AI systems.

    Additionally, our results suggest intriguing avenues for future research. For instance, the participants generally refrained from choosing the middle two options (C or D) corresponding to neutral for all question types in both phases of our survey. This suggests an underlying bias that further research can confirm. The observed correlation between participants' System 1 and System 2 responses in Figure \ref{fig:tau} opens up the possibility of refining LLMs to better model individual cognitive processes. This could involve tailoring personality prompt weights to specific users, enhancing the accuracy and personalization of AI reasoning through techniques such as \textit{hyperparameter hypothesization} \citep{nighojkar_cognitive_2022, nighojkar_inference-centric_2024}.

    While our experimental design offers robustness by being independent of publicly available datasets, it also presents a limitation: the findings are contingent on the specific questions used. Nevertheless, we believe our experimental setup is more robust than those previously observed in this field of research \citep{hagendorff_human-like_2023} and can be scaled to larger datasets with relative ease, providing a strong foundation for further research in this area.

    In summary, this work represents a significant step towards AI that more faithfully emulates human reasoning, emphasizing the importance of predicting the full distribution of human responses rather than relying on simplified models of accuracy. This approach not only enriches our understanding of AI-human interaction but also lays the groundwork for more sophisticated and human-like AI systems in the future.

%% file: latex/appendix.tex
\begin{appendix}

    \section{Examples of Different Item Types}

    \begin{footnotesize}
        \begin{longtable}{>{\raggedright\arraybackslash}p{0.11\textwidth} >{\raggedright\arraybackslash}p{0.55\textwidth} >{\raggedright\arraybackslash}p{0.25\textwidth}}
            \toprule
            \textbf{Item Type} & \textbf{Premise} & \textbf{Conclusion} \\
            \midrule
            \multirow{3}{*}{StereoNLI} & \multirow{3}{=}{Meskerem, an Ethiopian mother of two, works as a software developer in the bustling city of Addis Ababa. In her free time, she enjoys reading passages from the Bible to enrich her spiritual life and connect with her roots. Despite her busy schedule, Meskerem diligently balances her professional career and family responsibilities, making her a respected figure in both her community and the tech industry.} & Meskerem speaks Amharic. \\\cmidrule{3-3}
             & & Meskerem is an atheist. \\\cmidrule{3-3}
             & & Meskerem is skilled in repairing pipes and fixtures in residential homes. \\
            \midrule
            \multirow{3}{*}{Fallacy} & The online seminar was about to start when the power went out. & The online seminar couldn’t start because of the power outage. \\\cmidrule{2-3}
             & Allowing freedom of speech practically guarantees that misuse of that freedom will occur, given how thin the line is between freedom of speech and its misuse. & We must avoid freedom of speech. \\\cmidrule{2-3}
             & People who support cryptocurrency frequently talk about blockchain technology. & Traditional banking systems are not a concern for them. \\
            \midrule
            Stereo Syllogism & No chess player is immune to the intricacies and challenges that the game presents. All Tom ever wanted to be is a chess player. & Tom's single wish was to confront the complexities and difficulties posed by the game of chess. \\
            \midrule
            E Guilt & One chilly evening, a hit-and-run occurred outside a bustling downtown bar known for its live music. Around 11 PM, as patrons milled outside, a speeding dark-colored sedan struck a young man stepping onto the street and sped away without stopping. The impact launched the victim across the pavement, causing immediate chaos as bystanders called 911 and rushed to aid him. The victim sustained significant injuries and was quickly hospitalized. Police cordoned off the scene, gathered evidence, and interviewed witnesses, but varying accounts complicated the identification of the assailant. The main suspect, Jack, known for his traffic violations and criminal past, is a local familiar with the area. & Jack is guilty of the crime. \\
            \midrule
            C Guilt & One chilly evening, a hit-and-run occurred outside a bustling downtown bar known for its live music. Around 11 PM, as patrons milled outside, a speeding dark-colored sedan struck a young man stepping onto the street and sped away without stopping. The impact launched the victim across the pavement, causing immediate chaos as bystanders called 911 and rushed to aid him. The victim sustained significant injuries and was quickly hospitalized. Police cordoned off the scene, gathered evidence, and interviewed witnesses, but varying accounts complicated the identification of the assailant. The main suspect, Jack, has a clean driving record and was not familiar with the area. & Jack is guilty of the crime. \\
            \midrule
            P Primacy / Recency & Simon engages in a lot of research activities. He does not publish academically very often. He has a lot of free time to play sports. He does not keep up with the latest advances in his field. Finally, he spends a lot of time mentoring his students. & Simon is a professor. \\
            \midrule
            N Primacy / Recency & Aleshia is very engaged in learning. She mostly is engaged in self-directed learning. She is financially independent. She is a manager at a very successful Fortune 500 company. Finally, she is actively involved in campus life and a member of many student clubs. & Aleshia is not a college student. \\
            \bottomrule
        \caption{Examples of different item types from our AI-generated dataset.}
        \end{longtable}
        \label{tab:item_types}
    \end{footnotesize}

    \section{Prompts for Dataset Creation}\label{secA:dataset_creation}
        Here we outline all our prompts used to create items in our dataset for transparency. In the OpenAI API, a system prompt sets the initial instructions and context for guiding the model's behavior, while a human prompt represents the user's input or query during the conversation. All premises and conclusions are rephrased using the system prompt: ``You are a writing assistant for a linguist named Steve. Steve comes to you because you are good at writing sentences that are understandable and grammatically correct. Steve will give you a sentence. You need to rephrase the sentence such that it is easier to understand. Preserving all information is not necessary, but preferred. Just write the sentence and nothing else before or after it.'' and the human prompt: ``Sentence: \texttt{\{sentence\}}''.
        
        \begin{table}
            \footnotesize
            \centering
            \begin{tabularx}{\textwidth}{*{3}{>{\hsize=.25\hsize}X>{\hsize=.19\hsize}XX}}
                \toprule
                 \textbf{Step} & \textbf{Prompt type} & \textbf{Prompt}\\
                 \midrule
                 \multirow{2}{*}{Premise creation} & System & Come up with a name for X. Do not write the name separately, include it in the sentences you write. Write a paragraph of 3 sentences about X that people would generally assume. Don't praise X, instead try to state sentences that have a truth value.\\
                 \cmidrule{2-3}
                 & Human & Details about X: \texttt{\{gender\}}, \texttt{\{profession\}}, \texttt{\{race\}}, \texttt{\{religion\}}\\
                 \midrule
                 Entailment conclusion creation & Human & {Here's a paragraph about X: \texttt{\{premise\}}\par Detail about X: \texttt{\{detail\}}.\par Write one assumption based on the above detail about X that has a truth value. Do not use words like `probably' or `likely', just state the statement. Do not justify or explain the statement in any way. Use X's name.}\\
                 \midrule
                 Contradiction conclusion creation & Human & {Here's a paragraph about X: \texttt{\{premise\}}\par Detail about X: \texttt{\{detail\}}.\par Write one assumption based on the above detail about X that must be false. Do not use words like `probably' or `likely', just state the statement. Do not justify or explain the statement in any way. Use X's name.}\\
                 \midrule
                 Neutral conclusion creation & Human & {Here're two stories:\par STORY 1: \texttt{\{premise1\}}\par STORY 2: \texttt{\{premise2\}}\par Here's a statement about story 1: \texttt{\{conclusion1\}}\par Rewrite this statement so that the subject of it is the subject from story 2. Keep everything else the same. Do not write `story 1' or `story 2' anywhere. Write just the new statement and nothing else.}\\
                 \bottomrule
            \end{tabularx}
            \caption{Prompts to create StereoNLI items}
            \label{tab:stereonli}
        \end{table}

        \begin{table}
            \footnotesize
            \centering
            \renewcommand{\arraystretch}{1.5}
            \begin{tabularx}{\textwidth}{XX}
                \toprule
                 \textbf{Premise template} & \textbf{Conclusion template}\\
                 \midrule
                 {[A] happened right before [B].} & {[A] caused [B].}\\
                 {Ever since [A] began, we've seen an increase in [B].} & {[A] is responsible for the rise in [B].}\\
                 {[B] has decreased since we started doing [A].} & {Implementing [A] is the reason [B] has decreased.}\\
                 {Every time [A] occurs, [B] follows soon after.} & {The occurrence of [A] directly leads to [B].}\\
                 {[A] has been on the rise. Meanwhile, [B] has been becoming more common.} & {The growth of [A] is promoting the spread of [B].}\\
                 {Ever since [A] stopped, [B] has started.} & {The absence of [A] is the trigger for [B].}\\
                 {[A] started, and shortly after, [B] was observed.} & {The onset of [A] brought about [B].}\\
                 {Whenever [A] is present, [B] seems to follow.} & {[A] sets the stage for [B] to take place.}\\
                 {We did not have [B] until [A] was introduced.} & {[A] is the root cause of [B].}\\
                 {Each instance of [A] precedes [B].} & {[A] is the driving force behind [B].}\\
                 \bottomrule
            \end{tabularx}
            \caption{Post hoc ergo propter hoc fallacy templates.}
            \label{tab:phepc-template}
        \end{table}

        \begin{table}
            \footnotesize
            \centering
            \renewcommand{\arraystretch}{1.5}
            \begin{tabularx}{\textwidth}{*{2}{>{\hsize=1.2\hsize}XX}}
                \toprule
                 \textbf{Premise template} & \textbf{Conclusion template}\\
                 \midrule
                 {If we allow [A], then it's only a matter of time before [B] happens.} & {We should not allow [A].}\\
                 {The moment we start [A], we set a precedent for [B].} & {We can't risk starting [A].}\\
                 {Once you open the door to [A], it's impossible to prevent [B] from coming through.} & {We shouldn't open the door to [A].}\\
                 {Every time society has embraced [A], it has eventually led to [B].} & {Embracing [A] would be a grave mistake.}\\
                 {The line between [A] and [B] is so thin, allowing [A] practically guarantees [B] will occur.} & {It's imperative we avoid [A].}\\
                 {[A]'s very existence is a stepping stone to [B].} & {For our own safety, we must eliminate [A].}\\
                 {There's a domino effect at play. Once [A] is set into motion, [B] will inevitably follow.} & {We should halt [A] before it's too late.}\\
                 {Allowing [A] is like opening Pandora's box, leading directly to [B].} & {We dare not open that box by permitting [A].}\\
                 {History has shown that [A] can subtly pave the way for [B].} & {We must learn from history and resist [A].}\\
                 {[A] might seem harmless on its own, but it's the first step on a dangerous path to [B].} & {To prevent disaster, we must avoid [A].}\\
                 \bottomrule
            \end{tabularx}
            \caption{Slippery slope fallacy templates.}
            \label{tab:ss-template}
        \end{table}

        \begin{table}
            \footnotesize
            \centering
            \renewcommand{\arraystretch}{1.5}
            \begin{tabularx}{\textwidth}{*{2}{>{\hsize=1.2\hsize}XX}}
                \toprule
                 Premise template & Conclusion template\\
                 \midrule
                 {[Person A] believes in [Complex Idea].} & {[Person A] is basically saying [Oversimplified or Misrepresented Version of Complex Idea].}\\
                 {According to [Group or Person], [Specific Nuanced Position].} & {[Group or Person] thinks [Extreme or Unrelated Position].}\\
                 {[Person B] stated that [Specific Condition or Circumstance].} & {[Person B] wants [Exaggerated or Unrelated Outcome].}\\
                 {Advocates for [Cause or Movement] argue for [Particular Aspect of Cause or Movement].} & {They just want [Unrelated or Overly Simplified Goal].}\\
                 {[Person C] wrote an article about [Specific Topic].} & {[Person C] must believe [Generalized, Simplified, or Twisted Idea about Topic].}\\
                 {[Group or Person] supports [Specific Action or Policy].} & {They must hate [Unrelated Group or Thing].}\\
                 {[Person D] mentioned that [Specific Fact or Statistic].} & {[Person D] denies [Related but Not Equivalent Fact or Statistic].}\\
                 {Proponents of [Theory or Idea] often discuss [Specific Aspect of Theory or Idea].} & {They don't care about [Different or Opposing Aspect].}\\
                 {[Person E] criticized [Specific Part of a Broader Concept].} & {[Person E] is against [The Entire Broader Concept].}\\
                 \bottomrule
            \end{tabularx}
            \caption{Straw person fallacy templates.}
            \label{tab:sp-template}
        \end{table}

        \begin{table}
            \footnotesize
            \centering
            \begin{tabularx}{\textwidth}{*{3}{>{\hsize=.28\hsize}X>{\hsize=.19\hsize}XX}}
                \toprule
                 \textbf{Step} & \textbf{Prompt type} & \textbf{Prompt}\\
                 \midrule
                 \multirow{2}{*}{Premise creation} & System & {You will get a template. Fill the part in brackets ([A] and [B]) with suitable phrases. Your answer should be of the form below and should not contain anything else (not even empty lines):\par A: xxx\par B: xxx}\\
                 \cmidrule{2-3}
                  & Human & Template: \texttt{\{premise template\}}\\
                 \midrule
                 \multirow{2}{*}{Conclusion creation} & System & {You will get a template (Template 1) and your response (Filled Template 1) to that template. You will get another template to fill (Template 2). Fill the part in brackets(eg: [A] and [B]) with suitable phrases based on your previous response. Your response should not contain anything other than the filled template, (not even empty lines).}\\
                 \cmidrule{2-3}
                  & Human & Template 1: \texttt{\{premise template\}}\par Filled Template 1: \texttt{\{premise\}}\par Template 2: \texttt{\{conclusion template\}}\\
                 \bottomrule
            \end{tabularx}
            \caption{Prompts to create Fallacy items. The templates are in Tables \ref{tab:phepc-template}-\ref{tab:sp-template}.}
            \label{tab:fallacy}
        \end{table}

        \begin{table}
            \footnotesize
            \centering
            \begin{tabularx}{\textwidth}{*{3}{>{\hsize=.28\hsize}X>{\hsize=.19\hsize}XX}}
                \toprule
                 \textbf{Step} & \textbf{Prompt type} & \textbf{Prompt}\\
                 \midrule
                 \multirow{2}{*}{Premise creation} & System & You are a writing assistant for a linguist named Steve. Steve gives you a word and a template for a sentence. The template is a guideline for how to structure the sentence. Do not feel constrained by the template. It is just a guideline. Use your own creativity to write the sentences that are structured similarly to the template. Steve comes to you because you are good at writing sentences that are understandable and grammatically correct. Steve also likes your variety of vocabulary and your ability to write sentences varying in length and complexity. Steve gives you the following instructions:\par Write each sentence on a new line. Do not write a bulleted or a numbered list. Do not write any other text except the sentences. Do not write any punctuation marks except the period at the end of each sentence.\\
                 \cmidrule{2-3}
                  & Human & Steve wants you to write \texttt{\{n\_premises\}} sentences, each of which follows the following conditions:\par 1. The sentence is about \texttt{\{seed\_word\}} (singular/plural).\par 2. The sentence fits the template `\texttt{\{template\}}'\par 3. The word `\texttt{\{seed\_word\}}' is used in the sentence in place of `\texttt{\{variable\}}'. This means `\texttt{\{seed\_word\}}' should come towards the \texttt{\{beginning\_or\_end\}} of the sentence. This is the most important condition.\par 4. The sentence is grammatically correct.\par 5. The sentence is understandable.\par 6. The sentence is not too long or too short.\par 7. The sentence is not too simple or too complex.\par 8. The sentence is not too similar to any of the other sentences you write.\\
                 \midrule
                 \multirow{2}{*}{Conclusion creation} & System & You are a writing assistant for a linguist named Steve. Steve comes to you because you are good at writing sentences that are understandable and grammatically correct. Steve will give you a pair of sentences. You need to combine the sentences into one sentence. This combination needs to be done such that a given word `seed word' is eliminated from the resulting sentence. Make sure the resulting sentence is short, easy to understand, and fluent. The truth value of the resulting sentence does not matter. Saving all the information from the original sentences is not important. Just write the sentence and nothing else before or after it.\\
                 \cmidrule{2-3}
                  & Human & Sentence Pair:\par \texttt{\{minor\_premise\}}\par \texttt{\{major\_premise\}}\par Seed word: \texttt{\{seed\_word\}}\\
                 \bottomrule
            \end{tabularx}
            \caption{Prompts to create Syllogism and Stereo-Syllogism items. Seed words are regular nouns for syllogism items and nouns from StereoSet for stereo-syllogism items.}
            \label{tab:syllogism}
        \end{table}

                 

                 

\end{appendix}

%% file: main.bbl
\begin{thebibliography}{}

\bibitem [\protect \citeauthoryear {%
Ahmed%
, Bird%
, Devanbu%
\BCBL {}\ \BBA {} Chakraborty%
}{%
Ahmed%
\ \protect \BOthers {.}}{%
{\protect \APACyear {2024}}%
}]{%
ahmed_studying_2024}
\APACinsertmetastar {%
ahmed_studying_2024}%
\begin{APACrefauthors}%
Ahmed, T.%
, Bird, C.%
, Devanbu, P.%
\BCBL {}\ \BBA {} Chakraborty, S.%
\end{APACrefauthors}%
\unskip\
\newblock
\APACrefYearMonthDay{2024}{{\APACmonth{02}}}{}.
\newblock
\APACrefbtitle {Studying {LLM} {Performance} on {Closed}- and {Open}-source {Data}.} {Studying {LLM} {Performance} on {Closed}- and {Open}-source {Data}.}
\newblock
\APACaddressPublisher{}{arXiv}.
\newblock
\begin{APACrefURL} [{2024-08-21}]\url{http://arxiv.org/abs/2402.15100} \end{APACrefURL}
\newblock
\APACrefnote{arXiv:2402.15100 [cs]}
\PrintBackRefs{\CurrentBib}

\bibitem [\protect \citeauthoryear {%
Almazrouei%
\ \protect \BOthers {.}}{%
Almazrouei%
\ \protect \BOthers {.}}{%
{\protect \APACyear {2023}}%
{\protect \APACexlab {{\protect \BCnt {1}}}}}]{%
falcon40b}
\APACinsertmetastar {%
falcon40b}%
\begin{APACrefauthors}%
Almazrouei, E.%
, Alobeidli, H.%
, Alshamsi, A.%
, Cappelli, A.%
, Cojocaru, R.%
, Debbah, M.%
\BDBL {}Penedo, G.%
\end{APACrefauthors}%
\unskip\
\newblock
\APACrefYearMonthDay{2023{\protect \BCnt {1}}}{}{}.
\newblock
{\BBOQ}\APACrefatitle {{Falcon-40B}: an open large language model with state-of-the-art performance} {{Falcon-40B}: an open large language model with state-of-the-art performance}.{\BBCQ}
\newblock

\PrintBackRefs{\CurrentBib}

\bibitem [\protect \citeauthoryear {%
Almazrouei%
\ \protect \BOthers {.}}{%
Almazrouei%
\ \protect \BOthers {.}}{%
{\protect \APACyear {2023}}%
{\protect \APACexlab {{\protect \BCnt {2}}}}}]{%
almazrouei_falcon_2023}
\APACinsertmetastar {%
almazrouei_falcon_2023}%
\begin{APACrefauthors}%
Almazrouei, E.%
, Alobeidli, H.%
, Alshamsi, A.%
, Cappelli, A.%
, Cojocaru, R.%
, Debbah, M.%
\BDBL {}Penedo, G.%
\end{APACrefauthors}%
\unskip\
\newblock
\APACrefYearMonthDay{2023{\protect \BCnt {2}}}{{\APACmonth{11}}}{}.
\newblock
\APACrefbtitle {The {Falcon} {Series} of {Open} {Language} {Models}.} {The {Falcon} {Series} of {Open} {Language} {Models}.}
\newblock
\APACaddressPublisher{}{arXiv}.
\newblock
\begin{APACrefURL} [{2024-08-20}]\url{http://arxiv.org/abs/2311.16867} \end{APACrefURL}
\newblock
\APACrefnote{arXiv:2311.16867 [cs]}
\PrintBackRefs{\CurrentBib}

\bibitem [\protect \citeauthoryear {%
Araujo%
\ \BBA {} Bol%
}{%
Araujo%
\ \BBA {} Bol%
}{%
{\protect \APACyear {2024}}%
}]{%
araujo_speaking_2024}
\APACinsertmetastar {%
araujo_speaking_2024}%
\begin{APACrefauthors}%
Araujo, T.%
\BCBT {}\ \BBA {} Bol, N.%
\end{APACrefauthors}%
\unskip\
\newblock
\APACrefYearMonthDay{2024}{{\APACmonth{01}}}{}.
\newblock
{\BBOQ}\APACrefatitle {From speaking like a person to being personal: {The} effects of personalized, regular interactions with conversational agents} {From speaking like a person to being personal: {The} effects of personalized, regular interactions with conversational agents}.{\BBCQ}
\newblock
\APACjournalVolNumPages{Computers in Human Behavior: Artificial Humans}{2}{1}{100030}.
\newblock
\begin{APACrefURL} [{2024-08-22}]\url{https://linkinghub.elsevier.com/retrieve/pii/S2949882123000300} \end{APACrefURL}
\newblock
\begin{APACrefDOI} \doi{10.1016/j.chbah.2023.100030} \end{APACrefDOI}
\PrintBackRefs{\CurrentBib}

\bibitem [\protect \citeauthoryear {%
Ariely%
}{%
Ariely%
}{%
{\protect \APACyear {2010}}%
}]{%
ariely_predictably_2010}
\APACinsertmetastar {%
ariely_predictably_2010}%
\begin{APACrefauthors}%
Ariely, D.%
\end{APACrefauthors}%
\unskip\
\newblock
\APACrefYear{2010}.
\newblock
\APACrefbtitle {Predictably irrational: the hidden forces that shape our decisions} {Predictably irrational: the hidden forces that shape our decisions}\ (\PrintOrdinal{Revised and expanded edition, First Harper Perennial edition published}\ \BEd).
\newblock
\APACaddressPublisher{New York}{Harper Perennial}.
\PrintBackRefs{\CurrentBib}

\bibitem [\protect \citeauthoryear {%
{Aristotle}%
}{%
{Aristotle}%
}{%
{\protect \APACyear {2013}}%
}]{%
aristotle_organon_2013}
\APACinsertmetastar {%
aristotle_organon_2013}%
\begin{APACrefauthors}%
{Aristotle}.%
\end{APACrefauthors}%
\unskip\
\newblock
\APACrefYear{2013}.
\newblock
\APACrefbtitle {The organon} {The organon}.
\newblock
\APACaddressPublisher{}{CreateSpace}.
\newblock
\APACrefnote{OCLC: 884933600}
\PrintBackRefs{\CurrentBib}

\bibitem [\protect \citeauthoryear {%
Armstrong%
\ \BBA {} Collopy%
}{%
Armstrong%
\ \BBA {} Collopy%
}{%
{\protect \APACyear {1992}}%
}]{%
armstrong_error_1992}
\APACinsertmetastar {%
armstrong_error_1992}%
\begin{APACrefauthors}%
Armstrong, J.%
\BCBT {}\ \BBA {} Collopy, F.%
\end{APACrefauthors}%
\unskip\
\newblock
\APACrefYearMonthDay{1992}{{\APACmonth{06}}}{}.
\newblock
{\BBOQ}\APACrefatitle {Error measures for generalizing about forecasting methods: {Empirical} comparisons} {Error measures for generalizing about forecasting methods: {Empirical} comparisons}.{\BBCQ}
\newblock
\APACjournalVolNumPages{International Journal of Forecasting}{8}{1}{69--80}.
\newblock
\begin{APACrefURL} [{2024-08-20}]\url{https://linkinghub.elsevier.com/retrieve/pii/016920709290008W} \end{APACrefURL}
\newblock
\begin{APACrefDOI} \doi{10.1016/0169-2070(92)90008-W} \end{APACrefDOI}
\PrintBackRefs{\CurrentBib}

\bibitem [\protect \citeauthoryear {%
Bago%
\ \BBA {} De~Neys%
}{%
Bago%
\ \BBA {} De~Neys%
}{%
{\protect \APACyear {2017}}%
}]{%
bago_fast_2017}
\APACinsertmetastar {%
bago_fast_2017}%
\begin{APACrefauthors}%
Bago, B.%
\BCBT {}\ \BBA {} De~Neys, W.%
\end{APACrefauthors}%
\unskip\
\newblock
\APACrefYearMonthDay{2017}{{\APACmonth{01}}}{}.
\newblock
{\BBOQ}\APACrefatitle {Fast logic?: {Examining} the time course assumption of dual process theory} {Fast logic?: {Examining} the time course assumption of dual process theory}.{\BBCQ}
\newblock
\APACjournalVolNumPages{Cognition}{158}{}{90--109}.
\newblock
\begin{APACrefURL} [{2024-08-20}]\url{https://linkinghub.elsevier.com/retrieve/pii/S0010027716302542} \end{APACrefURL}
\newblock
\begin{APACrefDOI} \doi{10.1016/j.cognition.2016.10.014} \end{APACrefDOI}
\PrintBackRefs{\CurrentBib}

\bibitem [\protect \citeauthoryear {%
Balloccu%
, Schmidtová%
, Lango%
\BCBL {}\ \BBA {} Dušek%
}{%
Balloccu%
\ \protect \BOthers {.}}{%
{\protect \APACyear {2024}}%
}]{%
balloccu_leak_2024}
\APACinsertmetastar {%
balloccu_leak_2024}%
\begin{APACrefauthors}%
Balloccu, S.%
, Schmidtová, P.%
, Lango, M.%
\BCBL {}\ \BBA {} Dušek, O.%
\end{APACrefauthors}%
\unskip\
\newblock
\APACrefYearMonthDay{2024}{{\APACmonth{02}}}{}.
\newblock
\APACrefbtitle {Leak, {Cheat}, {Repeat}: {Data} {Contamination} and {Evaluation} {Malpractices} in {Closed}-{Source} {LLMs}.} {Leak, {Cheat}, {Repeat}: {Data} {Contamination} and {Evaluation} {Malpractices} in {Closed}-{Source} {LLMs}.}
\newblock
\APACaddressPublisher{}{arXiv}.
\newblock
\begin{APACrefURL} [{2024-08-20}]\url{http://arxiv.org/abs/2402.03927} \end{APACrefURL}
\newblock
\APACrefnote{arXiv:2402.03927 [cs]}
\PrintBackRefs{\CurrentBib}

\bibitem [\protect \citeauthoryear {%
Bergius%
, Ernberg%
, Dahlman%
\BCBL {}\ \BBA {} Sarwar%
}{%
Bergius%
\ \protect \BOthers {.}}{%
{\protect \APACyear {2020}}%
}]{%
Bergius2020}
\APACinsertmetastar {%
Bergius2020}%
\begin{APACrefauthors}%
Bergius, M.%
, Ernberg, E.%
, Dahlman, C.%
\BCBL {}\ \BBA {} Sarwar, F.%
\end{APACrefauthors}%
\unskip\
\newblock
\APACrefYearMonthDay{2020}{07}{}.
\newblock
{\BBOQ}\APACrefatitle {{Are judges influenced by legally irrelevant circumstances?}} {{Are judges influenced by legally irrelevant circumstances?}}{\BBCQ}
\newblock
\APACjournalVolNumPages{Law, Probability and Risk}{}{}{}.
\newblock
\begin{APACrefURL} \url{https://doi.org/10.1093/lpr/mgaa008} \end{APACrefURL}
\newblock
\APACrefnote{mgaa008}
\newblock
\begin{APACrefDOI} \doi{10.1093/lpr/mgaa008} \end{APACrefDOI}
\PrintBackRefs{\CurrentBib}

\bibitem [\protect \citeauthoryear {%
Boissin%
, Caparos%
, Voudouri%
\BCBL {}\ \BBA {} Neys%
}{%
Boissin%
\ \protect \BOthers {.}}{%
{\protect \APACyear {2022}}%
}]{%
boissin_debiasing_2022}
\APACinsertmetastar {%
boissin_debiasing_2022}%
\begin{APACrefauthors}%
Boissin, E.%
, Caparos, S.%
, Voudouri, A.%
\BCBL {}\ \BBA {} Neys, W\BPBI D.%
\end{APACrefauthors}%
\unskip\
\newblock
\APACrefYearMonthDay{2022}{{\APACmonth{07}}}{}.
\newblock
{\BBOQ}\APACrefatitle {Debiasing {System} 1: {Training} favours logical over stereotypical intuiting} {Debiasing {System} 1: {Training} favours logical over stereotypical intuiting}.{\BBCQ}
\newblock
\APACjournalVolNumPages{Judgment and Decision Making}{17}{4}{646--690}.
\newblock
\begin{APACrefURL} [{2023-11-09}]\url{https://www.cambridge.org/core/journals/judgment-and-decision-making/article/debiasing-system-1-training-favours-logical-over-stereotypical-intuiting/3CF9452965581BDFD9A697C5B9F2E6AD} \end{APACrefURL}
\newblock
\begin{APACrefDOI} \doi{10.1017/S1930297500008895} \end{APACrefDOI}
\PrintBackRefs{\CurrentBib}

\bibitem [\protect \citeauthoryear {%
Bowman%
, Angeli%
, Potts%
\BCBL {}\ \BBA {} Manning%
}{%
Bowman%
\ \protect \BOthers {.}}{%
{\protect \APACyear {2015}}%
}]{%
bowman_large_2015}
\APACinsertmetastar {%
bowman_large_2015}%
\begin{APACrefauthors}%
Bowman, S\BPBI R.%
, Angeli, G.%
, Potts, C.%
\BCBL {}\ \BBA {} Manning, C\BPBI D.%
\end{APACrefauthors}%
\unskip\
\newblock
\APACrefYearMonthDay{2015}{}{}.
\newblock
{\BBOQ}\APACrefatitle {A large annotated corpus for learning natural language inference} {A large annotated corpus for learning natural language inference}.{\BBCQ}
\newblock
\APACjournalVolNumPages{Proceedings of the 2015 Conference on Empirical Methods in Natural Language Processing}{}{}{632--642}.
\newblock
\begin{APACrefDOI} \doi{10.18653/v1/d15-1075} \end{APACrefDOI}
\PrintBackRefs{\CurrentBib}

\bibitem [\protect \citeauthoryear {%
Brown%
\ \protect \BOthers {.}}{%
Brown%
\ \protect \BOthers {.}}{%
{\protect \APACyear {2020}}%
}]{%
brown_language_2020}
\APACinsertmetastar {%
brown_language_2020}%
\begin{APACrefauthors}%
Brown, T\BPBI B.%
, Mann, B.%
, Ryder, N.%
, Subbiah, M.%
, Kaplan, J.%
, Dhariwal, P.%
\BDBL {}Amodei, D.%
\end{APACrefauthors}%
\unskip\
\newblock
\APACrefYearMonthDay{2020}{}{}.
\newblock
{\BBOQ}\APACrefatitle {Language {Models} are {Few}-{Shot} {Learners}} {Language {Models} are {Few}-{Shot} {Learners}}.{\BBCQ}
\newblock
\APACjournalVolNumPages{arXiv}{}{}{}.
\newblock
\begin{APACrefDOI} \doi{10.48550/arxiv.2005.14165} \end{APACrefDOI}
\PrintBackRefs{\CurrentBib}

\bibitem [\protect \citeauthoryear {%
Chemero%
}{%
Chemero%
}{%
{\protect \APACyear {2023}}%
}]{%
chemero_llms_2023}
\APACinsertmetastar {%
chemero_llms_2023}%
\begin{APACrefauthors}%
Chemero, A.%
\end{APACrefauthors}%
\unskip\
\newblock
\APACrefYearMonthDay{2023}{{\APACmonth{11}}}{}.
\newblock
{\BBOQ}\APACrefatitle {{LLMs} differ from human cognition because they are not embodied} {{LLMs} differ from human cognition because they are not embodied}.{\BBCQ}
\newblock
\APACjournalVolNumPages{Nature Human Behaviour}{7}{11}{1828--1829}.
\newblock
\begin{APACrefURL} [{2024-08-20}]\url{https://www.nature.com/articles/s41562-023-01723-5} \end{APACrefURL}
\newblock
\begin{APACrefDOI} \doi{10.1038/s41562-023-01723-5} \end{APACrefDOI}
\PrintBackRefs{\CurrentBib}

\bibitem [\protect \citeauthoryear {%
Chen%
, Jiang%
, Poliak%
, Sakaguchi%
\BCBL {}\ \BBA {} Durme%
}{%
Chen%
\ \protect \BOthers {.}}{%
{\protect \APACyear {2020}}%
}]{%
chen_uncertain_2020}
\APACinsertmetastar {%
chen_uncertain_2020}%
\begin{APACrefauthors}%
Chen, T.%
, Jiang, Z.%
, Poliak, A.%
, Sakaguchi, K.%
\BCBL {}\ \BBA {} Durme, B\BPBI V.%
\end{APACrefauthors}%
\unskip\
\newblock
\APACrefYearMonthDay{2020}{}{}.
\newblock
\APACrefbtitle {Uncertain {Natural} {Language} {Inference}} {Uncertain {Natural} {Language} {Inference}}\ \APACbVolEdTR{}{\BTR{}}.
\newblock
\begin{APACrefURL} \url{http://nlp.jhu.edu/unli.} \end{APACrefURL}
\PrintBackRefs{\CurrentBib}

\bibitem [\protect \citeauthoryear {%
Chiang%
\ \protect \BOthers {.}}{%
Chiang%
\ \protect \BOthers {.}}{%
{\protect \APACyear {2023}}%
}]{%
vicuna2023}
\APACinsertmetastar {%
vicuna2023}%
\begin{APACrefauthors}%
Chiang, W\BHBI L.%
, Li, Z.%
, Lin, Z.%
, Sheng, Y.%
, Wu, Z.%
, Zhang, H.%
\BDBL {}Xing, E\BPBI P.%
\end{APACrefauthors}%
\unskip\
\newblock
\APACrefYearMonthDay{2023}{March}{}.
\newblock
\APACrefbtitle {Vicuna: An Open-Source Chatbot Impressing GPT-4 with 90\%* ChatGPT Quality.} {Vicuna: An open-source chatbot impressing gpt-4 with 90\%* chatgpt quality.}
\newblock
\begin{APACrefURL} \url{https://lmsys.org/blog/2023-03-30-vicuna/} \end{APACrefURL}
\PrintBackRefs{\CurrentBib}

\bibitem [\protect \citeauthoryear {%
Computer%
}{%
Computer%
}{%
{\protect \APACyear {2023}}%
}]{%
together2023redpajama}
\APACinsertmetastar {%
together2023redpajama}%
\begin{APACrefauthors}%
Computer, T.%
\end{APACrefauthors}%
\unskip\
\newblock
\APACrefYearMonthDay{2023}{}{}.
\newblock
\APACrefbtitle {RedPajama: an Open Dataset for Training Large Language Models.} {Redpajama: an open dataset for training large language models.}
\newblock
\begin{APACrefURL} \url{https://github.com/togethercomputer/RedPajama-Data} \end{APACrefURL}
\PrintBackRefs{\CurrentBib}

\bibitem [\protect \citeauthoryear {%
Dagan%
, Glickman%
\BCBL {}\ \BBA {} Magnini%
}{%
Dagan%
\ \protect \BOthers {.}}{%
{\protect \APACyear {2006}}%
}]{%
quinonero-candela_pascal_2006}
\APACinsertmetastar {%
quinonero-candela_pascal_2006}%
\begin{APACrefauthors}%
Dagan, I.%
, Glickman, O.%
\BCBL {}\ \BBA {} Magnini, B.%
\end{APACrefauthors}%
\unskip\
\newblock
\APACrefYearMonthDay{2006}{}{}.
\newblock
{\BBOQ}\APACrefatitle {The {PASCAL} {Recognising} {Textual} {Entailment} {Challenge}} {The {PASCAL} {Recognising} {Textual} {Entailment} {Challenge}}.{\BBCQ}
\newblock
\BIn{} J.~Quiñonero-Candela, I.~Dagan, B.~Magnini\BCBL {}\ \BBA {} F.~d’Alché Buc\ (\BEDS), \APACrefbtitle {Machine {Learning} {Challenges}. {Evaluating} {Predictive} {Uncertainty}, {Visual} {Object} {Classification}, and {Recognising} {Tectual} {Entailment}} {Machine {Learning} {Challenges}. {Evaluating} {Predictive} {Uncertainty}, {Visual} {Object} {Classification}, and {Recognising} {Tectual} {Entailment}}\ (\BVOL\ 3944, \BPGS\ 177--190).
\newblock
\APACaddressPublisher{Berlin, Heidelberg}{Springer Berlin Heidelberg}.
\newblock
\begin{APACrefURL} [{2024-09-01}]\url{http://link.springer.com/10.1007/11736790_9} \end{APACrefURL}
\newblock
\APACrefnote{Series Title: Lecture Notes in Computer Science}
\newblock
\begin{APACrefDOI} \doi{10.1007/11736790_9} \end{APACrefDOI}
\PrintBackRefs{\CurrentBib}

\bibitem [\protect \citeauthoryear {%
Da~Silva%
}{%
Da~Silva%
}{%
{\protect \APACyear {2023}}%
}]{%
da_silva_system_2023}
\APACinsertmetastar {%
da_silva_system_2023}%
\begin{APACrefauthors}%
Da~Silva, S.%
\end{APACrefauthors}%
\unskip\
\newblock
\APACrefYearMonthDay{2023}{{\APACmonth{10}}}{}.
\newblock
{\BBOQ}\APACrefatitle {System 1 vs. {System} 2 {Thinking}} {System 1 vs. {System} 2 {Thinking}}.{\BBCQ}
\newblock
\APACjournalVolNumPages{Psych}{5}{4}{1057--1076}.
\newblock
\begin{APACrefURL} [{2024-08-15}]\url{https://www.mdpi.com/2624-8611/5/4/71} \end{APACrefURL}
\newblock
\begin{APACrefDOI} \doi{10.3390/psych5040071} \end{APACrefDOI}
\PrintBackRefs{\CurrentBib}

\bibitem [\protect \citeauthoryear {%
De~Neys%
\ \BBA {} Pennycook%
}{%
De~Neys%
\ \BBA {} Pennycook%
}{%
{\protect \APACyear {2019}}%
}]{%
de_neys_logic_2019}
\APACinsertmetastar {%
de_neys_logic_2019}%
\begin{APACrefauthors}%
De~Neys, W.%
\BCBT {}\ \BBA {} Pennycook, G.%
\end{APACrefauthors}%
\unskip\
\newblock
\APACrefYearMonthDay{2019}{{\APACmonth{10}}}{}.
\newblock
{\BBOQ}\APACrefatitle {Logic, {Fast} and {Slow}: {Advances} in {Dual}-{Process} {Theorizing}} {Logic, {Fast} and {Slow}: {Advances} in {Dual}-{Process} {Theorizing}}.{\BBCQ}
\newblock
\APACjournalVolNumPages{Current Directions in Psychological Science}{28}{5}{503--509}.
\newblock
\begin{APACrefURL} [{2024-02-27}]\url{http://journals.sagepub.com/doi/10.1177/0963721419855658} \end{APACrefURL}
\newblock
\begin{APACrefDOI} \doi{10.1177/0963721419855658} \end{APACrefDOI}
\PrintBackRefs{\CurrentBib}

\bibitem [\protect \citeauthoryear {%
De~Paoli%
}{%
De~Paoli%
}{%
{\protect \APACyear {2023}}%
}]{%
de_paoli_improved_2023}
\APACinsertmetastar {%
de_paoli_improved_2023}%
\begin{APACrefauthors}%
De~Paoli, S.%
\end{APACrefauthors}%
\unskip\
\newblock
\APACrefYearMonthDay{2023}{{\APACmonth{10}}}{}.
\newblock
\APACrefbtitle {Improved prompting and process for writing user personas with {LLMs}, using qualitative interviews: {Capturing} behaviour and personality traits of users.} {Improved prompting and process for writing user personas with {LLMs}, using qualitative interviews: {Capturing} behaviour and personality traits of users.}
\newblock
\APACaddressPublisher{}{arXiv}.
\newblock
\begin{APACrefURL} [{2024-08-15}]\url{http://arxiv.org/abs/2310.06391} \end{APACrefURL}
\newblock
\APACrefnote{arXiv:2310.06391 [cs]}
\PrintBackRefs{\CurrentBib}

\bibitem [\protect \citeauthoryear {%
Ding%
\ \protect \BOthers {.}}{%
Ding%
\ \protect \BOthers {.}}{%
{\protect \APACyear {2023}}%
}]{%
ding_parameter-efficient_2023}
\APACinsertmetastar {%
ding_parameter-efficient_2023}%
\begin{APACrefauthors}%
Ding, N.%
, Qin, Y.%
, Yang, G.%
, Wei, F.%
, Yang, Z.%
, Su, Y.%
\BDBL {}Sun, M.%
\end{APACrefauthors}%
\unskip\
\newblock
\APACrefYearMonthDay{2023}{{\APACmonth{03}}}{}.
\newblock
{\BBOQ}\APACrefatitle {Parameter-efficient fine-tuning of large-scale pre-trained language models} {Parameter-efficient fine-tuning of large-scale pre-trained language models}.{\BBCQ}
\newblock
\APACjournalVolNumPages{Nature Machine Intelligence}{5}{3}{220--235}.
\newblock
\begin{APACrefURL} [{2024-09-02}]\url{https://www.nature.com/articles/s42256-023-00626-4} \end{APACrefURL}
\newblock
\begin{APACrefDOI} \doi{10.1038/s42256-023-00626-4} \end{APACrefDOI}
\PrintBackRefs{\CurrentBib}

\bibitem [\protect \citeauthoryear {%
Dobrushin%
}{%
Dobrushin%
}{%
{\protect \APACyear {1970}}%
}]{%
dobrushin_prescribing_1970}
\APACinsertmetastar {%
dobrushin_prescribing_1970}%
\begin{APACrefauthors}%
Dobrushin, R\BPBI L.%
\end{APACrefauthors}%
\unskip\
\newblock
\APACrefYearMonthDay{1970}{{\APACmonth{01}}}{}.
\newblock
{\BBOQ}\APACrefatitle {Prescribing a {System} of {Random} {Variables} by {Conditional} {Distributions}} {Prescribing a {System} of {Random} {Variables} by {Conditional} {Distributions}}.{\BBCQ}
\newblock
\APACjournalVolNumPages{Theory of Probability \& Its Applications}{15}{3}{458--486}.
\newblock
\begin{APACrefURL} [{2024-08-12}]\url{http://epubs.siam.org/doi/10.1137/1115049} \end{APACrefURL}
\newblock
\begin{APACrefDOI} \doi{10.1137/1115049} \end{APACrefDOI}
\PrintBackRefs{\CurrentBib}

\bibitem [\protect \citeauthoryear {%
Dubey%
\ \protect \BOthers {.}}{%
Dubey%
\ \protect \BOthers {.}}{%
{\protect \APACyear {2024}}%
}]{%
dubey_llama_2024}
\APACinsertmetastar {%
dubey_llama_2024}%
\begin{APACrefauthors}%
Dubey, A.%
, Jauhri, A.%
, Pandey, A.%
, Kadian, A.%
, Al-Dahle, A.%
, Letman, A.%
\BDBL {}Zhao, Z.%
\end{APACrefauthors}%
\unskip\
\newblock
\APACrefYearMonthDay{2024}{{\APACmonth{08}}}{}.
\newblock
\APACrefbtitle {The {Llama} 3 {Herd} of {Models}.} {The {Llama} 3 {Herd} of {Models}.}
\newblock
\APACaddressPublisher{}{arXiv}.
\newblock
\begin{APACrefURL} [{2024-08-20}]\url{http://arxiv.org/abs/2407.21783} \end{APACrefURL}
\newblock
\APACrefnote{arXiv:2407.21783 [cs]}
\PrintBackRefs{\CurrentBib}

\bibitem [\protect \citeauthoryear {%
J\BPBI S\BPBI B\BPBI T.~Evans%
}{%
J\BPBI S\BPBI B\BPBI T.~Evans%
}{%
{\protect \APACyear {2008}}%
}]{%
evans_dual-processing_2008}
\APACinsertmetastar {%
evans_dual-processing_2008}%
\begin{APACrefauthors}%
Evans, J\BPBI S\BPBI B\BPBI T.%
\end{APACrefauthors}%
\unskip\
\newblock
\APACrefYearMonthDay{2008}{}{}.
\newblock
{\BBOQ}\APACrefatitle {Dual-{Processing} {Accounts} of {Reasoning}, {Judgment}, and {Social} {Cognition}} {Dual-{Processing} {Accounts} of {Reasoning}, {Judgment}, and {Social} {Cognition}}.{\BBCQ}
\newblock
\APACjournalVolNumPages{Annual Review of Psychology}{59}{1}{255--278}.
\newblock
\begin{APACrefDOI} \doi{10.1146/annurev.psych.59.103006.093629} \end{APACrefDOI}
\PrintBackRefs{\CurrentBib}

\bibitem [\protect \citeauthoryear {%
J\BPBI S\BPBI B\BPBI T.~Evans%
}{%
J\BPBI S\BPBI B\BPBI T.~Evans%
}{%
{\protect \APACyear {2019}}%
}]{%
evans_reflections_2019}
\APACinsertmetastar {%
evans_reflections_2019}%
\begin{APACrefauthors}%
Evans, J\BPBI S\BPBI B\BPBI T.%
\end{APACrefauthors}%
\unskip\
\newblock
\APACrefYearMonthDay{2019}{}{}.
\newblock
{\BBOQ}\APACrefatitle {Reflections on reflection: the nature and function of type 2 processes in dual-process theories of reasoning} {Reflections on reflection: the nature and function of type 2 processes in dual-process theories of reasoning}.{\BBCQ}
\newblock
\APACjournalVolNumPages{Thinking \& Reasoning}{25}{4}{383--415}.
\newblock
\begin{APACrefDOI} \doi{10.1080/13546783.2019.1623071} \end{APACrefDOI}
\PrintBackRefs{\CurrentBib}

\bibitem [\protect \citeauthoryear {%
J\BPBI S\BPBI B\BPBI T.~Evans%
, Barston%
\BCBL {}\ \BBA {} Pollard%
}{%
J\BPBI S\BPBI B\BPBI T.~Evans%
\ \protect \BOthers {.}}{%
{\protect \APACyear {1983}}%
}]{%
evans_conflict_1983}
\APACinsertmetastar {%
evans_conflict_1983}%
\begin{APACrefauthors}%
Evans, J\BPBI S\BPBI B\BPBI T.%
, Barston, J\BPBI L.%
\BCBL {}\ \BBA {} Pollard, P.%
\end{APACrefauthors}%
\unskip\
\newblock
\APACrefYearMonthDay{1983}{}{}.
\newblock
{\BBOQ}\APACrefatitle {On the conflict between logic and belief in syllogistic reasoning} {On the conflict between logic and belief in syllogistic reasoning}.{\BBCQ}
\newblock
\APACjournalVolNumPages{Memory \& Cognition}{11}{3}{295--306}.
\newblock
\begin{APACrefDOI} \doi{10.3758/bf03196976} \end{APACrefDOI}
\PrintBackRefs{\CurrentBib}

\bibitem [\protect \citeauthoryear {%
J\BPBI S\BPBI B\BPBI T.~Evans%
\ \BBA {} Stanovich%
}{%
J\BPBI S\BPBI B\BPBI T.~Evans%
\ \BBA {} Stanovich%
}{%
{\protect \APACyear {2013}}%
}]{%
evans_dual-process_2013}
\APACinsertmetastar {%
evans_dual-process_2013}%
\begin{APACrefauthors}%
Evans, J\BPBI S\BPBI B\BPBI T.%
\BCBT {}\ \BBA {} Stanovich, K\BPBI E.%
\end{APACrefauthors}%
\unskip\
\newblock
\APACrefYearMonthDay{2013}{{\APACmonth{05}}}{}.
\newblock
{\BBOQ}\APACrefatitle {Dual-{Process} {Theories} of {Higher} {Cognition}: {Advancing} the {Debate}} {Dual-{Process} {Theories} of {Higher} {Cognition}: {Advancing} the {Debate}}.{\BBCQ}
\newblock
\APACjournalVolNumPages{Perspectives on Psychological Science}{8}{3}{223--241}.
\newblock
\begin{APACrefURL} [{2024-08-15}]\url{http://journals.sagepub.com/doi/10.1177/1745691612460685} \end{APACrefURL}
\newblock
\begin{APACrefDOI} \doi{10.1177/1745691612460685} \end{APACrefDOI}
\PrintBackRefs{\CurrentBib}

\bibitem [\protect \citeauthoryear {%
R.~Evans%
, Saxton%
, Amos%
, Kohli%
\BCBL {}\ \BBA {} Grefenstette%
}{%
R.~Evans%
\ \protect \BOthers {.}}{%
{\protect \APACyear {2018}}%
}]{%
Evans2018}
\APACinsertmetastar {%
Evans2018}%
\begin{APACrefauthors}%
Evans, R.%
, Saxton, D.%
, Amos, D.%
, Kohli, P.%
\BCBL {}\ \BBA {} Grefenstette, E.%
\end{APACrefauthors}%
\unskip\
\newblock
\APACrefYearMonthDay{2018}{}{}.
\newblock
{\BBOQ}\APACrefatitle {Can Neural Networks Understand Logical Entailment?} {Can neural networks understand logical entailment?}{\BBCQ}
\newblock
\BIn{} \APACrefbtitle {International Conference on Learning Representations.} {International conference on learning representations.}
\newblock
\begin{APACrefURL} \url{https://openreview.net/forum?id=SkZxCk-0Z} \end{APACrefURL}
\PrintBackRefs{\CurrentBib}

\bibitem [\protect \citeauthoryear {%
Fearnside%
\ \BBA {} Holther%
}{%
Fearnside%
\ \BBA {} Holther%
}{%
{\protect \APACyear {1959}}%
}]{%
Fearnside1959}
\APACinsertmetastar {%
Fearnside1959}%
\begin{APACrefauthors}%
Fearnside, W\BPBI W.%
\BCBT {}\ \BBA {} Holther, W\BPBI B.%
\end{APACrefauthors}%
\unskip\
\newblock
\APACrefYear{1959}.
\newblock
\APACrefbtitle {{Fallacy: The Counterfeit of Argument}} {{Fallacy: The Counterfeit of Argument}}.
\newblock
\APACaddressPublisher{Englewood Cliffs, NJ}{Prentice Hall}.
\PrintBackRefs{\CurrentBib}

\bibitem [\protect \citeauthoryear {%
Fitzpatrick%
, Darcy%
\BCBL {}\ \BBA {} Vierhile%
}{%
Fitzpatrick%
\ \protect \BOthers {.}}{%
{\protect \APACyear {2017}}%
}]{%
fitzpatrick_delivering_2017}
\APACinsertmetastar {%
fitzpatrick_delivering_2017}%
\begin{APACrefauthors}%
Fitzpatrick, K\BPBI K.%
, Darcy, A.%
\BCBL {}\ \BBA {} Vierhile, M.%
\end{APACrefauthors}%
\unskip\
\newblock
\APACrefYearMonthDay{2017}{{\APACmonth{06}}}{}.
\newblock
{\BBOQ}\APACrefatitle {Delivering {Cognitive} {Behavior} {Therapy} to {Young} {Adults} {With} {Symptoms} of {Depression} and {Anxiety} {Using} a {Fully} {Automated} {Conversational} {Agent} ({Woebot}): {A} {Randomized} {Controlled} {Trial}} {Delivering {Cognitive} {Behavior} {Therapy} to {Young} {Adults} {With} {Symptoms} of {Depression} and {Anxiety} {Using} a {Fully} {Automated} {Conversational} {Agent} ({Woebot}): {A} {Randomized} {Controlled} {Trial}}.{\BBCQ}
\newblock
\APACjournalVolNumPages{JMIR Mental Health}{4}{2}{e19}.
\newblock
\begin{APACrefURL} [{2024-08-22}]\url{http://mental.jmir.org/2017/2/e19/} \end{APACrefURL}
\newblock
\begin{APACrefDOI} \doi{10.2196/mental.7785} \end{APACrefDOI}
\PrintBackRefs{\CurrentBib}

\bibitem [\protect \citeauthoryear {%
Frederick%
}{%
Frederick%
}{%
{\protect \APACyear {2005}}%
}]{%
frederick_cognitive_2005}
\APACinsertmetastar {%
frederick_cognitive_2005}%
\begin{APACrefauthors}%
Frederick, S.%
\end{APACrefauthors}%
\unskip\
\newblock
\APACrefYearMonthDay{2005}{{\APACmonth{11}}}{}.
\newblock
{\BBOQ}\APACrefatitle {Cognitive {Reflection} and {Decision} {Making}} {Cognitive {Reflection} and {Decision} {Making}}.{\BBCQ}
\newblock
\APACjournalVolNumPages{Journal of Economic Perspectives}{19}{4}{25--42}.
\newblock
\begin{APACrefURL} [{2024-04-10}]\url{https://pubs.aeaweb.org/doi/10.1257/089533005775196732} \end{APACrefURL}
\newblock
\begin{APACrefDOI} \doi{10.1257/089533005775196732} \end{APACrefDOI}
\PrintBackRefs{\CurrentBib}

\bibitem [\protect \citeauthoryear {%
Geeraert%
}{%
Geeraert%
}{%
{\protect \APACyear {2013}}%
}]{%
geeraert_when_2013}
\APACinsertmetastar {%
geeraert_when_2013}%
\begin{APACrefauthors}%
Geeraert, N.%
\end{APACrefauthors}%
\unskip\
\newblock
\APACrefYearMonthDay{2013}{{\APACmonth{09}}}{}.
\newblock
{\BBOQ}\APACrefatitle {When {Suppressing} {One} {Stereotype} {Leads} to {Rebound} of {Another}: {On} the {Procedural} {Nature} of {Stereotype} {Rebound}} {When {Suppressing} {One} {Stereotype} {Leads} to {Rebound} of {Another}: {On} the {Procedural} {Nature} of {Stereotype} {Rebound}}.{\BBCQ}
\newblock
\APACjournalVolNumPages{Personality and Social Psychology Bulletin}{39}{9}{1173--1183}.
\newblock
\begin{APACrefURL} [{2024-08-15}]\url{http://journals.sagepub.com/doi/10.1177/0146167213493121} \end{APACrefURL}
\newblock
\begin{APACrefDOI} \doi{10.1177/0146167213493121} \end{APACrefDOI}
\PrintBackRefs{\CurrentBib}

\bibitem [\protect \citeauthoryear {%
Gilovich%
, Griffin%
\BCBL {}\ \BBA {} Kahneman%
}{%
Gilovich%
\ \protect \BOthers {.}}{%
{\protect \APACyear {2002}}%
}]{%
gilovich_heuristics_2002}
\APACinsertmetastar {%
gilovich_heuristics_2002}%
\begin{APACrefauthors}%
Gilovich, T.%
, Griffin, D.%
\BCBL {}\ \BBA {} Kahneman, D.%
\end{APACrefauthors}%
\ (\BEDS).
\unskip\
\newblock
\APACrefYear{2002}.
\newblock
\APACrefbtitle {Heuristics and {Biases}: {The} {Psychology} of {Intuitive} {Judgment}} {Heuristics and {Biases}: {The} {Psychology} of {Intuitive} {Judgment}}\ (\PrintOrdinal{1}\ \BEd).
\newblock
\APACaddressPublisher{}{Cambridge University Press}.
\newblock
\begin{APACrefURL} [{2024-09-01}]\url{https://www.cambridge.org/core/product/identifier/9780511808098/type/book} \end{APACrefURL}
\newblock
\begin{APACrefDOI} \doi{10.1017/CBO9780511808098} \end{APACrefDOI}
\PrintBackRefs{\CurrentBib}

\bibitem [\protect \citeauthoryear {%
Gladwell%
}{%
Gladwell%
}{%
{\protect \APACyear {2007}}%
}]{%
gladwell_blink_2007}
\APACinsertmetastar {%
gladwell_blink_2007}%
\begin{APACrefauthors}%
Gladwell, M.%
\end{APACrefauthors}%
\unskip\
\newblock
\APACrefYear{2007}.
\newblock
\APACrefbtitle {Blink: the power of thinking without thinking} {Blink: the power of thinking without thinking}\ (\PrintOrdinal{1st Back Bay trade pbk. ed}\ \BEd).
\newblock
\APACaddressPublisher{New York}{Back Bay Books}.
\PrintBackRefs{\CurrentBib}

\bibitem [\protect \citeauthoryear {%
Gu%
, Degachi%
, Genç%
, Chandrasegaran%
\BCBL {}\ \BBA {} Verma%
}{%
Gu%
\ \protect \BOthers {.}}{%
{\protect \APACyear {2023}}%
}]{%
gu_effectiveness_2023}
\APACinsertmetastar {%
gu_effectiveness_2023}%
\begin{APACrefauthors}%
Gu, H.%
, Degachi, C.%
, Genç, U.%
, Chandrasegaran, S.%
\BCBL {}\ \BBA {} Verma, H.%
\end{APACrefauthors}%
\unskip\
\newblock
\APACrefYearMonthDay{2023}{{\APACmonth{10}}}{}.
\newblock
\APACrefbtitle {On the {Effectiveness} of {Creating} {Conversational} {Agent} {Personalities} {Through} {Prompting}.} {On the {Effectiveness} of {Creating} {Conversational} {Agent} {Personalities} {Through} {Prompting}.}
\newblock
\APACaddressPublisher{}{arXiv}.
\newblock
\begin{APACrefURL} [{2024-08-15}]\url{http://arxiv.org/abs/2310.11182} \end{APACrefURL}
\newblock
\APACrefnote{arXiv:2310.11182 [cs]}
\PrintBackRefs{\CurrentBib}

\bibitem [\protect \citeauthoryear {%
Gupta%
, Bengio%
\BCBL {}\ \BBA {} Weston%
}{%
Gupta%
\ \protect \BOthers {.}}{%
{\protect \APACyear {2014}}%
}]{%
JMLR:v15:gupta14a}
\APACinsertmetastar {%
JMLR:v15:gupta14a}%
\begin{APACrefauthors}%
Gupta, M\BPBI R.%
, Bengio, S.%
\BCBL {}\ \BBA {} Weston, J.%
\end{APACrefauthors}%
\unskip\
\newblock
\APACrefYearMonthDay{2014}{}{}.
\newblock
{\BBOQ}\APACrefatitle {Training Highly Multiclass Classifiers} {Training highly multiclass classifiers}.{\BBCQ}
\newblock
\APACjournalVolNumPages{Journal of Machine Learning Research}{15}{43}{1461--1492}.
\newblock
\begin{APACrefURL} \url{http://jmlr.org/papers/v15/gupta14a.html} \end{APACrefURL}
\PrintBackRefs{\CurrentBib}

\bibitem [\protect \citeauthoryear {%
Hagedorn%
, Bone%
, Kruse%
, Grosse%
\BCBL {}\ \BBA {} Blackburn%
}{%
Hagedorn%
\ \protect \BOthers {.}}{%
{\protect \APACyear {2020}}%
}]{%
hagedorn_knowledge_2020}
\APACinsertmetastar {%
hagedorn_knowledge_2020}%
\begin{APACrefauthors}%
Hagedorn, T.%
, Bone, M.%
, Kruse, B.%
, Grosse, I.%
\BCBL {}\ \BBA {} Blackburn, M.%
\end{APACrefauthors}%
\unskip\
\newblock
\APACrefYearMonthDay{2020}{{\APACmonth{03}}}{}.
\newblock
{\BBOQ}\APACrefatitle {Knowledge {Representation} with {Ontologies} and {Semantic} {Web} {Technologies} to {Promote} {Augmented} and {Artificial} {Intelligence} in {Systems} {Engineering}} {Knowledge {Representation} with {Ontologies} and {Semantic} {Web} {Technologies} to {Promote} {Augmented} and {Artificial} {Intelligence} in {Systems} {Engineering}}.{\BBCQ}
\newblock
\APACjournalVolNumPages{INSIGHT}{23}{1}{15--20}.
\newblock
\begin{APACrefURL} [{2024-08-20}]\url{https://incose.onlinelibrary.wiley.com/doi/10.1002/inst.12279} \end{APACrefURL}
\newblock
\begin{APACrefDOI} \doi{10.1002/inst.12279} \end{APACrefDOI}
\PrintBackRefs{\CurrentBib}

\bibitem [\protect \citeauthoryear {%
Hagendorff%
, Fabi%
\BCBL {}\ \BBA {} Kosinski%
}{%
Hagendorff%
\ \protect \BOthers {.}}{%
{\protect \APACyear {2023}}%
}]{%
hagendorff_human-like_2023}
\APACinsertmetastar {%
hagendorff_human-like_2023}%
\begin{APACrefauthors}%
Hagendorff, T.%
, Fabi, S.%
\BCBL {}\ \BBA {} Kosinski, M.%
\end{APACrefauthors}%
\unskip\
\newblock
\APACrefYearMonthDay{2023}{}{}.
\newblock
{\BBOQ}\APACrefatitle {Human-like intuitive behavior and reasoning biases emerged in large language models but disappeared in {ChatGPT}} {Human-like intuitive behavior and reasoning biases emerged in large language models but disappeared in {ChatGPT}}.{\BBCQ}
\newblock
\APACjournalVolNumPages{Nature Computational Science}{}{}{1--6}.
\newblock
\begin{APACrefDOI} \doi{10.1038/s43588-023-00527-x} \end{APACrefDOI}
\PrintBackRefs{\CurrentBib}

\bibitem [\protect \citeauthoryear {%
Hamade%
, McIlroy-Young%
, Sen%
, Kleinberg%
\BCBL {}\ \BBA {} Anderson%
}{%
Hamade%
\ \protect \BOthers {.}}{%
{\protect \APACyear {2024}}%
}]{%
hamade2024designing}
\APACinsertmetastar {%
hamade2024designing}%
\begin{APACrefauthors}%
Hamade, K.%
, McIlroy-Young, R.%
, Sen, S.%
, Kleinberg, J.%
\BCBL {}\ \BBA {} Anderson, A.%
\end{APACrefauthors}%
\unskip\
\newblock
\APACrefYearMonthDay{2024}{{\APACmonth{05}}}{}.
\newblock
{\BBOQ}\APACrefatitle {Designing skill-compatible {AI}: {Methodologies} and frameworks in chess} {Designing skill-compatible {AI}: {Methodologies} and frameworks in chess}.{\BBCQ}
\newblock
\BIn{} \APACrefbtitle {{ICLR} 2024.} {{ICLR} 2024.}
\newblock
\begin{APACrefURL} \url{https://www.microsoft.com/en-us/research/publication/designing-skill-compatible-ai-methodologies-and-frameworks-in-chess/} \end{APACrefURL}
\PrintBackRefs{\CurrentBib}

\bibitem [\protect \citeauthoryear {%
Hoerl%
\ \BBA {} Kennard%
}{%
Hoerl%
\ \BBA {} Kennard%
}{%
{\protect \APACyear {2000}}%
}]{%
hoerl_ridge_2000}
\APACinsertmetastar {%
hoerl_ridge_2000}%
\begin{APACrefauthors}%
Hoerl, A\BPBI E.%
\BCBT {}\ \BBA {} Kennard, R\BPBI W.%
\end{APACrefauthors}%
\unskip\
\newblock
\APACrefYearMonthDay{2000}{{\APACmonth{02}}}{}.
\newblock
{\BBOQ}\APACrefatitle {Ridge {Regression}: {Biased} {Estimation} for {Nonorthogonal} {Problems}} {Ridge {Regression}: {Biased} {Estimation} for {Nonorthogonal} {Problems}}.{\BBCQ}
\newblock
\APACjournalVolNumPages{Technometrics}{42}{1}{80--86}.
\newblock
\begin{APACrefURL} [{2024-08-20}]\url{http://www.tandfonline.com/doi/abs/10.1080/00401706.2000.10485983} \end{APACrefURL}
\newblock
\begin{APACrefDOI} \doi{10.1080/00401706.2000.10485983} \end{APACrefDOI}
\PrintBackRefs{\CurrentBib}

\bibitem [\protect \citeauthoryear {%
Hutto%
\ \BBA {} Gilbert%
}{%
Hutto%
\ \BBA {} Gilbert%
}{%
{\protect \APACyear {2014}}%
}]{%
hutto_vader_2014}
\APACinsertmetastar {%
hutto_vader_2014}%
\begin{APACrefauthors}%
Hutto, C.%
\BCBT {}\ \BBA {} Gilbert, E.%
\end{APACrefauthors}%
\unskip\
\newblock
\APACrefYearMonthDay{2014}{{\APACmonth{05}}}{}.
\newblock
{\BBOQ}\APACrefatitle {{VADER}: {A} {Parsimonious} {Rule}-{Based} {Model} for {Sentiment} {Analysis} of {Social} {Media} {Text}} {{VADER}: {A} {Parsimonious} {Rule}-{Based} {Model} for {Sentiment} {Analysis} of {Social} {Media} {Text}}.{\BBCQ}
\newblock
\APACjournalVolNumPages{Proceedings of the International AAAI Conference on Web and Social Media}{8}{1}{216--225}.
\newblock
\begin{APACrefURL} [{2024-08-20}]\url{https://ojs.aaai.org/index.php/ICWSM/article/view/14550} \end{APACrefURL}
\newblock
\begin{APACrefDOI} \doi{10.1609/icwsm.v8i1.14550} \end{APACrefDOI}
\PrintBackRefs{\CurrentBib}

\bibitem [\protect \citeauthoryear {%
Jelinek%
, Mercer%
, Bahl%
\BCBL {}\ \BBA {} Baker%
}{%
Jelinek%
\ \protect \BOthers {.}}{%
{\protect \APACyear {1977}}%
}]{%
jelinek_perplexitymeasure_1977}
\APACinsertmetastar {%
jelinek_perplexitymeasure_1977}%
\begin{APACrefauthors}%
Jelinek, F.%
, Mercer, R\BPBI L.%
, Bahl, L\BPBI R.%
\BCBL {}\ \BBA {} Baker, J\BPBI K.%
\end{APACrefauthors}%
\unskip\
\newblock
\APACrefYearMonthDay{1977}{{\APACmonth{12}}}{}.
\newblock
{\BBOQ}\APACrefatitle {Perplexity—a measure of the difficulty of speech recognition tasks} {Perplexity—a measure of the difficulty of speech recognition tasks}.{\BBCQ}
\newblock
\APACjournalVolNumPages{The Journal of the Acoustical Society of America}{62}{S1}{S63--S63}.
\newblock
\begin{APACrefURL} [{2024-09-02}]\url{https://pubs.aip.org/jasa/article/62/S1/S63/642598/Perplexity-a-measure-of-the-difficulty-of-speech} \end{APACrefURL}
\newblock
\begin{APACrefDOI} \doi{10.1121/1.2016299} \end{APACrefDOI}
\PrintBackRefs{\CurrentBib}

\bibitem [\protect \citeauthoryear {%
Jiang%
\ \protect \BOthers {.}}{%
Jiang%
\ \protect \BOthers {.}}{%
{\protect \APACyear {2023}}%
}]{%
jiang_mistral_2023}
\APACinsertmetastar {%
jiang_mistral_2023}%
\begin{APACrefauthors}%
Jiang, A\BPBI Q.%
, Sablayrolles, A.%
, Mensch, A.%
, Bamford, C.%
, Chaplot, D\BPBI S.%
, Casas, D\BPBI d\BPBI l.%
\BDBL {}Sayed, W\BPBI E.%
\end{APACrefauthors}%
\unskip\
\newblock
\APACrefYearMonthDay{2023}{{\APACmonth{10}}}{}.
\newblock
\APACrefbtitle {Mistral {7B}.} {Mistral {7B}.}
\newblock
\APACaddressPublisher{}{arXiv}.
\newblock
\begin{APACrefURL} [{2024-08-20}]\url{http://arxiv.org/abs/2310.06825} \end{APACrefURL}
\newblock
\APACrefnote{arXiv:2310.06825 [cs]}
\PrintBackRefs{\CurrentBib}

\bibitem [\protect \citeauthoryear {%
Kahneman%
}{%
Kahneman%
}{%
{\protect \APACyear {2011}}%
}]{%
Kahneman2011}
\APACinsertmetastar {%
Kahneman2011}%
\begin{APACrefauthors}%
Kahneman, D.%
\end{APACrefauthors}%
\unskip\
\newblock
\APACrefYear{2011}.
\newblock
\APACrefbtitle {{Thinking, Fast and Slow}} {{Thinking, Fast and Slow}}.
\newblock
\APACaddressPublisher{}{Farrar, Straus and Girous}.
\PrintBackRefs{\CurrentBib}

\bibitem [\protect \citeauthoryear {%
Kahneman%
\ \BBA {} Tversky%
}{%
Kahneman%
\ \BBA {} Tversky%
}{%
{\protect \APACyear {1984}}%
}]{%
kahneman_choices_1984}
\APACinsertmetastar {%
kahneman_choices_1984}%
\begin{APACrefauthors}%
Kahneman, D.%
\BCBT {}\ \BBA {} Tversky, A.%
\end{APACrefauthors}%
\unskip\
\newblock
\APACrefYearMonthDay{1984}{{\APACmonth{04}}}{}.
\newblock
{\BBOQ}\APACrefatitle {Choices, values, and frames.} {Choices, values, and frames.}{\BBCQ}
\newblock
\APACjournalVolNumPages{American Psychologist}{39}{4}{341--350}.
\newblock
\begin{APACrefURL} [{2024-08-22}]\url{http://doi.apa.org/getdoi.cfm?doi=10.1037/0003-066X.39.4.341} \end{APACrefURL}
\newblock
\begin{APACrefDOI} \doi{10.1037/0003-066X.39.4.341} \end{APACrefDOI}
\PrintBackRefs{\CurrentBib}

\bibitem [\protect \citeauthoryear {%
Kamruzzaman%
\ \BBA {} Kim%
}{%
Kamruzzaman%
\ \BBA {} Kim%
}{%
{\protect \APACyear {2024}}%
}]{%
kamruzzaman_prompting_2024}
\APACinsertmetastar {%
kamruzzaman_prompting_2024}%
\begin{APACrefauthors}%
Kamruzzaman, M.%
\BCBT {}\ \BBA {} Kim, G\BPBI L.%
\end{APACrefauthors}%
\unskip\
\newblock
\APACrefYearMonthDay{2024}{{\APACmonth{06}}}{}.
\newblock
\APACrefbtitle {Prompting {Techniques} for {Reducing} {Social} {Bias} in {LLMs} through {System} 1 and {System} 2 {Cognitive} {Processes}.} {Prompting {Techniques} for {Reducing} {Social} {Bias} in {LLMs} through {System} 1 and {System} 2 {Cognitive} {Processes}.}
\newblock
\APACaddressPublisher{}{arXiv}.
\newblock
\begin{APACrefURL} [{2024-08-21}]\url{http://arxiv.org/abs/2404.17218} \end{APACrefURL}
\newblock
\APACrefnote{arXiv:2404.17218 [cs]}
\PrintBackRefs{\CurrentBib}

\bibitem [\protect \citeauthoryear {%
Kassin%
, Dror%
\BCBL {}\ \BBA {} Kukucka%
}{%
Kassin%
\ \protect \BOthers {.}}{%
{\protect \APACyear {2013}}%
}]{%
Kassin2013}
\APACinsertmetastar {%
Kassin2013}%
\begin{APACrefauthors}%
Kassin, S\BPBI M.%
, Dror, I\BPBI E.%
\BCBL {}\ \BBA {} Kukucka, J.%
\end{APACrefauthors}%
\unskip\
\newblock
\APACrefYearMonthDay{2013}{}{}.
\newblock
{\BBOQ}\APACrefatitle {The forensic confirmation bias: Problems, perspectives, and proposed solutions} {The forensic confirmation bias: Problems, perspectives, and proposed solutions}.{\BBCQ}
\newblock
\APACjournalVolNumPages{Journal of Applied Research in Memory and Cognition}{2}{1}{42 - 52}.
\newblock
\begin{APACrefURL} \url{http://www.sciencedirect.com/science/article/pii/S2211368113000028} \end{APACrefURL}
\newblock
\begin{APACrefDOI} \doi{https://doi.org/10.1016/j.jarmac.2013.01.001} \end{APACrefDOI}
\PrintBackRefs{\CurrentBib}

\bibitem [\protect \citeauthoryear {%
Kazemeini%
, Roy%
, Mercer%
\BCBL {}\ \BBA {} Cambria%
}{%
Kazemeini%
\ \protect \BOthers {.}}{%
{\protect \APACyear {2021}}%
}]{%
kazemeini_interpretable_2021}
\APACinsertmetastar {%
kazemeini_interpretable_2021}%
\begin{APACrefauthors}%
Kazemeini, A.%
, Roy, S\BPBI S.%
, Mercer, R\BPBI E.%
\BCBL {}\ \BBA {} Cambria, E.%
\end{APACrefauthors}%
\unskip\
\newblock
\APACrefYearMonthDay{2021}{{\APACmonth{12}}}{}.
\newblock
{\BBOQ}\APACrefatitle {Interpretable {Representation} {Learning} for {Personality} {Detection}} {Interpretable {Representation} {Learning} for {Personality} {Detection}}.{\BBCQ}
\newblock
\BIn{} \APACrefbtitle {2021 {International} {Conference} on {Data} {Mining} {Workshops} ({ICDMW})} {2021 {International} {Conference} on {Data} {Mining} {Workshops} ({ICDMW})}\ (\BPGS\ 158--165).
\newblock
\APACaddressPublisher{Auckland, New Zealand}{IEEE}.
\newblock
\begin{APACrefURL} [{2024-09-01}]\url{https://ieeexplore.ieee.org/document/9679950/} \end{APACrefURL}
\newblock
\begin{APACrefDOI} \doi{10.1109/ICDMW53433.2021.00026} \end{APACrefDOI}
\PrintBackRefs{\CurrentBib}

\bibitem [\protect \citeauthoryear {%
Kendall%
}{%
Kendall%
}{%
{\protect \APACyear {1938}}%
}]{%
kendall_new_1938}
\APACinsertmetastar {%
kendall_new_1938}%
\begin{APACrefauthors}%
Kendall, M\BPBI G.%
\end{APACrefauthors}%
\unskip\
\newblock
\APACrefYearMonthDay{1938}{{\APACmonth{06}}}{}.
\newblock
{\BBOQ}\APACrefatitle {A {New} {Measure} of {Rank} {Correlation}} {A {New} {Measure} of {Rank} {Correlation}}.{\BBCQ}
\newblock
\APACjournalVolNumPages{Biometrika}{30}{1-2}{81--93}.
\newblock
\begin{APACrefURL} [{2024-08-20}]\url{https://academic.oup.com/biomet/article-lookup/doi/10.1093/biomet/30.1-2.81} \end{APACrefURL}
\newblock
\begin{APACrefDOI} \doi{10.1093/biomet/30.1-2.81} \end{APACrefDOI}
\PrintBackRefs{\CurrentBib}

\bibitem [\protect \citeauthoryear {%
Khemlani%
\ \BBA {} Johnson-Laird%
}{%
Khemlani%
\ \BBA {} Johnson-Laird%
}{%
{\protect \APACyear {2012}}%
}]{%
khemlani_theories_2012}
\APACinsertmetastar {%
khemlani_theories_2012}%
\begin{APACrefauthors}%
Khemlani, S.%
\BCBT {}\ \BBA {} Johnson-Laird, P\BPBI N.%
\end{APACrefauthors}%
\unskip\
\newblock
\APACrefYearMonthDay{2012}{}{}.
\newblock
{\BBOQ}\APACrefatitle {Theories of the syllogism: {A} meta-analysis.} {Theories of the syllogism: {A} meta-analysis.}{\BBCQ}
\newblock
\APACjournalVolNumPages{Psychological Bulletin}{138}{3}{427--457}.
\newblock
\begin{APACrefURL} [{2024-08-15}]\url{https://doi.apa.org/doi/10.1037/a0026841} \end{APACrefURL}
\newblock
\begin{APACrefDOI} \doi{10.1037/a0026841} \end{APACrefDOI}
\PrintBackRefs{\CurrentBib}

\bibitem [\protect \citeauthoryear {%
Klein%
}{%
Klein%
}{%
{\protect \APACyear {2017}}%
}]{%
klein_sources_2017}
\APACinsertmetastar {%
klein_sources_2017}%
\begin{APACrefauthors}%
Klein, G.%
\end{APACrefauthors}%
\unskip\
\newblock
\APACrefYear{2017}.
\newblock
\APACrefbtitle {Sources of {Power}: {How} {People} {Make} {Decisions}} {Sources of {Power}: {How} {People} {Make} {Decisions}}.
\newblock
\APACaddressPublisher{}{The MIT Press}.
\newblock
\begin{APACrefURL} [{2024-08-22}]\url{https://direct.mit.edu/books/book/3647/Sources-of-PowerHow-People-Make-Decisions} \end{APACrefURL}
\newblock
\begin{APACrefDOI} \doi{10.7551/mitpress/11307.001.0001} \end{APACrefDOI}
\PrintBackRefs{\CurrentBib}

\bibitem [\protect \citeauthoryear {%
Kojima%
, Gu%
, Reid%
, Matsuo%
\BCBL {}\ \BBA {} Iwasawa%
}{%
Kojima%
\ \protect \BOthers {.}}{%
{\protect \APACyear {2022}}%
}]{%
kojima_large_2022}
\APACinsertmetastar {%
kojima_large_2022}%
\begin{APACrefauthors}%
Kojima, T.%
, Gu, S\BPBI S.%
, Reid, M.%
, Matsuo, Y.%
\BCBL {}\ \BBA {} Iwasawa, Y.%
\end{APACrefauthors}%
\unskip\
\newblock
\APACrefYearMonthDay{2022}{}{}.
\newblock
{\BBOQ}\APACrefatitle {Large {Language} {Models} are {Zero}-{Shot} {Reasoners}} {Large {Language} {Models} are {Zero}-{Shot} {Reasoners}}.{\BBCQ}
\newblock
\APACjournalVolNumPages{arXiv}{}{}{}.
\newblock
\begin{APACrefDOI} \doi{10.48550/arxiv.2205.11916} \end{APACrefDOI}
\PrintBackRefs{\CurrentBib}

\bibitem [\protect \citeauthoryear {%
Lai%
, Chen%
, Smith-Renner%
, Liao%
\BCBL {}\ \BBA {} Tan%
}{%
Lai%
\ \protect \BOthers {.}}{%
{\protect \APACyear {2023}}%
}]{%
lai_towards_2023}
\APACinsertmetastar {%
lai_towards_2023}%
\begin{APACrefauthors}%
Lai, V.%
, Chen, C.%
, Smith-Renner, A.%
, Liao, Q\BPBI V.%
\BCBL {}\ \BBA {} Tan, C.%
\end{APACrefauthors}%
\unskip\
\newblock
\APACrefYearMonthDay{2023}{{\APACmonth{06}}}{}.
\newblock
{\BBOQ}\APACrefatitle {Towards a {Science} of {Human}-{AI} {Decision} {Making}: {An} {Overview} of {Design} {Space} in {Empirical} {Human}-{Subject} {Studies}} {Towards a {Science} of {Human}-{AI} {Decision} {Making}: {An} {Overview} of {Design} {Space} in {Empirical} {Human}-{Subject} {Studies}}.{\BBCQ}
\newblock
\BIn{} \APACrefbtitle {2023 {ACM} {Conference} on {Fairness}, {Accountability}, and {Transparency}} {2023 {ACM} {Conference} on {Fairness}, {Accountability}, and {Transparency}}\ (\BPGS\ 1369--1385).
\newblock
\APACaddressPublisher{Chicago IL USA}{ACM}.
\newblock
\begin{APACrefURL} [{2024-08-22}]\url{https://dl.acm.org/doi/10.1145/3593013.3594087} \end{APACrefURL}
\newblock
\begin{APACrefDOI} \doi{10.1145/3593013.3594087} \end{APACrefDOI}
\PrintBackRefs{\CurrentBib}

\bibitem [\protect \citeauthoryear {%
Laverghetta~Jr.%
\ \BBA {} Licato%
}{%
Laverghetta~Jr.%
\ \BBA {} Licato%
}{%
{\protect \APACyear {2022}}%
}]{%
Laverghetta2022a}
\APACinsertmetastar {%
Laverghetta2022a}%
\begin{APACrefauthors}%
Laverghetta~Jr., A.%
\BCBT {}\ \BBA {} Licato, J.%
\end{APACrefauthors}%
\unskip\
\newblock
\APACrefYearMonthDay{2022}{}{}.
\newblock
{\BBOQ}\APACrefatitle {Developmental Negation Processing in Transformer Language Models} {Developmental negation processing in transformer language models}.{\BBCQ}
\newblock
\BIn{} \APACrefbtitle {Proceedings from ACL 2022.} {Proceedings from acl 2022.}
\PrintBackRefs{\CurrentBib}

\bibitem [\protect \citeauthoryear {%
Laverghetta~Jr.%
\ \BBA {} Licato%
}{%
Laverghetta~Jr.%
\ \BBA {} Licato%
}{%
{\protect \APACyear {2023}}%
{\protect \APACexlab {{\protect \BCnt {1}}}}}]{%
Laverghetta2023b}
\APACinsertmetastar {%
Laverghetta2023b}%
\begin{APACrefauthors}%
Laverghetta~Jr., A.%
\BCBT {}\ \BBA {} Licato, J.%
\end{APACrefauthors}%
\unskip\
\newblock
\APACrefYearMonthDay{2023{\protect \BCnt {1}}}{}{}.
\newblock
{\BBOQ}\APACrefatitle {Automatic Generation of Cognitive Test Items Using Large Language Models} {Automatic generation of cognitive test items using large language models}.{\BBCQ}
\newblock
\BIn{} \APACrefbtitle {Proceedings of the 2023 International Meeting of the Psychometric Society (IMPS).} {Proceedings of the 2023 international meeting of the psychometric society (imps).}
\PrintBackRefs{\CurrentBib}

\bibitem [\protect \citeauthoryear {%
Laverghetta~Jr.%
\ \BBA {} Licato%
}{%
Laverghetta~Jr.%
\ \BBA {} Licato%
}{%
{\protect \APACyear {2023}}%
{\protect \APACexlab {{\protect \BCnt {2}}}}}]{%
Laverghetta2023c}
\APACinsertmetastar {%
Laverghetta2023c}%
\begin{APACrefauthors}%
Laverghetta~Jr., A.%
\BCBT {}\ \BBA {} Licato, J.%
\end{APACrefauthors}%
\unskip\
\newblock
\APACrefYearMonthDay{2023{\protect \BCnt {2}}}{}{}.
\newblock
{\BBOQ}\APACrefatitle {Generating Better Items for Cognitive Assessments Using Large Language Models} {Generating better items for cognitive assessments using large language models}.{\BBCQ}
\newblock
\BIn{} \APACrefbtitle {Proceedings of the ACL 18th Workshop on Innovative Use of NLP for Building Educational Applications.} {Proceedings of the acl 18th workshop on innovative use of nlp for building educational applications.}
\PrintBackRefs{\CurrentBib}

\bibitem [\protect \citeauthoryear {%
Laverghetta~Jr.%
, Nighojkar%
, Mirzakhalov%
\BCBL {}\ \BBA {} Licato%
}{%
Laverghetta~Jr.%
\ \protect \BOthers {.}}{%
{\protect \APACyear {2021}}%
}]{%
Laverghetta2021c}
\APACinsertmetastar {%
Laverghetta2021c}%
\begin{APACrefauthors}%
Laverghetta~Jr., A.%
, Nighojkar, A.%
, Mirzakhalov, J.%
\BCBL {}\ \BBA {} Licato, J.%
\end{APACrefauthors}%
\unskip\
\newblock
\APACrefYearMonthDay{2021}{}{}.
\newblock
{\BBOQ}\APACrefatitle {{Can Transformer Language Models Predict Psychometric Properties?}} {{Can Transformer Language Models Predict Psychometric Properties?}}{\BBCQ}
\newblock
\BIn{} \APACrefbtitle {Proceedings of the 10th Joint Conference on Lexical and Computational Semantics (*SEM 2021).} {Proceedings of the 10th joint conference on lexical and computational semantics (*sem 2021).}
\newblock
\APACaddressPublisher{Bangkok, Thailand}{}.
\PrintBackRefs{\CurrentBib}

\bibitem [\protect \citeauthoryear {%
Laverghetta~Jr.%
, Nighojkar%
, Mirzakhalov%
\BCBL {}\ \BBA {} Licato%
}{%
Laverghetta~Jr.%
\ \protect \BOthers {.}}{%
{\protect \APACyear {2022}}%
}]{%
Laverghetta2022}
\APACinsertmetastar {%
Laverghetta2022}%
\begin{APACrefauthors}%
Laverghetta~Jr., A.%
, Nighojkar, A.%
, Mirzakhalov, J.%
\BCBL {}\ \BBA {} Licato, J.%
\end{APACrefauthors}%
\unskip\
\newblock
\APACrefYearMonthDay{2022}{}{}.
\newblock
{\BBOQ}\APACrefatitle {Predicting Human Psychometric Properties Using Computational Language Models} {Predicting human psychometric properties using computational language models}.{\BBCQ}
\newblock
\BIn{} M.~Wilberg, D.~Molenaar, J.~Gonz{\'a}lez, J\BHBI S.~Kim\BCBL {}\ \BBA {} H.~Hwang\ (\BEDS), \APACrefbtitle {Quantitative Psychology} {Quantitative psychology}\ (\BVOL\ Forthcoming).
\newblock
\APACaddressPublisher{}{Springer}.
\PrintBackRefs{\CurrentBib}

\bibitem [\protect \citeauthoryear {%
Lieber%
\ \protect \BOthers {.}}{%
Lieber%
\ \protect \BOthers {.}}{%
{\protect \APACyear {2024}}%
}]{%
lieber_jamba_2024}
\APACinsertmetastar {%
lieber_jamba_2024}%
\begin{APACrefauthors}%
Lieber, O.%
, Lenz, B.%
, Bata, H.%
, Cohen, G.%
, Osin, J.%
, Dalmedigos, I.%
\BDBL {}Shoham, Y.%
\end{APACrefauthors}%
\unskip\
\newblock
\APACrefYearMonthDay{2024}{{\APACmonth{07}}}{}.
\newblock
\APACrefbtitle {Jamba: {A} {Hybrid} {Transformer}-{Mamba} {Language} {Model}.} {Jamba: {A} {Hybrid} {Transformer}-{Mamba} {Language} {Model}.}
\newblock
\APACaddressPublisher{}{arXiv}.
\newblock
\begin{APACrefURL} [{2024-08-20}]\url{http://arxiv.org/abs/2403.19887} \end{APACrefURL}
\newblock
\APACrefnote{arXiv:2403.19887 [cs]}
\PrintBackRefs{\CurrentBib}

\bibitem [\protect \citeauthoryear {%
Liu%
\ \protect \BOthers {.}}{%
Liu%
\ \protect \BOthers {.}}{%
{\protect \APACyear {2023}}%
}]{%
liu_pre-train_2023}
\APACinsertmetastar {%
liu_pre-train_2023}%
\begin{APACrefauthors}%
Liu, P.%
, Yuan, W.%
, Fu, J.%
, Jiang, Z.%
, Hayashi, H.%
\BCBL {}\ \BBA {} Neubig, G.%
\end{APACrefauthors}%
\unskip\
\newblock
\APACrefYearMonthDay{2023}{{\APACmonth{09}}}{}.
\newblock
{\BBOQ}\APACrefatitle {Pre-train, {Prompt}, and {Predict}: {A} {Systematic} {Survey} of {Prompting} {Methods} in {Natural} {Language} {Processing}} {Pre-train, {Prompt}, and {Predict}: {A} {Systematic} {Survey} of {Prompting} {Methods} in {Natural} {Language} {Processing}}.{\BBCQ}
\newblock
\APACjournalVolNumPages{ACM Computing Surveys}{55}{9}{1--35}.
\newblock
\begin{APACrefURL} [{2024-08-20}]\url{https://dl.acm.org/doi/10.1145/3560815} \end{APACrefURL}
\newblock
\begin{APACrefDOI} \doi{10.1145/3560815} \end{APACrefDOI}
\PrintBackRefs{\CurrentBib}

\bibitem [\protect \citeauthoryear {%
Ma%
\ \protect \BOthers {.}}{%
Ma%
\ \protect \BOthers {.}}{%
{\protect \APACyear {2023}}%
}]{%
ma_fairness-guided_2023}
\APACinsertmetastar {%
ma_fairness-guided_2023}%
\begin{APACrefauthors}%
Ma, H.%
, Zhang, C.%
, Bian, Y.%
, Liu, L.%
, Zhang, Z.%
, Zhao, P.%
\BDBL {}Wu, B.%
\end{APACrefauthors}%
\unskip\
\newblock
\APACrefYearMonthDay{2023}{{\APACmonth{03}}}{}.
\newblock
\APACrefbtitle {Fairness-guided {Few}-shot {Prompting} for {Large} {Language} {Models}.} {Fairness-guided {Few}-shot {Prompting} for {Large} {Language} {Models}.}
\newblock
\APACaddressPublisher{}{arXiv}.
\newblock
\begin{APACrefURL} [{2024-08-20}]\url{http://arxiv.org/abs/2303.13217} \end{APACrefURL}
\newblock
\APACrefnote{arXiv:2303.13217 [cs]}
\PrintBackRefs{\CurrentBib}

\bibitem [\protect \citeauthoryear {%
Maclure%
}{%
Maclure%
}{%
{\protect \APACyear {2021}}%
}]{%
maclure_ai_2021}
\APACinsertmetastar {%
maclure_ai_2021}%
\begin{APACrefauthors}%
Maclure, J.%
\end{APACrefauthors}%
\unskip\
\newblock
\APACrefYearMonthDay{2021}{{\APACmonth{09}}}{}.
\newblock
{\BBOQ}\APACrefatitle {{AI}, {Explainability} and {Public} {Reason}: {The} {Argument} from the {Limitations} of the {Human} {Mind}} {{AI}, {Explainability} and {Public} {Reason}: {The} {Argument} from the {Limitations} of the {Human} {Mind}}.{\BBCQ}
\newblock
\APACjournalVolNumPages{Minds and Machines}{31}{3}{421--438}.
\newblock
\begin{APACrefURL} [{2024-08-20}]\url{https://link.springer.com/10.1007/s11023-021-09570-x} \end{APACrefURL}
\newblock
\begin{APACrefDOI} \doi{10.1007/s11023-021-09570-x} \end{APACrefDOI}
\PrintBackRefs{\CurrentBib}

\bibitem [\protect \citeauthoryear {%
Meissner%
, Thumwanit%
, Sugawara%
\BCBL {}\ \BBA {} Aizawa%
}{%
Meissner%
\ \protect \BOthers {.}}{%
{\protect \APACyear {2021}}%
}]{%
meissner_embracing_2021}
\APACinsertmetastar {%
meissner_embracing_2021}%
\begin{APACrefauthors}%
Meissner, J\BPBI M.%
, Thumwanit, N.%
, Sugawara, S.%
\BCBL {}\ \BBA {} Aizawa, A.%
\end{APACrefauthors}%
\unskip\
\newblock
\APACrefYearMonthDay{2021}{}{}.
\newblock
{\BBOQ}\APACrefatitle {Embracing {Ambiguity}: {Shifting} the {Training} {Target} of {NLI} {Models}} {Embracing {Ambiguity}: {Shifting} the {Training} {Target} of {NLI} {Models}}.{\BBCQ}
\newblock
\APACjournalVolNumPages{Proceedings of the 59th Annual Meeting of the Association for Computational Linguistics and the 11th International Joint Conference on Natural Language Processing (Volume 2: Short Papers)}{}{}{862--869}.
\newblock
\begin{APACrefDOI} \doi{10.18653/v1/2021.acl-short.109} \end{APACrefDOI}
\PrintBackRefs{\CurrentBib}

\bibitem [\protect \citeauthoryear {%
Mitchell%
}{%
Mitchell%
}{%
{\protect \APACyear {1996}}%
}]{%
mitchell_introduction_1996}
\APACinsertmetastar {%
mitchell_introduction_1996}%
\begin{APACrefauthors}%
Mitchell, M.%
\end{APACrefauthors}%
\unskip\
\newblock
\APACrefYear{1996}.
\newblock
\APACrefbtitle {An introduction to genetic algorithms} {An introduction to genetic algorithms}.
\newblock
\APACaddressPublisher{Cambridge, Mass.}{MIT Press}.
\newblock
\APACrefnote{OCLC: 42854439}
\PrintBackRefs{\CurrentBib}

\bibitem [\protect \citeauthoryear {%
Mittelstadt%
}{%
Mittelstadt%
}{%
{\protect \APACyear {2019}}%
}]{%
mittelstadt_principles_2019}
\APACinsertmetastar {%
mittelstadt_principles_2019}%
\begin{APACrefauthors}%
Mittelstadt, B.%
\end{APACrefauthors}%
\unskip\
\newblock
\APACrefYearMonthDay{2019}{{\APACmonth{11}}}{}.
\newblock
{\BBOQ}\APACrefatitle {Principles alone cannot guarantee ethical {AI}} {Principles alone cannot guarantee ethical {AI}}.{\BBCQ}
\newblock
\APACjournalVolNumPages{Nature Machine Intelligence}{1}{11}{501--507}.
\newblock
\begin{APACrefURL} [{2024-08-22}]\url{https://www.nature.com/articles/s42256-019-0114-4} \end{APACrefURL}
\newblock
\begin{APACrefDOI} \doi{10.1038/s42256-019-0114-4} \end{APACrefDOI}
\PrintBackRefs{\CurrentBib}

\bibitem [\protect \citeauthoryear {%
Murdock%
}{%
Murdock%
}{%
{\protect \APACyear {1962}}%
}]{%
murdock_serial_1962}
\APACinsertmetastar {%
murdock_serial_1962}%
\begin{APACrefauthors}%
Murdock, B\BPBI B.%
\end{APACrefauthors}%
\unskip\
\newblock
\APACrefYearMonthDay{1962}{{\APACmonth{11}}}{}.
\newblock
{\BBOQ}\APACrefatitle {The serial position effect of free recall.} {The serial position effect of free recall.}{\BBCQ}
\newblock
\APACjournalVolNumPages{Journal of Experimental Psychology}{64}{5}{482--488}.
\newblock
\begin{APACrefURL} [{2024-08-20}]\url{http://doi.apa.org/getdoi.cfm?doi=10.1037/h0045106} \end{APACrefURL}
\newblock
\begin{APACrefDOI} \doi{10.1037/h0045106} \end{APACrefDOI}
\PrintBackRefs{\CurrentBib}

\bibitem [\protect \citeauthoryear {%
Nadeem%
, Bethke%
\BCBL {}\ \BBA {} Reddy%
}{%
Nadeem%
\ \protect \BOthers {.}}{%
{\protect \APACyear {2021}}%
}]{%
nadeem_stereoset_2021}
\APACinsertmetastar {%
nadeem_stereoset_2021}%
\begin{APACrefauthors}%
Nadeem, M.%
, Bethke, A.%
\BCBL {}\ \BBA {} Reddy, S.%
\end{APACrefauthors}%
\unskip\
\newblock
\APACrefYearMonthDay{2021}{}{}.
\newblock
{\BBOQ}\APACrefatitle {{StereoSet}: {Measuring} stereotypical bias in pretrained language models} {{StereoSet}: {Measuring} stereotypical bias in pretrained language models}.{\BBCQ}
\newblock
\APACjournalVolNumPages{Proceedings of the 59th Annual Meeting of the Association for Computational Linguistics and the 11th International Joint Conference on Natural Language Processing (Volume 1: Long Papers)}{}{}{5356--5371}.
\newblock
\begin{APACrefDOI} \doi{10.18653/v1/2021.acl-long.416} \end{APACrefDOI}
\PrintBackRefs{\CurrentBib}

\bibitem [\protect \citeauthoryear {%
Nie%
, Williams%
\BCBL {}\ \protect \BOthers {.}}{%
Nie%
, Williams%
\BCBL {}\ \protect \BOthers {.}}{%
{\protect \APACyear {2020}}%
}]{%
nie_adversarial_2020}
\APACinsertmetastar {%
nie_adversarial_2020}%
\begin{APACrefauthors}%
Nie, Y.%
, Williams, A.%
, Dinan, E.%
, Bansal, M.%
, Weston, J.%
\BCBL {}\ \BBA {} Kiela, D.%
\end{APACrefauthors}%
\unskip\
\newblock
\APACrefYearMonthDay{2020}{}{}.
\newblock
{\BBOQ}\APACrefatitle {Adversarial {NLI}: {A} {New} {Benchmark} for {Natural} {Language} {Understanding}} {Adversarial {NLI}: {A} {New} {Benchmark} for {Natural} {Language} {Understanding}}.{\BBCQ}
\newblock
\APACjournalVolNumPages{Proceedings of the 58th Annual Meeting of the Association for Computational Linguistics}{}{}{4885--4901}.
\newblock
\begin{APACrefDOI} \doi{10.18653/v1/2020.acl-main.441} \end{APACrefDOI}
\PrintBackRefs{\CurrentBib}

\bibitem [\protect \citeauthoryear {%
Nie%
, Zhou%
\BCBL {}\ \BBA {} Bansal%
}{%
Nie%
, Zhou%
\BCBL {}\ \BBA {} Bansal%
}{%
{\protect \APACyear {2020}}%
}]{%
nie_what_2020}
\APACinsertmetastar {%
nie_what_2020}%
\begin{APACrefauthors}%
Nie, Y.%
, Zhou, X.%
\BCBL {}\ \BBA {} Bansal, M.%
\end{APACrefauthors}%
\unskip\
\newblock
\APACrefYearMonthDay{2020}{}{}.
\newblock
{\BBOQ}\APACrefatitle {What {Can} {We} {Learn} from {Collective} {Human} {Opinions} on {Natural} {Language} {Inference} {Data}?} {What {Can} {We} {Learn} from {Collective} {Human} {Opinions} on {Natural} {Language} {Inference} {Data}?}{\BBCQ}
\newblock
\APACjournalVolNumPages{Proceedings of the 2020 Conference on Empirical Methods in Natural Language Processing (EMNLP)}{}{}{9131--9143}.
\newblock
\begin{APACrefDOI} \doi{10.18653/v1/2020.emnlp-main.734} \end{APACrefDOI}
\PrintBackRefs{\CurrentBib}

\bibitem [\protect \citeauthoryear {%
Nighojkar%
}{%
Nighojkar%
}{%
{\protect \APACyear {2024}}%
}]{%
nighojkar_inference-centric_2024}
\APACinsertmetastar {%
nighojkar_inference-centric_2024}%
\begin{APACrefauthors}%
Nighojkar, A.%
\end{APACrefauthors}%
\unskip\
\newblock
\APACrefYear{2024}.
\unskip\
\newblock
\APACrefbtitle {An {Inference}-{Centric} {Approach} to {Natural} {Language} {Processing} and {Cognitive} {Modeling}} {An {Inference}-{Centric} {Approach} to {Natural} {Language} {Processing} and {Cognitive} {Modeling}}\ \APACtypeAddressSchool {Ph.{D}.}{United States -- Florida}{University of South Florida}.
\unskip\
\newblock
\begin{APACrefURL} \url{https://www.proquest.com/dissertations-theses/inference-centric-approach-natural-language/docview/3084657113/se-2?accountid=14745} \end{APACrefURL}
\unskip\
\newblock
\APACrefnote{ISBN: 9798383569337 Publication Title: ProQuest Dissertations and Theses 31334853}
\PrintBackRefs{\CurrentBib}

\bibitem [\protect \citeauthoryear {%
Nighojkar%
, Jr%
\BCBL {}\ \BBA {} Licato%
}{%
Nighojkar%
\ \protect \BOthers {.}}{%
{\protect \APACyear {2023}}%
}]{%
nighojkar_no_2023}
\APACinsertmetastar {%
nighojkar_no_2023}%
\begin{APACrefauthors}%
Nighojkar, A.%
, Jr, A\BPBI L.%
\BCBL {}\ \BBA {} Licato, J.%
\end{APACrefauthors}%
\unskip\
\newblock
\APACrefYearMonthDay{2023}{}{}.
\newblock
{\BBOQ}\APACrefatitle {No {Strong} {Feelings} {One} {Way} or {Another}: {Re}-operationalizing {Neutrality} in {Natural} {Language} {Inference}} {No {Strong} {Feelings} {One} {Way} or {Another}: {Re}-operationalizing {Neutrality} in {Natural} {Language} {Inference}}.{\BBCQ}
\newblock
\APACjournalVolNumPages{Proceedings of the 17th Linguistic Annotation Workshop (LAW-XVII)}{}{}{199--210}.
\newblock
\begin{APACrefDOI} \doi{10.18653/v1/2023.law-1.20} \end{APACrefDOI}
\PrintBackRefs{\CurrentBib}

\bibitem [\protect \citeauthoryear {%
Nighojkar%
, Khlyzova%
\BCBL {}\ \BBA {} Licato%
}{%
Nighojkar%
\ \protect \BOthers {.}}{%
{\protect \APACyear {2022}}%
}]{%
nighojkar_cognitive_2022}
\APACinsertmetastar {%
nighojkar_cognitive_2022}%
\begin{APACrefauthors}%
Nighojkar, A.%
, Khlyzova, A.%
\BCBL {}\ \BBA {} Licato, J.%
\end{APACrefauthors}%
\unskip\
\newblock
\APACrefYearMonthDay{2022}{{\APACmonth{08}}}{}.
\newblock
\APACrefbtitle {Cognitive {Modeling} of {Semantic} {Fluency} {Using} {Transformers}.} {Cognitive {Modeling} of {Semantic} {Fluency} {Using} {Transformers}.}
\newblock
\APACaddressPublisher{}{arXiv}.
\newblock
\begin{APACrefURL} [{2024-08-20}]\url{http://arxiv.org/abs/2208.09719} \end{APACrefURL}
\newblock
\APACrefnote{arXiv:2208.09719 [cs]}
\PrintBackRefs{\CurrentBib}

\bibitem [\protect \citeauthoryear {%
Nighojkar%
\ \BBA {} Licato%
}{%
Nighojkar%
\ \BBA {} Licato%
}{%
{\protect \APACyear {2021}}%
{\protect \APACexlab {{\protect \BCnt {1}}}}}]{%
nighojkar_improving_2021}
\APACinsertmetastar {%
nighojkar_improving_2021}%
\begin{APACrefauthors}%
Nighojkar, A.%
\BCBT {}\ \BBA {} Licato, J.%
\end{APACrefauthors}%
\unskip\
\newblock
\APACrefYearMonthDay{2021{\protect \BCnt {1}}}{}{}.
\newblock
{\BBOQ}\APACrefatitle {Improving paraphrase detection with the adversarial paraphrasing task} {Improving paraphrase detection with the adversarial paraphrasing task}.{\BBCQ}.
\PrintBackRefs{\CurrentBib}

\bibitem [\protect \citeauthoryear {%
Nighojkar%
\ \BBA {} Licato%
}{%
Nighojkar%
\ \BBA {} Licato%
}{%
{\protect \APACyear {2021}}%
{\protect \APACexlab {{\protect \BCnt {2}}}}}]{%
nighojkar_mutual_2021}
\APACinsertmetastar {%
nighojkar_mutual_2021}%
\begin{APACrefauthors}%
Nighojkar, A.%
\BCBT {}\ \BBA {} Licato, J.%
\end{APACrefauthors}%
\unskip\
\newblock
\APACrefYearMonthDay{2021{\protect \BCnt {2}}}{}{}.
\newblock
{\BBOQ}\APACrefatitle {Mutual {Implication} as a {Measure} of {Textual} {Equivalence}} {Mutual {Implication} as a {Measure} of {Textual} {Equivalence}}.{\BBCQ}
\newblock
\APACjournalVolNumPages{The International FLAIRS Conference Proceedings}{34}{1}{}.
\newblock
\begin{APACrefDOI} \doi{10.32473/flairs.v34i1.128519} \end{APACrefDOI}
\PrintBackRefs{\CurrentBib}

\bibitem [\protect \citeauthoryear {%
Nisbett%
\ \BBA {} Wilson%
}{%
Nisbett%
\ \BBA {} Wilson%
}{%
{\protect \APACyear {1977}}%
}]{%
nisbett_telling_1977}
\APACinsertmetastar {%
nisbett_telling_1977}%
\begin{APACrefauthors}%
Nisbett, R\BPBI E.%
\BCBT {}\ \BBA {} Wilson, T\BPBI D.%
\end{APACrefauthors}%
\unskip\
\newblock
\APACrefYearMonthDay{1977}{{\APACmonth{05}}}{}.
\newblock
{\BBOQ}\APACrefatitle {Telling more than we can know: {Verbal} reports on mental processes.} {Telling more than we can know: {Verbal} reports on mental processes.}{\BBCQ}
\newblock
\APACjournalVolNumPages{Psychological Review}{84}{3}{231--259}.
\newblock
\begin{APACrefURL} [{2024-08-22}]\url{https://doi.apa.org/doi/10.1037/0033-295X.84.3.231} \end{APACrefURL}
\newblock
\begin{APACrefDOI} \doi{10.1037/0033-295X.84.3.231} \end{APACrefDOI}
\PrintBackRefs{\CurrentBib}

\bibitem [\protect \citeauthoryear {%
Nye%
, Tessler%
, Tenenbaum%
\BCBL {}\ \BBA {} Lake%
}{%
Nye%
\ \protect \BOthers {.}}{%
{\protect \APACyear {2021}}%
}]{%
Nye2021}
\APACinsertmetastar {%
Nye2021}%
\begin{APACrefauthors}%
Nye, M.%
, Tessler, M.%
, Tenenbaum, J.%
\BCBL {}\ \BBA {} Lake, B\BPBI M.%
\end{APACrefauthors}%
\unskip\
\newblock
\APACrefYearMonthDay{2021}{}{}.
\newblock
{\BBOQ}\APACrefatitle {Improving Coherence and Consistency in Neural Sequence Models with Dual-System, Neuro-Symbolic Reasoning} {Improving coherence and consistency in neural sequence models with dual-system, neuro-symbolic reasoning}.{\BBCQ}
\newblock
\BIn{} M.~Ranzato, A.~Beygelzimer, Y.~Dauphin, P.~Liang\BCBL {}\ \BBA {} J\BPBI W.~Vaughan\ (\BEDS), \APACrefbtitle {Advances in Neural Information Processing Systems} {Advances in neural information processing systems}\ (\BVOL~34, \BPGS\ 25192--25204).
\newblock
\APACaddressPublisher{}{Curran Associates, Inc.}
\newblock
\begin{APACrefURL} \url{https://proceedings.neurips.cc/paper/2021/file/d3e2e8f631bd9336ed25b8162aef8782-Paper.pdf} \end{APACrefURL}
\PrintBackRefs{\CurrentBib}

\bibitem [\protect \citeauthoryear {%
O'Neil%
}{%
O'Neil%
}{%
{\protect \APACyear {2017}}%
}]{%
oneil_weapons_2017}
\APACinsertmetastar {%
oneil_weapons_2017}%
\begin{APACrefauthors}%
O'Neil, C.%
\end{APACrefauthors}%
\unskip\
\newblock
\APACrefYear{2017}.
\newblock
\APACrefbtitle {Weapons of math destruction: how big data increases inequality and threatens democracy} {Weapons of math destruction: how big data increases inequality and threatens democracy}\ (\PrintOrdinal{First paperback edition}\ \BEd).
\newblock
\APACaddressPublisher{New York}{B/D/W/Y Broadway Books}.
\PrintBackRefs{\CurrentBib}

\bibitem [\protect \citeauthoryear {%
OpenAI%
\ \protect \BOthers {.}}{%
OpenAI%
\ \protect \BOthers {.}}{%
{\protect \APACyear {2023}}%
}]{%
openai_gpt-4_2023}
\APACinsertmetastar {%
openai_gpt-4_2023}%
\begin{APACrefauthors}%
OpenAI%
, Achiam, J.%
, Adler, S.%
, Agarwal, S.%
, Ahmad, L.%
, Akkaya, I.%
\BDBL {}Zoph, B.%
\end{APACrefauthors}%
\unskip\
\newblock
\APACrefYearMonthDay{2023}{}{}.
\newblock
\APACrefbtitle {{GPT}-4 {Technical} {Report}.} {{GPT}-4 {Technical} {Report}.}
\newblock
\APACaddressPublisher{}{arXiv}.
\newblock
\begin{APACrefURL} [{2024-08-12}]\url{https://arxiv.org/abs/2303.08774} \end{APACrefURL}
\newblock
\APACrefnote{Version Number: 6}
\newblock
\begin{APACrefDOI} \doi{10.48550/ARXIV.2303.08774} \end{APACrefDOI}
\PrintBackRefs{\CurrentBib}

\bibitem [\protect \citeauthoryear {%
Papineni%
, Roukos%
, Ward%
\BCBL {}\ \BBA {} Zhu%
}{%
Papineni%
\ \protect \BOthers {.}}{%
{\protect \APACyear {2001}}%
}]{%
papineni_bleu_2001}
\APACinsertmetastar {%
papineni_bleu_2001}%
\begin{APACrefauthors}%
Papineni, K.%
, Roukos, S.%
, Ward, T.%
\BCBL {}\ \BBA {} Zhu, W\BHBI J.%
\end{APACrefauthors}%
\unskip\
\newblock
\APACrefYearMonthDay{2001}{}{}.
\newblock
{\BBOQ}\APACrefatitle {{BLEU}: a method for automatic evaluation of machine translation} {{BLEU}: a method for automatic evaluation of machine translation}.{\BBCQ}
\newblock
\BIn{} \APACrefbtitle {Proceedings of the 40th {Annual} {Meeting} on {Association} for {Computational} {Linguistics} - {ACL} '02} {Proceedings of the 40th {Annual} {Meeting} on {Association} for {Computational} {Linguistics} - {ACL} '02}\ (\BPG~311).
\newblock
\APACaddressPublisher{Philadelphia, Pennsylvania}{Association for Computational Linguistics}.
\newblock
\begin{APACrefURL} [{2024-08-15}]\url{http://portal.acm.org/citation.cfm?doid=1073083.1073135} \end{APACrefURL}
\newblock
\begin{APACrefDOI} \doi{10.3115/1073083.1073135} \end{APACrefDOI}
\PrintBackRefs{\CurrentBib}

\bibitem [\protect \citeauthoryear {%
Pavlick%
\ \BBA {} Kwiatkowski%
}{%
Pavlick%
\ \BBA {} Kwiatkowski%
}{%
{\protect \APACyear {2019}}%
}]{%
pavlick_inherent_2019}
\APACinsertmetastar {%
pavlick_inherent_2019}%
\begin{APACrefauthors}%
Pavlick, E.%
\BCBT {}\ \BBA {} Kwiatkowski, T.%
\end{APACrefauthors}%
\unskip\
\newblock
\APACrefYearMonthDay{2019}{}{}.
\newblock
{\BBOQ}\APACrefatitle {Inherent {Disagreements} in {Human} {Textual} {Inferences}} {Inherent {Disagreements} in {Human} {Textual} {Inferences}}.{\BBCQ}
\newblock
\APACjournalVolNumPages{Transactions of the Association for Computational Linguistics}{7}{}{677--694}.
\newblock
\begin{APACrefDOI} \doi{10.1162/tacl_a_00293} \end{APACrefDOI}
\PrintBackRefs{\CurrentBib}

\bibitem [\protect \citeauthoryear {%
Peer%
\ \BBA {} Gamliel%
}{%
Peer%
\ \BBA {} Gamliel%
}{%
{\protect \APACyear {2013}}%
}]{%
Peer2013}
\APACinsertmetastar {%
Peer2013}%
\begin{APACrefauthors}%
Peer, E.%
\BCBT {}\ \BBA {} Gamliel, E.%
\end{APACrefauthors}%
\unskip\
\newblock
\APACrefYearMonthDay{2013}{01}{}.
\newblock
{\BBOQ}\APACrefatitle {Heuristics and biases in judicial decisions} {Heuristics and biases in judicial decisions}.{\BBCQ}
\newblock
\APACjournalVolNumPages{Court Review}{49}{}{114-118}.
\PrintBackRefs{\CurrentBib}

\bibitem [\protect \citeauthoryear {%
Pellert%
, Lechner%
, Wagner%
, Rammstedt%
\BCBL {}\ \BBA {} Strohmaier%
}{%
Pellert%
\ \protect \BOthers {.}}{%
{\protect \APACyear {2024}}%
}]{%
pellert_ai_2024}
\APACinsertmetastar {%
pellert_ai_2024}%
\begin{APACrefauthors}%
Pellert, M.%
, Lechner, C\BPBI M.%
, Wagner, C.%
, Rammstedt, B.%
\BCBL {}\ \BBA {} Strohmaier, M.%
\end{APACrefauthors}%
\unskip\
\newblock
\APACrefYearMonthDay{2024}{{\APACmonth{01}}}{}.
\newblock
{\BBOQ}\APACrefatitle {{AI} {Psychometrics}: {Assessing} the {Psychological} {Profiles} of {Large} {Language} {Models} {Through} {Psychometric} {Inventories}} {{AI} {Psychometrics}: {Assessing} the {Psychological} {Profiles} of {Large} {Language} {Models} {Through} {Psychometric} {Inventories}}.{\BBCQ}
\newblock
\APACjournalVolNumPages{Perspectives on Psychological Science}{}{}{17456916231214460}.
\newblock
\begin{APACrefURL} [{2024-09-01}]\url{http://journals.sagepub.com/doi/10.1177/17456916231214460} \end{APACrefURL}
\newblock
\begin{APACrefDOI} \doi{10.1177/17456916231214460} \end{APACrefDOI}
\PrintBackRefs{\CurrentBib}

\bibitem [\protect \citeauthoryear {%
Penedo%
\ \protect \BOthers {.}}{%
Penedo%
\ \protect \BOthers {.}}{%
{\protect \APACyear {2023}}%
}]{%
penedo_refinedweb_2023}
\APACinsertmetastar {%
penedo_refinedweb_2023}%
\begin{APACrefauthors}%
Penedo, G.%
, Malartic, Q.%
, Hesslow, D.%
, Cojocaru, R.%
, Cappelli, A.%
, Alobeidli, H.%
\BDBL {}Launay, J.%
\end{APACrefauthors}%
\unskip\
\newblock
\APACrefYearMonthDay{2023}{{\APACmonth{06}}}{}.
\newblock
\APACrefbtitle {The {RefinedWeb} {Dataset} for {Falcon} {LLM}: {Outperforming} {Curated} {Corpora} with {Web} {Data}, and {Web} {Data} {Only}.} {The {RefinedWeb} {Dataset} for {Falcon} {LLM}: {Outperforming} {Curated} {Corpora} with {Web} {Data}, and {Web} {Data} {Only}.}
\newblock
\APACaddressPublisher{}{arXiv}.
\newblock
\begin{APACrefURL} [{2024-08-20}]\url{http://arxiv.org/abs/2306.01116} \end{APACrefURL}
\newblock
\APACrefnote{arXiv:2306.01116 [cs]}
\PrintBackRefs{\CurrentBib}

\bibitem [\protect \citeauthoryear {%
Prescott%
\ \BBA {} Wilson%
}{%
Prescott%
\ \BBA {} Wilson%
}{%
{\protect \APACyear {2023}}%
}]{%
prescott_understanding_2023}
\APACinsertmetastar {%
prescott_understanding_2023}%
\begin{APACrefauthors}%
Prescott, T\BPBI J.%
\BCBT {}\ \BBA {} Wilson, S\BPBI P.%
\end{APACrefauthors}%
\unskip\
\newblock
\APACrefYearMonthDay{2023}{{\APACmonth{05}}}{}.
\newblock
{\BBOQ}\APACrefatitle {Understanding brain functional architecture through robotics} {Understanding brain functional architecture through robotics}.{\BBCQ}
\newblock
\APACjournalVolNumPages{Science Robotics}{8}{78}{eadg6014}.
\newblock
\begin{APACrefURL} [{2024-08-20}]\url{https://www.science.org/doi/10.1126/scirobotics.adg6014} \end{APACrefURL}
\newblock
\begin{APACrefDOI} \doi{10.1126/scirobotics.adg6014} \end{APACrefDOI}
\PrintBackRefs{\CurrentBib}

\bibitem [\protect \citeauthoryear {%
Quinlan%
}{%
Quinlan%
}{%
{\protect \APACyear {1986}}%
}]{%
quinlan_induction_1986}
\APACinsertmetastar {%
quinlan_induction_1986}%
\begin{APACrefauthors}%
Quinlan, J\BPBI R.%
\end{APACrefauthors}%
\unskip\
\newblock
\APACrefYearMonthDay{1986}{{\APACmonth{03}}}{}.
\newblock
{\BBOQ}\APACrefatitle {Induction of decision trees} {Induction of decision trees}.{\BBCQ}
\newblock
\APACjournalVolNumPages{Machine Learning}{1}{1}{81--106}.
\newblock
\begin{APACrefURL} [{2024-08-20}]\url{http://link.springer.com/10.1007/BF00116251} \end{APACrefURL}
\newblock
\begin{APACrefDOI} \doi{10.1007/BF00116251} \end{APACrefDOI}
\PrintBackRefs{\CurrentBib}

\bibitem [\protect \citeauthoryear {%
Rachlinski%
, Wistrich%
\BCBL {}\ \BBA {} Guthrie%
}{%
Rachlinski%
\ \protect \BOthers {.}}{%
{\protect \APACyear {2015}}%
}]{%
Rachlinski2015}
\APACinsertmetastar {%
Rachlinski2015}%
\begin{APACrefauthors}%
Rachlinski, J\BPBI J.%
, Wistrich, A\BPBI J.%
\BCBL {}\ \BBA {} Guthrie, C.%
\end{APACrefauthors}%
\unskip\
\newblock
\APACrefYearMonthDay{2015}{}{}.
\newblock
{\BBOQ}\APACrefatitle {Can Judges Make Reliable Numeric Judgments? Distorted Damages and Skewed Sentences} {Can judges make reliable numeric judgments? distorted damages and skewed sentences}.{\BBCQ}
\newblock
\APACjournalVolNumPages{Indiana Law Journal}{90}{}{}.
\PrintBackRefs{\CurrentBib}

\bibitem [\protect \citeauthoryear {%
Rafailov%
\ \protect \BOthers {.}}{%
Rafailov%
\ \protect \BOthers {.}}{%
{\protect \APACyear {2023}}%
}]{%
rafailov_direct_2023}
\APACinsertmetastar {%
rafailov_direct_2023}%
\begin{APACrefauthors}%
Rafailov, R.%
, Sharma, A.%
, Mitchell, E.%
, Ermon, S.%
, Manning, C\BPBI D.%
\BCBL {}\ \BBA {} Finn, C.%
\end{APACrefauthors}%
\unskip\
\newblock
\APACrefYearMonthDay{2023}{}{}.
\newblock
\APACrefbtitle {Direct {Preference} {Optimization}: {Your} {Language} {Model} is {Secretly} a {Reward} {Model}.} {Direct {Preference} {Optimization}: {Your} {Language} {Model} is {Secretly} a {Reward} {Model}.}
\newblock
\APACaddressPublisher{}{arXiv}.
\newblock
\begin{APACrefURL} [{2024-08-12}]\url{https://arxiv.org/abs/2305.18290} \end{APACrefURL}
\newblock
\APACrefnote{Version Number: 3}
\newblock
\begin{APACrefDOI} \doi{10.48550/ARXIV.2305.18290} \end{APACrefDOI}
\PrintBackRefs{\CurrentBib}

\bibitem [\protect \citeauthoryear {%
Raffel%
\ \protect \BOthers {.}}{%
Raffel%
\ \protect \BOthers {.}}{%
{\protect \APACyear {2023}}%
}]{%
raffel_exploring_2023}
\APACinsertmetastar {%
raffel_exploring_2023}%
\begin{APACrefauthors}%
Raffel, C.%
, Shazeer, N.%
, Roberts, A.%
, Lee, K.%
, Narang, S.%
, Matena, M.%
\BDBL {}Liu, P\BPBI J.%
\end{APACrefauthors}%
\unskip\
\newblock
\APACrefYearMonthDay{2023}{{\APACmonth{09}}}{}.
\newblock
\APACrefbtitle {Exploring the {Limits} of {Transfer} {Learning} with a {Unified} {Text}-to-{Text} {Transformer}.} {Exploring the {Limits} of {Transfer} {Learning} with a {Unified} {Text}-to-{Text} {Transformer}.}
\newblock
\APACaddressPublisher{}{arXiv}.
\newblock
\begin{APACrefURL} [{2024-08-20}]\url{http://arxiv.org/abs/1910.10683} \end{APACrefURL}
\newblock
\APACrefnote{arXiv:1910.10683 [cs, stat]}
\PrintBackRefs{\CurrentBib}

\bibitem [\protect \citeauthoryear {%
Robertson%
}{%
Robertson%
}{%
{\protect \APACyear {2004}}%
}]{%
robertson_understanding_2004}
\APACinsertmetastar {%
robertson_understanding_2004}%
\begin{APACrefauthors}%
Robertson, S.%
\end{APACrefauthors}%
\unskip\
\newblock
\APACrefYearMonthDay{2004}{{\APACmonth{10}}}{}.
\newblock
{\BBOQ}\APACrefatitle {Understanding inverse document frequency: on theoretical arguments for {IDF}} {Understanding inverse document frequency: on theoretical arguments for {IDF}}.{\BBCQ}
\newblock
\APACjournalVolNumPages{Journal of Documentation}{60}{5}{503--520}.
\newblock
\begin{APACrefURL} [{2024-08-22}]\url{https://www.emerald.com/insight/content/doi/10.1108/00220410410560582/full/html} \end{APACrefURL}
\newblock
\begin{APACrefDOI} \doi{10.1108/00220410410560582} \end{APACrefDOI}
\PrintBackRefs{\CurrentBib}

\bibitem [\protect \citeauthoryear {%
Roccas%
, Sagiv%
, Schwartz%
\BCBL {}\ \BBA {} Knafo%
}{%
Roccas%
\ \protect \BOthers {.}}{%
{\protect \APACyear {2002}}%
}]{%
roccas_big_2002}
\APACinsertmetastar {%
roccas_big_2002}%
\begin{APACrefauthors}%
Roccas, S.%
, Sagiv, L.%
, Schwartz, S\BPBI H.%
\BCBL {}\ \BBA {} Knafo, A.%
\end{APACrefauthors}%
\unskip\
\newblock
\APACrefYearMonthDay{2002}{{\APACmonth{06}}}{}.
\newblock
{\BBOQ}\APACrefatitle {The {Big} {Five} {Personality} {Factors} and {Personal} {Values}} {The {Big} {Five} {Personality} {Factors} and {Personal} {Values}}.{\BBCQ}
\newblock
\APACjournalVolNumPages{Personality and Social Psychology Bulletin}{28}{6}{789--801}.
\newblock
\begin{APACrefURL} [{2024-08-11}]\url{http://journals.sagepub.com/doi/10.1177/0146167202289008} \end{APACrefURL}
\newblock
\begin{APACrefDOI} \doi{10.1177/0146167202289008} \end{APACrefDOI}
\PrintBackRefs{\CurrentBib}

\bibitem [\protect \citeauthoryear {%
Saha%
\ \protect \BOthers {.}}{%
Saha%
\ \protect \BOthers {.}}{%
{\protect \APACyear {2024}}%
}]{%
saha_system-1x_2024}
\APACinsertmetastar {%
saha_system-1x_2024}%
\begin{APACrefauthors}%
Saha, S.%
, Prasad, A.%
, Chen, J\BPBI C\BHBI Y.%
, Hase, P.%
, Stengel-Eskin, E.%
\BCBL {}\ \BBA {} Bansal, M.%
\end{APACrefauthors}%
\unskip\
\newblock
\APACrefYearMonthDay{2024}{{\APACmonth{07}}}{}.
\newblock
\APACrefbtitle {System-1.x: {Learning} to {Balance} {Fast} and {Slow} {Planning} with {Language} {Models}.} {System-1.x: {Learning} to {Balance} {Fast} and {Slow} {Planning} with {Language} {Models}.}
\newblock
\APACaddressPublisher{}{arXiv}.
\newblock
\begin{APACrefURL} [{2024-08-11}]\url{http://arxiv.org/abs/2407.14414} \end{APACrefURL}
\newblock
\APACrefnote{arXiv:2407.14414 [cs]}
\PrintBackRefs{\CurrentBib}

\bibitem [\protect \citeauthoryear {%
Sahoo%
\ \protect \BOthers {.}}{%
Sahoo%
\ \protect \BOthers {.}}{%
{\protect \APACyear {2024}}%
}]{%
sahoo_systematic_2024}
\APACinsertmetastar {%
sahoo_systematic_2024}%
\begin{APACrefauthors}%
Sahoo, P.%
, Singh, A\BPBI K.%
, Saha, S.%
, Jain, V.%
, Mondal, S.%
\BCBL {}\ \BBA {} Chadha, A.%
\end{APACrefauthors}%
\unskip\
\newblock
\APACrefYearMonthDay{2024}{{\APACmonth{02}}}{}.
\newblock
\APACrefbtitle {A {Systematic} {Survey} of {Prompt} {Engineering} in {Large} {Language} {Models}: {Techniques} and {Applications}.} {A {Systematic} {Survey} of {Prompt} {Engineering} in {Large} {Language} {Models}: {Techniques} and {Applications}.}
\newblock
\APACaddressPublisher{}{arXiv}.
\newblock
\begin{APACrefURL} [{2024-08-20}]\url{http://arxiv.org/abs/2402.07927} \end{APACrefURL}
\newblock
\APACrefnote{arXiv:2402.07927 [cs]}
\PrintBackRefs{\CurrentBib}

\bibitem [\protect \citeauthoryear {%
Sainz%
\ \protect \BOthers {.}}{%
Sainz%
\ \protect \BOthers {.}}{%
{\protect \APACyear {2023}}%
}]{%
sainz_nlp_2023}
\APACinsertmetastar {%
sainz_nlp_2023}%
\begin{APACrefauthors}%
Sainz, O.%
, Campos, J.%
, García-Ferrero, I.%
, Etxaniz, J.%
, De~Lacalle, O\BPBI L.%
\BCBL {}\ \BBA {} Agirre, E.%
\end{APACrefauthors}%
\unskip\
\newblock
\APACrefYearMonthDay{2023}{}{}.
\newblock
{\BBOQ}\APACrefatitle {{NLP} {Evaluation} in trouble: {On} the {Need} to {Measure} {LLM} {Data} {Contamination} for each {Benchmark}} {{NLP} {Evaluation} in trouble: {On} the {Need} to {Measure} {LLM} {Data} {Contamination} for each {Benchmark}}.{\BBCQ}
\newblock
\BIn{} \APACrefbtitle {Findings of the {Association} for {Computational} {Linguistics}: {EMNLP} 2023} {Findings of the {Association} for {Computational} {Linguistics}: {EMNLP} 2023}\ (\BPGS\ 10776--10787).
\newblock
\APACaddressPublisher{Singapore}{Association for Computational Linguistics}.
\newblock
\begin{APACrefURL} [{2024-09-01}]\url{https://aclanthology.org/2023.findings-emnlp.722} \end{APACrefURL}
\newblock
\begin{APACrefDOI} \doi{10.18653/v1/2023.findings-emnlp.722} \end{APACrefDOI}
\PrintBackRefs{\CurrentBib}

\bibitem [\protect \citeauthoryear {%
Sap%
\ \protect \BOthers {.}}{%
Sap%
\ \protect \BOthers {.}}{%
{\protect \APACyear {2022}}%
}]{%
sap_annotators_2022}
\APACinsertmetastar {%
sap_annotators_2022}%
\begin{APACrefauthors}%
Sap, M.%
, Swayamdipta, S.%
, Vianna, L.%
, Zhou, X.%
, Choi, Y.%
\BCBL {}\ \BBA {} Smith, N.%
\end{APACrefauthors}%
\unskip\
\newblock
\APACrefYearMonthDay{2022}{}{}.
\newblock
{\BBOQ}\APACrefatitle {Annotators with {Attitudes}: {How} {Annotator} {Beliefs} {And} {Identities} {Bias} {Toxic} {Language} {Detection}} {Annotators with {Attitudes}: {How} {Annotator} {Beliefs} {And} {Identities} {Bias} {Toxic} {Language} {Detection}}.{\BBCQ}
\newblock
\APACjournalVolNumPages{Proceedings of the 2022 Conference of the North American Chapter of the Association for Computational Linguistics: Human Language Technologies}{}{}{5884--5906}.
\newblock
\begin{APACrefDOI} \doi{10.18653/v1/2022.naacl-main.431} \end{APACrefDOI}
\PrintBackRefs{\CurrentBib}

\bibitem [\protect \citeauthoryear {%
Schmidhuber%
}{%
Schmidhuber%
}{%
{\protect \APACyear {2015}}%
}]{%
schmidhuber_deep_2015}
\APACinsertmetastar {%
schmidhuber_deep_2015}%
\begin{APACrefauthors}%
Schmidhuber, J.%
\end{APACrefauthors}%
\unskip\
\newblock
\APACrefYearMonthDay{2015}{{\APACmonth{01}}}{}.
\newblock
{\BBOQ}\APACrefatitle {Deep learning in neural networks: {An} overview} {Deep learning in neural networks: {An} overview}.{\BBCQ}
\newblock
\APACjournalVolNumPages{Neural Networks}{61}{}{85--117}.
\newblock
\begin{APACrefURL} [{2024-08-20}]\url{https://linkinghub.elsevier.com/retrieve/pii/S0893608014002135} \end{APACrefURL}
\newblock
\begin{APACrefDOI} \doi{10.1016/j.neunet.2014.09.003} \end{APACrefDOI}
\PrintBackRefs{\CurrentBib}

\bibitem [\protect \citeauthoryear {%
Stanovich%
\ \BBA {} West%
}{%
Stanovich%
\ \BBA {} West%
}{%
{\protect \APACyear {2000}}%
}]{%
stanovich_individual_2000}
\APACinsertmetastar {%
stanovich_individual_2000}%
\begin{APACrefauthors}%
Stanovich, K\BPBI E.%
\BCBT {}\ \BBA {} West, R\BPBI F.%
\end{APACrefauthors}%
\unskip\
\newblock
\APACrefYearMonthDay{2000}{{\APACmonth{10}}}{}.
\newblock
{\BBOQ}\APACrefatitle {Individual differences in reasoning: {Implications} for the rationality debate?} {Individual differences in reasoning: {Implications} for the rationality debate?}{\BBCQ}
\newblock
\APACjournalVolNumPages{Behavioral and Brain Sciences}{23}{5}{645--665}.
\newblock
\begin{APACrefURL} [{2024-08-15}]\url{https://www.cambridge.org/core/product/identifier/S0140525X00003435/type/journal_article} \end{APACrefURL}
\newblock
\begin{APACrefDOI} \doi{10.1017/S0140525X00003435} \end{APACrefDOI}
\PrintBackRefs{\CurrentBib}

\bibitem [\protect \citeauthoryear {%
Stanovich%
, West%
\BCBL {}\ \BBA {} Toplak%
}{%
Stanovich%
\ \protect \BOthers {.}}{%
{\protect \APACyear {2018}}%
}]{%
stanovich_rationality_2018}
\APACinsertmetastar {%
stanovich_rationality_2018}%
\begin{APACrefauthors}%
Stanovich, K\BPBI E.%
, West, R\BPBI F.%
\BCBL {}\ \BBA {} Toplak, M\BPBI E.%
\end{APACrefauthors}%
\unskip\
\newblock
\APACrefYear{2018}.
\newblock
\APACrefbtitle {The rationality quotient: toward a test of rational thinking} {The rationality quotient: toward a test of rational thinking}\ (\PrintOrdinal{First paperback edition}\ \BEd).
\newblock
\APACaddressPublisher{Cambridge, Massachusetts London, England}{The MIT Press}.
\PrintBackRefs{\CurrentBib}

\bibitem [\protect \citeauthoryear {%
Stone%
\ \protect \BOthers {.}}{%
Stone%
\ \protect \BOthers {.}}{%
{\protect \APACyear {2020}}%
}]{%
stone_artificial_2020}
\APACinsertmetastar {%
stone_artificial_2020}%
\begin{APACrefauthors}%
Stone, M.%
, Aravopoulou, E.%
, Ekinci, Y.%
, Evans, G.%
, Hobbs, M.%
, Labib, A.%
\BDBL {}Machtynger, L.%
\end{APACrefauthors}%
\unskip\
\newblock
\APACrefYearMonthDay{2020}{{\APACmonth{04}}}{}.
\newblock
{\BBOQ}\APACrefatitle {Artificial intelligence ({AI}) in strategic marketing decision-making: a research agenda} {Artificial intelligence ({AI}) in strategic marketing decision-making: a research agenda}.{\BBCQ}
\newblock
\APACjournalVolNumPages{The Bottom Line}{33}{2}{183--200}.
\newblock
\begin{APACrefURL} [{2024-08-22}]\url{https://www.emerald.com/insight/content/doi/10.1108/BL-03-2020-0022/full/html} \end{APACrefURL}
\newblock
\begin{APACrefDOI} \doi{10.1108/BL-03-2020-0022} \end{APACrefDOI}
\PrintBackRefs{\CurrentBib}

\bibitem [\protect \citeauthoryear {%
Sullivan%
\ \BBA {} Artino%
}{%
Sullivan%
\ \BBA {} Artino%
}{%
{\protect \APACyear {2013}}%
}]{%
sullivan_analyzing_2013}
\APACinsertmetastar {%
sullivan_analyzing_2013}%
\begin{APACrefauthors}%
Sullivan, G\BPBI M.%
\BCBT {}\ \BBA {} Artino, A\BPBI R.%
\end{APACrefauthors}%
\unskip\
\newblock
\APACrefYearMonthDay{2013}{{\APACmonth{12}}}{}.
\newblock
{\BBOQ}\APACrefatitle {Analyzing and {Interpreting} {Data} {From} {Likert}-{Type} {Scales}} {Analyzing and {Interpreting} {Data} {From} {Likert}-{Type} {Scales}}.{\BBCQ}
\newblock
\APACjournalVolNumPages{Journal of Graduate Medical Education}{5}{4}{541--542}.
\newblock
\begin{APACrefURL} [{2024-08-22}]\url{https://meridian.allenpress.com/jgme/article/5/4/541/34037/Analyzing-and-Interpreting-Data-From-LikertType} \end{APACrefURL}
\newblock
\begin{APACrefDOI} \doi{10.4300/JGME-5-4-18} \end{APACrefDOI}
\PrintBackRefs{\CurrentBib}

\bibitem [\protect \citeauthoryear {%
Sun%
\ \protect \BOthers {.}}{%
Sun%
\ \protect \BOthers {.}}{%
{\protect \APACyear {2024}}%
}]{%
sun_survey_2024}
\APACinsertmetastar {%
sun_survey_2024}%
\begin{APACrefauthors}%
Sun, J.%
, Zheng, C.%
, Xie, E.%
, Liu, Z.%
, Chu, R.%
, Qiu, J.%
\BDBL {}Li, Z.%
\end{APACrefauthors}%
\unskip\
\newblock
\APACrefYearMonthDay{2024}{{\APACmonth{01}}}{}.
\newblock
\APACrefbtitle {A {Survey} of {Reasoning} with {Foundation} {Models}.} {A {Survey} of {Reasoning} with {Foundation} {Models}.}
\newblock
\APACaddressPublisher{}{arXiv}.
\newblock
\begin{APACrefURL} [{2024-08-20}]\url{http://arxiv.org/abs/2312.11562} \end{APACrefURL}
\newblock
\APACrefnote{arXiv:2312.11562 [cs]}
\PrintBackRefs{\CurrentBib}

\bibitem [\protect \citeauthoryear {%
Team%
}{%
Team%
}{%
{\protect \APACyear {2024}}%
}]{%
gemma_2024}
\APACinsertmetastar {%
gemma_2024}%
\begin{APACrefauthors}%
Team, G.%
\end{APACrefauthors}%
\unskip\
\newblock
\APACrefYearMonthDay{2024}{}{}.
\newblock
{\BBOQ}\APACrefatitle {Gemma} {Gemma}.{\BBCQ}
\newblock
\APACjournalVolNumPages{Online}{}{}{}.
\newblock
\begin{APACrefURL} \url{https://www.kaggle.com/m/3301} \end{APACrefURL}
\newblock
\begin{APACrefDOI} \doi{10.34740/KAGGLE/M/3301} \end{APACrefDOI}
\PrintBackRefs{\CurrentBib}

\bibitem [\protect \citeauthoryear {%
Teknium%
, theemozilla%
, karan4d%
\BCBL {}\ \BBA {} huemin\_art%
}{%
Teknium%
\ \protect \BOthers {.}}{%
{\protect \APACyear {2024}}%
}]{%
Nous-Hermes-2-Mistral-7B-DPO}
\APACinsertmetastar {%
Nous-Hermes-2-Mistral-7B-DPO}%
\begin{APACrefauthors}%
Teknium%
, theemozilla%
, karan4d%
\BCBL {}\ \BBA {} huemin\_art.%
\end{APACrefauthors}%
\unskip\
\newblock
\APACrefYearMonthDay{2024}{}{}.
\newblock
\APACrefbtitle {{N}ous {H}ermes 2 {M}istral 7{B} {DPO}.} {{N}ous {H}ermes 2 {M}istral 7{B} {DPO}.}
\newblock
\begin{APACrefURL} \url{https://huggingface.co/NousResearch/Nous-Hermes-2-Mistral-7B-DPO} \end{APACrefURL}
\PrintBackRefs{\CurrentBib}

\bibitem [\protect \citeauthoryear {%
Templeton%
}{%
Templeton%
}{%
{\protect \APACyear {2024}}%
}]{%
templeton2024scaling}
\APACinsertmetastar {%
templeton2024scaling}%
\begin{APACrefauthors}%
Templeton, A.%
\end{APACrefauthors}%
\unskip\
\newblock
\APACrefYear{2024}.
\newblock
\APACrefbtitle {Scaling monosemanticity: Extracting interpretable features from claude 3 sonnet} {Scaling monosemanticity: Extracting interpretable features from claude 3 sonnet}.
\newblock
\APACaddressPublisher{}{Anthropic}.
\PrintBackRefs{\CurrentBib}

\bibitem [\protect \citeauthoryear {%
{Tin Kam Ho}%
}{%
{Tin Kam Ho}%
}{%
{\protect \APACyear {1998}}%
}]{%
tin_kam_ho_random_1998}
\APACinsertmetastar {%
tin_kam_ho_random_1998}%
\begin{APACrefauthors}%
{Tin Kam Ho}.%
\end{APACrefauthors}%
\unskip\
\newblock
\APACrefYearMonthDay{1998}{{\APACmonth{08}}}{}.
\newblock
{\BBOQ}\APACrefatitle {The random subspace method for constructing decision forests} {The random subspace method for constructing decision forests}.{\BBCQ}
\newblock
\APACjournalVolNumPages{IEEE Transactions on Pattern Analysis and Machine Intelligence}{20}{8}{832--844}.
\newblock
\begin{APACrefURL} [{2024-08-20}]\url{http://ieeexplore.ieee.org/document/709601/} \end{APACrefURL}
\newblock
\begin{APACrefDOI} \doi{10.1109/34.709601} \end{APACrefDOI}
\PrintBackRefs{\CurrentBib}

\bibitem [\protect \citeauthoryear {%
Vatsal%
\ \BBA {} Dubey%
}{%
Vatsal%
\ \BBA {} Dubey%
}{%
{\protect \APACyear {2024}}%
}]{%
vatsal_survey_2024}
\APACinsertmetastar {%
vatsal_survey_2024}%
\begin{APACrefauthors}%
Vatsal, S.%
\BCBT {}\ \BBA {} Dubey, H.%
\end{APACrefauthors}%
\unskip\
\newblock
\APACrefYearMonthDay{2024}{{\APACmonth{07}}}{}.
\newblock
\APACrefbtitle {A {Survey} of {Prompt} {Engineering} {Methods} in {Large} {Language} {Models} for {Different} {NLP} {Tasks}.} {A {Survey} of {Prompt} {Engineering} {Methods} in {Large} {Language} {Models} for {Different} {NLP} {Tasks}.}
\newblock
\APACaddressPublisher{}{arXiv}.
\newblock
\begin{APACrefURL} [{2024-08-15}]\url{http://arxiv.org/abs/2407.12994} \end{APACrefURL}
\newblock
\APACrefnote{arXiv:2407.12994 [cs]}
\PrintBackRefs{\CurrentBib}

\bibitem [\protect \citeauthoryear {%
D.~Walton%
}{%
D.~Walton%
}{%
{\protect \APACyear {1985}}%
}]{%
Walton1985}
\APACinsertmetastar {%
Walton1985}%
\begin{APACrefauthors}%
Walton, D.%
\end{APACrefauthors}%
\unskip\
\newblock
\APACrefYear{1985}.
\newblock
\APACrefbtitle {Arguer's Position: A Pragmatic Study of Ad Hominem Attack, Criticism, Refutation, and Fallacy} {Arguer's position: A pragmatic study of ad hominem attack, criticism, refutation, and fallacy}.
\newblock
\APACaddressPublisher{}{Greenwood Press}.
\PrintBackRefs{\CurrentBib}

\bibitem [\protect \citeauthoryear {%
D.~Walton%
, Reed%
\BCBL {}\ \BBA {} Macagno%
}{%
D.~Walton%
\ \protect \BOthers {.}}{%
{\protect \APACyear {2008}}%
}]{%
Walton2008}
\APACinsertmetastar {%
Walton2008}%
\begin{APACrefauthors}%
Walton, D.%
, Reed, C.%
\BCBL {}\ \BBA {} Macagno, F.%
\end{APACrefauthors}%
\unskip\
\newblock
\APACrefYear{2008}.
\newblock
\APACrefbtitle {Argumentation Schemes} {Argumentation schemes}.
\newblock
\APACaddressPublisher{}{Cambridge University Press}.
\PrintBackRefs{\CurrentBib}

\bibitem [\protect \citeauthoryear {%
D\BPBI N.~Walton%
}{%
D\BPBI N.~Walton%
}{%
{\protect \APACyear {1990}}%
}]{%
Walton1990}
\APACinsertmetastar {%
Walton1990}%
\begin{APACrefauthors}%
Walton, D\BPBI N.%
\end{APACrefauthors}%
\unskip\
\newblock
\APACrefYearMonthDay{1990}{}{}.
\newblock
{\BBOQ}\APACrefatitle {What is Reasoning? What Is an Argument?} {What is reasoning? what is an argument?}{\BBCQ}
\newblock
\APACjournalVolNumPages{The Journal of Philosophy}{87}{8}{399--419}.
\newblock
\begin{APACrefURL} \url{http://www.jstor.org/stable/2026735} \end{APACrefURL}
\PrintBackRefs{\CurrentBib}

\bibitem [\protect \citeauthoryear {%
L.~Wang%
\ \protect \BOthers {.}}{%
L.~Wang%
\ \protect \BOthers {.}}{%
{\protect \APACyear {2023}}%
}]{%
wang_plan-and-solve_2023}
\APACinsertmetastar {%
wang_plan-and-solve_2023}%
\begin{APACrefauthors}%
Wang, L.%
, Xu, W.%
, Lan, Y.%
, Hu, Z.%
, Lan, Y.%
, Lee, R\BPBI K\BHBI W.%
\BCBL {}\ \BBA {} Lim, E\BHBI P.%
\end{APACrefauthors}%
\unskip\
\newblock
\APACrefYearMonthDay{2023}{}{}.
\newblock
{\BBOQ}\APACrefatitle {Plan-and-{Solve} {Prompting}: {Improving} {Zero}-{Shot} {Chain}-of-{Thought} {Reasoning} by {Large} {Language} {Models}} {Plan-and-{Solve} {Prompting}: {Improving} {Zero}-{Shot} {Chain}-of-{Thought} {Reasoning} by {Large} {Language} {Models}}.{\BBCQ}
\newblock
\APACjournalVolNumPages{arXiv}{}{}{}.
\PrintBackRefs{\CurrentBib}

\bibitem [\protect \citeauthoryear {%
X.~Wang%
\ \protect \BOthers {.}}{%
X.~Wang%
\ \protect \BOthers {.}}{%
{\protect \APACyear {2023}}%
}]{%
wang_self-consistency_2023}
\APACinsertmetastar {%
wang_self-consistency_2023}%
\begin{APACrefauthors}%
Wang, X.%
, Wei, J.%
, Schuurmans, D.%
, Le, Q.%
, Chi, E.%
, Narang, S.%
\BDBL {}Zhou, D.%
\end{APACrefauthors}%
\unskip\
\newblock
\APACrefYearMonthDay{2023}{{\APACmonth{03}}}{}.
\newblock
\APACrefbtitle {Self-{Consistency} {Improves} {Chain} of {Thought} {Reasoning} in {Language} {Models}.} {Self-{Consistency} {Improves} {Chain} of {Thought} {Reasoning} in {Language} {Models}.}
\newblock
\APACaddressPublisher{}{arXiv}.
\newblock
\begin{APACrefURL} [{2024-08-15}]\url{http://arxiv.org/abs/2203.11171} \end{APACrefURL}
\newblock
\APACrefnote{arXiv:2203.11171 [cs]}
\PrintBackRefs{\CurrentBib}

\bibitem [\protect \citeauthoryear {%
Wason%
}{%
Wason%
}{%
{\protect \APACyear {1966}}%
}]{%
Wason1966-WASR}
\APACinsertmetastar {%
Wason1966-WASR}%
\begin{APACrefauthors}%
Wason, P\BPBI C.%
\end{APACrefauthors}%
\unskip\
\newblock
\APACrefYearMonthDay{1966}{}{}.
\newblock
{\BBOQ}\APACrefatitle {Reasoning} {Reasoning}.{\BBCQ}
\newblock
\BIn{} P\BPBI C.~Wason\ (\BED), \APACrefbtitle {New Horizons in Psychology} {New horizons in psychology}\ (\BPGS\ 135--151).
\newblock
\APACaddressPublisher{}{Penguin Books}.
\PrintBackRefs{\CurrentBib}

\bibitem [\protect \citeauthoryear {%
Wei%
\ \protect \BOthers {.}}{%
Wei%
\ \protect \BOthers {.}}{%
{\protect \APACyear {2022}}%
}]{%
wei_chain_2022}
\APACinsertmetastar {%
wei_chain_2022}%
\begin{APACrefauthors}%
Wei, J.%
, Wang, X.%
, Schuurmans, D.%
, Bosma, M.%
, Ichter, B.%
, Xia, F.%
\BDBL {}Zhou, D.%
\end{APACrefauthors}%
\unskip\
\newblock
\APACrefYearMonthDay{2022}{}{}.
\newblock
{\BBOQ}\APACrefatitle {Chain of {Thought} {Prompting} {Elicits} {Reasoning} in {Large} {Language} {Models}} {Chain of {Thought} {Prompting} {Elicits} {Reasoning} in {Large} {Language} {Models}}.{\BBCQ}
\newblock
\APACjournalVolNumPages{arXiv}{}{}{}.
\newblock
\begin{APACrefDOI} \doi{10.48550/arxiv.2201.11903} \end{APACrefDOI}
\PrintBackRefs{\CurrentBib}

\bibitem [\protect \citeauthoryear {%
Wen%
, Cao%
, Yang%
, Yang%
\BCBL {}\ \BBA {} Liu%
}{%
Wen%
\ \protect \BOthers {.}}{%
{\protect \APACyear {2024}}%
}]{%
wen_affective-_2024}
\APACinsertmetastar {%
wen_affective-_2024}%
\begin{APACrefauthors}%
Wen, Z.%
, Cao, J.%
, Yang, Y.%
, Yang, R.%
\BCBL {}\ \BBA {} Liu, S.%
\end{APACrefauthors}%
\unskip\
\newblock
\APACrefYearMonthDay{2024}{{\APACmonth{03}}}{}.
\newblock
{\BBOQ}\APACrefatitle {Affective- {NLI}: {Towards} {Accurate} and {Interpretable} {Personality} {Recognition} in {Conversation}} {Affective- {NLI}: {Towards} {Accurate} and {Interpretable} {Personality} {Recognition} in {Conversation}}.{\BBCQ}
\newblock
\BIn{} \APACrefbtitle {2024 {IEEE} {International} {Conference} on {Pervasive} {Computing} and {Communications} ({PerCom})} {2024 {IEEE} {International} {Conference} on {Pervasive} {Computing} and {Communications} ({PerCom})}\ (\BPGS\ 184--193).
\newblock
\APACaddressPublisher{Biarritz, France}{IEEE}.
\newblock
\begin{APACrefURL} [{2024-09-01}]\url{https://ieeexplore.ieee.org/document/10494487/} \end{APACrefURL}
\newblock
\begin{APACrefDOI} \doi{10.1109/PerCom59722.2024.10494487} \end{APACrefDOI}
\PrintBackRefs{\CurrentBib}

\bibitem [\protect \citeauthoryear {%
Weston%
\ \BBA {} Sukhbaatar%
}{%
Weston%
\ \BBA {} Sukhbaatar%
}{%
{\protect \APACyear {2023}}%
}]{%
weston_system_2023}
\APACinsertmetastar {%
weston_system_2023}%
\begin{APACrefauthors}%
Weston, J.%
\BCBT {}\ \BBA {} Sukhbaatar, S.%
\end{APACrefauthors}%
\unskip\
\newblock
\APACrefYearMonthDay{2023}{{\APACmonth{11}}}{}.
\newblock
\APACrefbtitle {System 2 {Attention} (is something you might need too).} {System 2 {Attention} (is something you might need too).}
\newblock
\APACaddressPublisher{}{arXiv}.
\newblock
\begin{APACrefURL} [{2023-11-21}]\url{http://arxiv.org/abs/2311.11829} \end{APACrefURL}
\newblock
\APACrefnote{arXiv:2311.11829 [cs]}
\PrintBackRefs{\CurrentBib}

\bibitem [\protect \citeauthoryear {%
Williams%
, Nangia%
\BCBL {}\ \BBA {} Bowman%
}{%
Williams%
\ \protect \BOthers {.}}{%
{\protect \APACyear {2018}}%
}]{%
williams_broad-coverage_2018}
\APACinsertmetastar {%
williams_broad-coverage_2018}%
\begin{APACrefauthors}%
Williams, A.%
, Nangia, N.%
\BCBL {}\ \BBA {} Bowman, S.%
\end{APACrefauthors}%
\unskip\
\newblock
\APACrefYearMonthDay{2018}{}{}.
\newblock
{\BBOQ}\APACrefatitle {A {Broad}-{Coverage} {Challenge} {Corpus} for {Sentence} {Understanding} through {Inference}} {A {Broad}-{Coverage} {Challenge} {Corpus} for {Sentence} {Understanding} through {Inference}}.{\BBCQ}
\newblock
\APACjournalVolNumPages{Proceedings of the 2018 Conference of the North American Chapter of the Association for Computational Linguistics: Human Language Technologies, Volume 1 (Long Papers)}{}{}{1112--1122}.
\newblock
\begin{APACrefDOI} \doi{10.18653/v1/n18-1101} \end{APACrefDOI}
\PrintBackRefs{\CurrentBib}

\bibitem [\protect \citeauthoryear {%
Williams%
, Thrush%
\BCBL {}\ \BBA {} Kiela%
}{%
Williams%
\ \protect \BOthers {.}}{%
{\protect \APACyear {2022}}%
}]{%
williams-etal-2022-anlizing}
\APACinsertmetastar {%
williams-etal-2022-anlizing}%
\begin{APACrefauthors}%
Williams, A.%
, Thrush, T.%
\BCBL {}\ \BBA {} Kiela, D.%
\end{APACrefauthors}%
\unskip\
\newblock
\APACrefYearMonthDay{2022}{{\APACmonth{02}}}{}.
\newblock
{\BBOQ}\APACrefatitle {{ANLI}zing the Adversarial Natural Language Inference Dataset} {{ANLI}zing the adversarial natural language inference dataset}.{\BBCQ}
\newblock
\BIn{} A.~Ettinger, T.~Hunter\BCBL {}\ \BBA {} B.~Prickett\ (\BEDS), \APACrefbtitle {Proceedings of the Society for Computation in Linguistics 2022} {Proceedings of the society for computation in linguistics 2022}\ (\BPGS\ 23--54).
\newblock
\APACaddressPublisher{online}{Association for Computational Linguistics}.
\newblock
\begin{APACrefURL} \url{https://aclanthology.org/2022.scil-1.3} \end{APACrefURL}
\PrintBackRefs{\CurrentBib}

\bibitem [\protect \citeauthoryear {%
Wistrich%
, Rachlinski%
\BCBL {}\ \BBA {} Guthrie%
}{%
Wistrich%
\ \protect \BOthers {.}}{%
{\protect \APACyear {2015}}%
}]{%
Wistrich2015}
\APACinsertmetastar {%
Wistrich2015}%
\begin{APACrefauthors}%
Wistrich, A\BPBI J.%
, Rachlinski, J\BPBI J.%
\BCBL {}\ \BBA {} Guthrie, C.%
\end{APACrefauthors}%
\unskip\
\newblock
\APACrefYearMonthDay{2015}{}{}.
\newblock
{\BBOQ}\APACrefatitle {Heart Versus Head: Do Judges Follow the Law or Follow Their Feelings?.} {Heart versus head: Do judges follow the law or follow their feelings?.}{\BBCQ}
\newblock
\APACjournalVolNumPages{Texas Law Review}{93}{4}{855 - 923}.
\PrintBackRefs{\CurrentBib}

\bibitem [\protect \citeauthoryear {%
Workshop%
\ \protect \BOthers {.}}{%
Workshop%
\ \protect \BOthers {.}}{%
{\protect \APACyear {2023}}%
}]{%
workshop_bloom_2023}
\APACinsertmetastar {%
workshop_bloom_2023}%
\begin{APACrefauthors}%
Workshop, B.%
, Scao, T\BPBI L.%
, Fan, A.%
, Akiki, C.%
, Pavlick, E.%
, Ilić, S.%
\BDBL {}Wolf, T.%
\end{APACrefauthors}%
\unskip\
\newblock
\APACrefYearMonthDay{2023}{{\APACmonth{06}}}{}.
\newblock
\APACrefbtitle {{BLOOM}: {A} {176B}-{Parameter} {Open}-{Access} {Multilingual} {Language} {Model}.} {{BLOOM}: {A} {176B}-{Parameter} {Open}-{Access} {Multilingual} {Language} {Model}.}
\newblock
\APACaddressPublisher{}{arXiv}.
\newblock
\begin{APACrefURL} [{2024-08-20}]\url{http://arxiv.org/abs/2211.05100} \end{APACrefURL}
\newblock
\APACrefnote{arXiv:2211.05100 [cs]}
\PrintBackRefs{\CurrentBib}

\bibitem [\protect \citeauthoryear {%
Yang%
, Hu%
, Zhu%
\BCBL {}\ \BBA {} Nie%
}{%
Yang%
\ \protect \BOthers {.}}{%
{\protect \APACyear {2023}}%
}]{%
yang_belief_2023}
\APACinsertmetastar {%
yang_belief_2023}%
\begin{APACrefauthors}%
Yang, J.%
, Hu, Z.%
, Zhu, D.%
\BCBL {}\ \BBA {} Nie, D.%
\end{APACrefauthors}%
\unskip\
\newblock
\APACrefYearMonthDay{2023}{}{}.
\newblock
{\BBOQ}\APACrefatitle {Belief bias, conflict detection, and logical complexity} {Belief bias, conflict detection, and logical complexity}.{\BBCQ}
\newblock
\APACjournalVolNumPages{Current Psychology}{}{}{1--9}.
\newblock
\begin{APACrefDOI} \doi{10.1007/s12144-023-04562-9} \end{APACrefDOI}
\PrintBackRefs{\CurrentBib}

\bibitem [\protect \citeauthoryear {%
Yao%
\ \protect \BOthers {.}}{%
Yao%
\ \protect \BOthers {.}}{%
{\protect \APACyear {2023}}%
}]{%
yao_tree_2023}
\APACinsertmetastar {%
yao_tree_2023}%
\begin{APACrefauthors}%
Yao, S.%
, Yu, D.%
, Zhao, J.%
, Shafran, I.%
, Griffiths, T\BPBI L.%
, Cao, Y.%
\BCBL {}\ \BBA {} Narasimhan, K.%
\end{APACrefauthors}%
\unskip\
\newblock
\APACrefYearMonthDay{2023}{}{}.
\newblock
{\BBOQ}\APACrefatitle {Tree of {Thoughts}: {Deliberate} {Problem} {Solving} with {Large} {Language} {Models}} {Tree of {Thoughts}: {Deliberate} {Problem} {Solving} with {Large} {Language} {Models}}.{\BBCQ}
\newblock
\APACjournalVolNumPages{arXiv}{}{}{}.
\newblock
\begin{APACrefDOI} \doi{10.48550/arxiv.2305.10601} \end{APACrefDOI}
\PrintBackRefs{\CurrentBib}

\bibitem [\protect \citeauthoryear {%
Yu%
, Xu%
, Weston%
\BCBL {}\ \BBA {} Kulikov%
}{%
Yu%
\ \protect \BOthers {.}}{%
{\protect \APACyear {2024}}%
}]{%
yu_distilling_2024}
\APACinsertmetastar {%
yu_distilling_2024}%
\begin{APACrefauthors}%
Yu, P.%
, Xu, J.%
, Weston, J.%
\BCBL {}\ \BBA {} Kulikov, I.%
\end{APACrefauthors}%
\unskip\
\newblock
\APACrefYearMonthDay{2024}{{\APACmonth{07}}}{}.
\newblock
\APACrefbtitle {Distilling {System} 2 into {System} 1.} {Distilling {System} 2 into {System} 1.}
\newblock
\APACaddressPublisher{}{arXiv}.
\newblock
\begin{APACrefURL} [{2024-08-11}]\url{http://arxiv.org/abs/2407.06023} \end{APACrefURL}
\newblock
\APACrefnote{arXiv:2407.06023 [cs]}
\PrintBackRefs{\CurrentBib}

\bibitem [\protect \citeauthoryear {%
Zdaniuk%
}{%
Zdaniuk%
}{%
{\protect \APACyear {2014}}%
}]{%
michalos_ordinary_2014}
\APACinsertmetastar {%
michalos_ordinary_2014}%
\begin{APACrefauthors}%
Zdaniuk, B.%
\end{APACrefauthors}%
\unskip\
\newblock
\APACrefYearMonthDay{2014}{}{}.
\newblock
{\BBOQ}\APACrefatitle {Ordinary {Least}-{Squares} ({OLS}) {Model}} {Ordinary {Least}-{Squares} ({OLS}) {Model}}.{\BBCQ}
\newblock
\BIn{} A\BPBI C.~Michalos\ (\BED), \APACrefbtitle {Encyclopedia of {Quality} of {Life} and {Well}-{Being} {Research}} {Encyclopedia of {Quality} of {Life} and {Well}-{Being} {Research}}\ (\BPGS\ 4515--4517).
\newblock
\APACaddressPublisher{Dordrecht}{Springer Netherlands}.
\newblock
\begin{APACrefURL} [{2024-08-20}]\url{http://link.springer.com/10.1007/978-94-007-0753-5_2008} \end{APACrefURL}
\newblock
\begin{APACrefDOI} \doi{10.1007/978-94-007-0753-5_2008} \end{APACrefDOI}
\PrintBackRefs{\CurrentBib}

\bibitem [\protect \citeauthoryear {%
S.~Zhang%
, Gong%
\BCBL {}\ \BBA {} Choi%
}{%
S.~Zhang%
\ \protect \BOthers {.}}{%
{\protect \APACyear {2021}}%
}]{%
zhang_capturing_2021}
\APACinsertmetastar {%
zhang_capturing_2021}%
\begin{APACrefauthors}%
Zhang, S.%
, Gong, C.%
\BCBL {}\ \BBA {} Choi, E.%
\end{APACrefauthors}%
\unskip\
\newblock
\APACrefYearMonthDay{2021}{{\APACmonth{02}}}{}.
\newblock
{\BBOQ}\APACrefatitle {Capturing {Label} {Distribution}: {A} {Case} {Study} in {NLI}} {Capturing {Label} {Distribution}: {A} {Case} {Study} in {NLI}}.{\BBCQ}
\newblock
\APACjournalVolNumPages{arXiv}{}{}{}.
\newblock
\begin{APACrefURL} \url{http://arxiv.org/abs/2102.06859} \end{APACrefURL}
\PrintBackRefs{\CurrentBib}

\bibitem [\protect \citeauthoryear {%
S.~Zhang%
, Rudinger%
, Duh%
\BCBL {}\ \BBA {} Durme%
}{%
S.~Zhang%
\ \protect \BOthers {.}}{%
{\protect \APACyear {2017}}%
}]{%
zhang_ordinal_2017}
\APACinsertmetastar {%
zhang_ordinal_2017}%
\begin{APACrefauthors}%
Zhang, S.%
, Rudinger, R.%
, Duh, K.%
\BCBL {}\ \BBA {} Durme, B\BPBI V.%
\end{APACrefauthors}%
\unskip\
\newblock
\APACrefYearMonthDay{2017}{}{}.
\newblock
{\BBOQ}\APACrefatitle {Ordinal {Common}-sense {Inference}} {Ordinal {Common}-sense {Inference}}.{\BBCQ}
\newblock
\APACjournalVolNumPages{Transactions of the Association for Computational Linguistics}{5}{}{379--395}.
\newblock
\begin{APACrefDOI} \doi{10.1162/tacl_a_00068} \end{APACrefDOI}
\PrintBackRefs{\CurrentBib}

\bibitem [\protect \citeauthoryear {%
X\BPBI F.~Zhang%
\ \BBA {} Marneffe%
}{%
X\BPBI F.~Zhang%
\ \BBA {} Marneffe%
}{%
{\protect \APACyear {2021}}%
}]{%
zhang_identifying_2021}
\APACinsertmetastar {%
zhang_identifying_2021}%
\begin{APACrefauthors}%
Zhang, X\BPBI F.%
\BCBT {}\ \BBA {} Marneffe, M\BHBI C\BPBI d.%
\end{APACrefauthors}%
\unskip\
\newblock
\APACrefYearMonthDay{2021}{}{}.
\newblock
{\BBOQ}\APACrefatitle {Identifying inherent disagreement in natural language inference} {Identifying inherent disagreement in natural language inference}.{\BBCQ}
\newblock
\APACjournalVolNumPages{Proceedings of the 2021 Conference of the North American Chapter of the Association for Computational Linguistics: Human Language Technologies}{}{}{4908--4915}.
\newblock
\begin{APACrefDOI} \doi{10.18653/v1/2021.naacl-main.390} \end{APACrefDOI}
\PrintBackRefs{\CurrentBib}

\bibitem [\protect \citeauthoryear {%
Y.~Zhang%
, Yang%
, Yuan%
\BCBL {}\ \BBA {} Yao%
}{%
Y.~Zhang%
\ \protect \BOthers {.}}{%
{\protect \APACyear {2024}}%
}]{%
zhang_cumulative_2024}
\APACinsertmetastar {%
zhang_cumulative_2024}%
\begin{APACrefauthors}%
Zhang, Y.%
, Yang, J.%
, Yuan, Y.%
\BCBL {}\ \BBA {} Yao, A\BPBI C\BHBI C.%
\end{APACrefauthors}%
\unskip\
\newblock
\APACrefYearMonthDay{2024}{{\APACmonth{04}}}{}.
\newblock
\APACrefbtitle {Cumulative {Reasoning} with {Large} {Language} {Models}.} {Cumulative {Reasoning} with {Large} {Language} {Models}.}
\newblock
\APACaddressPublisher{}{arXiv}.
\newblock
\begin{APACrefURL} [{2024-08-15}]\url{http://arxiv.org/abs/2308.04371} \end{APACrefURL}
\newblock
\APACrefnote{arXiv:2308.04371 [cs]}
\PrintBackRefs{\CurrentBib}

\bibitem [\protect \citeauthoryear {%
Zhou%
, Nie%
\BCBL {}\ \BBA {} Bansal%
}{%
Zhou%
\ \protect \BOthers {.}}{%
{\protect \APACyear {2022}}%
}]{%
zhou_distributed_2022}
\APACinsertmetastar {%
zhou_distributed_2022}%
\begin{APACrefauthors}%
Zhou, X.%
, Nie, Y.%
\BCBL {}\ \BBA {} Bansal, M.%
\end{APACrefauthors}%
\unskip\
\newblock
\APACrefYearMonthDay{2022}{}{}.
\newblock
{\BBOQ}\APACrefatitle {Distributed {NLI}: {Learning} to {Predict} {Human} {Opinion} {Distributions} for {Language} {Reasoning}} {Distributed {NLI}: {Learning} to {Predict} {Human} {Opinion} {Distributions} for {Language} {Reasoning}}.{\BBCQ}
\newblock
\BIn{} \APACrefbtitle {Findings of the {Association} for {Computational} {Linguistics}: {ACL} 2022} {Findings of the {Association} for {Computational} {Linguistics}: {ACL} 2022}\ (\BPGS\ 972--987).
\newblock
\APACaddressPublisher{Dublin, Ireland}{Association for Computational Linguistics}.
\newblock
\begin{APACrefURL} [{2024-08-20}]\url{https://aclanthology.org/2022.findings-acl.79} \end{APACrefURL}
\newblock
\begin{APACrefDOI} \doi{10.18653/v1/2022.findings-acl.79} \end{APACrefDOI}
\PrintBackRefs{\CurrentBib}

\bibitem [\protect \citeauthoryear {%
Zollman%
, Sirnoorkar%
\BCBL {}\ \BBA {} Laverty%
}{%
Zollman%
\ \protect \BOthers {.}}{%
{\protect \APACyear {2023}}%
}]{%
zollman_analyzing_2023}
\APACinsertmetastar {%
zollman_analyzing_2023}%
\begin{APACrefauthors}%
Zollman, D\BPBI A.%
, Sirnoorkar, A.%
\BCBL {}\ \BBA {} Laverty, J\BPBI T.%
\end{APACrefauthors}%
\unskip\
\newblock
\APACrefYearMonthDay{2023}{{\APACmonth{10}}}{}.
\newblock
{\BBOQ}\APACrefatitle {Analyzing {AI} and student responses through the lens of sensemaking and mechanistic reasoning} {Analyzing {AI} and student responses through the lens of sensemaking and mechanistic reasoning}.{\BBCQ}
\newblock
\BIn{} \APACrefbtitle {2023 {Physics} {Education} {Research} {Conference} {Proceedings}} {2023 {Physics} {Education} {Research} {Conference} {Proceedings}}\ (\BPGS\ 415--420).
\newblock
\APACaddressPublisher{Sacramento, CA}{American Association of Physics Teachers}.
\newblock
\begin{APACrefURL} [{2024-08-20}]\url{https://www.per-central.org/items/detail.cfm?ID=16619} \end{APACrefURL}
\newblock
\begin{APACrefDOI} \doi{10.1119/perc.2023.pr.Zollman} \end{APACrefDOI}
\PrintBackRefs{\CurrentBib}

\end{thebibliography}
